%% file: main.tex
\documentclass{article}
\usepackage[dvipsnames]{xcolor}
\usepackage{float}
\usepackage[pdftex]{graphicx}
\usepackage{tikz}
\usepackage{pgfplots}
\pgfplotsset{compat=1.15}

\usepackage{answers}
\usepackage{xcolor}
\usepackage{setspace}
\usepackage{graphicx}
\usepackage[shortlabels]{enumitem}
\usepackage{multicol}
\usepackage{silence}
\usepackage{mathrsfs}
\usepackage[margin=1in]{geometry} 
\usepackage{amsmath,amsthm,amssymb}
\usepackage{mathtools}
\usepackage[utf8]{inputenc}
\usepackage[english]{babel}
\usepackage{caption}
\usepackage{subcaption}
\usepackage[square,sort,comma,numbers]{natbib}
\usepackage{bbm}
\usepackage{breqn}
\usepackage{subfiles}
\usepackage{amsmath,amssymb}
\usepackage[english]{babel}
\usepackage[nottoc]{tocbibind}

\usepackage[toc,page]{appendix}
\newcommand{\stoptocwriting}{%
  \addtocontents{toc}{\protect\setcounter{tocdepth}{-5}}}
\newcommand{\resumetocwriting}{%
  \addtocontents{toc}{\protect\setcounter{tocdepth}{\arabic{tocdepth}}}}
\addto{\captionsenglish}{}

\usepackage{thmtools}
\usepackage{thm-restate}

\usepackage{hyperref}
\hypersetup{unicode,hidelinks}

\usepackage{cleveref}

\usetikzlibrary{patterns}

\newtheorem{theorem}{Theorem}[section]
\newtheorem{corollary}{Corollary}[theorem]
\newtheorem{lemma}[theorem]{Lemma}
\newtheorem{fact}[theorem]{Fact}
\newtheorem{definition}[theorem]{Definition}
\newtheorem{assumption}{Assumption}

\newcommand{\wt}{\widetilde}
\newcommand{\wh}{\widehat}

\DeclareMathOperator{\nvar}{\wt{var}}

\newcommand{\R}{\mathbb{R}}

\let\Pr\relax
\DeclareMathOperator*{\Pr}{\mathbb{P}}
\renewcommand{\Im}{\text{Im}}
\renewcommand{\Re}{\text{Re}}
\newcommand{\Var}{\text{Var}}

\newcommand{\lam}{\lambda}
\newcommand{\eps}{\varepsilon}
\newcommand{\1}{\mathbbm{1}}

\newcommand{\norm}[1]{\|#1\|}
\newcommand{\abs}[1]{|#1| }
\newcommand{\Poi}{\text{Poi}}
\DeclareMathOperator*{\E}{\mathbb{E}}
\newcommand{\PE}{\Tilde \E}

\input{libtheorems}

\title{Sharp Constants in Uniformity Testing via the Huber Statistic}
\author{Shivam Gupta\\The University of Texas at Austin\\\texttt{shivamgupta@utexas.edu} \and Eric Price\\The University of Texas at Austin\\\texttt{ecprice@cs.utexas.edu}}

\begin{document}

\maketitle

\begin{abstract}%
  Uniformity testing is one of the most well-studied problems in
  property testing, with many known test statistics, including ones
  based on counting collisions, singletons, and the empirical TV
  distance.  It is known that the optimal sample complexity to
  distinguish the uniform distribution on $m$ elements from any
  $\eps$-far distribution with $1-\delta$ probability is
  $n = \Theta(\frac{\sqrt{m \log (1/\delta)}}{\eps^2} + \frac{\log
    (1/\delta)}{\eps^2})$, which is achieved by the empirical TV
  tester.  Yet in simulation, these theoretical analyses are
  misleading: in many cases, they do not correctly rank order the
  performance of existing testers, even in an asymptotic regime of all
  parameters tending to $0$ or $\infty$.

We explain this discrepancy by studying the \emph{constant factors}
required by the algorithms.  We show that the collisions tester
achieves a sharp maximal constant in the number of standard deviations
of separation between uniform and non-uniform inputs.  We then
introduce a new tester based on the Huber loss, and show that it not
only matches this separation, but also has tails corresponding to a
Gaussian with this separation.  This leads to a sample complexity of
$(1 + o(1))\frac{\sqrt{m \log (1/\delta)}}{\eps^2}$ in the regime
where this term is dominant, unlike all other existing testers.
\end{abstract}

\stoptocwriting
\input{intro}
\input{proofoverview}
 \newpage
\bibliographystyle{alpha}
\bibliography{references}
\newpage
\renewcommand{\contentsname}{Contents of Appendix}
 \tableofcontents
 \resumetocwriting
\newpage
\appendix
%
%
\input{variance-optimal.tex}
\newpage
\input{preliminaries}

\newpage
\input{huber_statistic}
\newpage
\input{mgf_computation_lemmas}
\input{lemmas}

\newpage
\input{lower_bounds}
\newpage
\input{tv_tester_superlinear}
\newpage
\input{squared_statistic}
\newpage
\input{empty_bins_n_Theta_m_eps_to_0}

\end{document}

%% file: libtheorems.tex
\usepackage{etoolbox}

\makeatletter

\newcommand{\define}[4][ignore]{%
  \ifstrequal{#1}{ignore}{}{
  \@namedef{thmtitle@#2}{#1}}%
  \@namedef{thm@#2}{#4}%
  \@namedef{thmtypen@#2}{lemma}%
  \newtheorem{thmtype@#2}[theorem]{#3}%
  \newtheorem*{thmtypealt@#2}{#3~\ref{#2}}%
}

\newcommand{\state}[1]{%
  \@namedef{curthm}{#1}
  \@ifundefined{thmtitle@#1}{
  \begin{thmtype@#1}
    }{
  \begin{thmtype@#1}[\@nameuse{thmtitle@#1}]
  }
    \label{#1}
    \@nameuse{thm@#1}
  \end{thmtype@#1}
  \@ifundefined{thmdone@#1}{
  \@namedef{thmdone@#1}{stated}%
  }{}
}

\newcommand{\restate}[1]{%
  \@namedef{curthm}{#1}
  \@ifundefined{thmtitle@#1}{
    \begin{thmtypealt@#1}
    }{
  \begin{thmtypealt@#1}[\@nameuse{thmtitle@#1}]
  }
    \@nameuse{thm@#1}
  \end{thmtypealt@#1}
  \@ifundefined{thmdone@#1}{
  \@namedef{thmdone@#1}{stated}%
  }{}
}

\newcommand{\thmlabel}[1]{
  \@ifundefined{thmdone@\@nameuse{curthm}}{\label{#1}
    }{\tag*{\eqref{#1}}}
}
\makeatother

%% file: intro.tex
\section{Introduction}

Property testing of distributions is an area of study initiated
in~\citep{goldreich2011testing} and~\citep{batu2000testing}.  The
foundation of these works is a test for \emph{uniformity}: given $n$
samples from a distribution $q$ on $[m]$, can we distinguish the case
that $q$ is uniform from the case that $q$ is $\eps$-far from uniform,
with probability $1-\delta$?  The remarkable result is that this is
often possible for $n \ll m$, when we cannot learn the actual
distribution.  Over the years, several different tests and bounds
have been established for uniformity.  In this paper we better
understand and explain the relative performance of these testers, then
introduce a new uniformity tester that outperforms all of them.

The first uniformity tester introduced was the \emph{collisions
  tester}~\citep{goldreich2011testing,batu2000testing}, which
counts the number of collisions among the samples.  It is equivalent
to Pearson's $\chi^2$ test, or any other statistic quadratic in the
histogram.  It succeeds with constant probability for
$n = O(\sqrt{m}/\eps^2)$~\citep{DGPP19}, which is
optimal~\citep{paninski2008coincidence}.


What happens for high-probability bounds?  Naive repetition gives a
multiplicative $O(\log \frac{1}{\delta})$ loss, but this can be
improved: Huang and Meyn \citep{DBLP:journals/tit/HuangM13} showed
that the \emph{singletons tester}~\citep{paninski2008coincidence}
achieves $\sqrt{m \log \frac{1}{\delta}}/\eps^2$, but only in the
setting of $n = o(m)$ and $\eps = \Omega(1)$.  The collisions tester,
however, really does involve a $\log \frac{1}{\delta}$
loss~\citep{peebles}.


Achieving optimal dependence of the whole range is the empirical
\emph{TV tester}~\citep{DGPP18}, which measures the TV distance between the empirical
distribution and uniform.  It needs
\begin{align}
  \label{eq:nbound}
  n = O\left(\frac{\sqrt{m \log \frac{1}{\delta}}}{\eps^2} + \frac{\log \frac{1}{\delta}}{\eps^2}\right)
\end{align}
which is optimal in all settings of parameters.

To summarize 20 years of theory, the TV tester is asymptotically
optimal while the collisions tester has poor $\delta$ dependence and
the singletons tester is good for $n \ll m$ but fails when $n \gg m$.
This suggests that, given an actual example of a uniformity testing
problem, the TV tester is as good as possible.

In Figure~\ref{fig:experiment-fig} we compare the TV tester to the
collisions tester in simulation.  We test the uniform distribution
against the distribution that puts $\frac{1+2 \eps}{m}$ mass on half
the bins, and $\frac{1 - 2 \eps}{m}$ mass on the remaining bins. This
is the worst case $\eps$-far distribution for both these
testers~\citep{DGPP18}. We find that, contrary to the theoretical
prediction, the collisions tester outperforms the TV tester on the
parameters we consider.  In our first experiment, with $m = n = 10^4$
and $\eps = 1/8$, the TV tester has twice the error rate as the
collisions tester (3.3\% vs 1.7\%).  In our second experiment, with
$m = n = 10^5$ and $\eps = 1/10$, the gap \emph{widens} to a factor
$10$ ($10^{-4}$ vs $10^{-5}$) despite the error rate $\delta$---the parameter the
collisions tester is suboptimal in---becoming much smaller.  This means
that our theory is giving the \emph{wrong advice}: a practitioner
should prefer the collisions tester to the TV tester here.


To better explain this, and to develop a new tester that outperforms
all existing ones, we need to start considering constant factors.

\paragraph{Designing a new tester.}
How should we design an efficient tester for uniformity?  We consider
``separable'' testers that take as input the histogram $Y_j$ (so $Y_j$
is the number of samples equal to $j$), compute a statistic
\[
  S = \sum_{j=1}^m f(Y_j),
\]
and output YES or NO based on whether $S$ lies below some threshold
$\tau$.  Existing testers are all either of this form, or use this as
the main subroutine (e.g., after Poissonization or taking the median
of multiple attempts).  Differences lie in the choice of $f$.
Quadratic functions $f(k) = (k - n/m)^2$ or $f(k) = \binom{k}{2}$ give
the $\chi^2$ or collisions
tester~\citep{goldreich2011testing,batu2000testing}, which are
equivalent because $\sum_j Y_j = n$ is fixed, so the two statistics
$S$ are linearly related.  The TV tester~\citep{DGPP18} uses
$f(k) = \abs{k - n/m}$, while the singletons
tester~\citep{paninski2008coincidence} uses $f(k) = 1_{k = 1}$.  But
how would one design $f$ from first principles to work well?

In this paper we introduce a natural approach to designing a test
statistic with good asymptotic constants.  First, we find the test
statistic $f$ that maximizes the number of standard deviations of
separation between YES and NO instances; then, we modify the tails of
$f$ so that $S$ has Gaussian tails.  This approach is summarized in
Figure~\ref{fig:table}.

\begin{figure}[H]
  \centering
  \begin{tabular}{|c|c|c|}
    \hline
    Tester & Optimal variance? & Subgaussian tails?\\
    \hline
    \hline
    Collisions/$\chi^2$ & \textbf{Yes} (Theorem~\ref{thm:quadraticopt}) & No (Theorem~\ref{thm:peebles})\\
    \hline
    TV & No (Theorem~\ref{thm:tv}) & \textbf{Yes}\\
    \hline
    Huber (new) & \textbf{Yes} & \textbf{Yes} (Theorem~\ref{thm:huber})\\
    \hline
  \end{tabular}
  \caption{Our main contributions are (1) that the collisions
    statistic achieves optimal variance, and (2) that the Huber
    statistic can get high-probability bounds matching this variance.}
  \label{fig:table}
\end{figure}

\paragraph{Step 1: Optimize variance.}
Intuitively, $S$ is a sum of $m$ terms $f(Y_j)$ that are \emph{nearly}
independent, so we expect central limit-type behavior
\[
  S \stackrel{\scalebox{.7}{$\approx$}}{\sim} N(\E[S], \Var[S]).
\]
That is, we expect our separable statistic to behave like a Gaussian,
with expectation and variance that depend on the particular statistic.
Because the hard alternative distributions $q$ are very close to $p$,
typically $\Var_q[S] = (1 + o(1)) \Var_p[S]$.  Then our ability to
distinguish $p$ and $q$ depends on how this variance compares to the
separation in means: we want to minimize this \emph{normalized variance}
\[
  \nvar_{p,q}(S) := \frac{\Var_p[S]}{(\E_q[S] - \E_p[S])^2}.
\]

We can set our threshold to lie halfway between $\E_p[S]$ and
$\E_q[S]$, so that, under the Gaussian approximation, the error
probability will be given by
\begin{align}
  \delta \approx \exp\left(-\frac{\left(\left(\E_q[S] -
          \E_p[S]\right)/2\right)^2}{2 \Var_p[S]} \right) =
  \exp\left(-\frac{1}{8 \nvar_{p, q}(S)} \right)\label{eq:deltagauss}
\end{align}
For any $q$, $\min_f \nvar_{p,q}(S_f)$ is a quadratic program in $f$,
so we can compute the variance-minimizing $f$ for any setting of
parameters.  We can also approximate it analytically in the asymptotic
limit.  We find that the quadratic statistics (like collisions or
$\chi^2$) are near-optimal:
\begin{restatable}{theorem}{quadraticopt}\label{thm:quadraticopt}
  Let $\eps^2 \ll \frac{n}{m} \lesssim 1$ and $n,m,1/\eps \to \infty$ with
  $m \gtrsim 1/\eps^4$.  Any separable statistic $S$ has normalized
  variance
  \[
    \nvar_{p, q}(S)  \geq (1 + o(1))\frac{1}{8} \frac{m}{n^2\eps^4}
  \]
  between the uniform distribution $p$ and the balanced nonuniform
  distribution $q$ with $q_k = \frac{1 \pm 2\eps}{m}$.

  Quadratic statistics (like collisions or $\chi^2$) match this, getting
  \begin{align}\label{eq:sep}
    \nvar_{p, q}(S)  \leq (1 + o(1))\frac{1}{8} \frac{m}{n^2\eps^4}
  \end{align}
  for \emph{any} $\eps$-far distribution $q$.
\end{restatable}

Theorem~\ref{thm:quadraticopt} shows that, if the Gaussian
approximations hold, the collisions tester has optimal constants.
Per~\eqref{eq:deltagauss}, for failure probability $\delta$ we need
$\nvar_{p,q}(S) = \frac{1}{8 \log \frac{1}{\delta}}$, or
\begin{align}\label{eq:nboundtight}
  n = (1 + o(1))\frac{1}{\eps^2} \sqrt{m \log \frac{1}{\delta}}
\end{align}
samples.  This matches the optimal complexity~\eqref{eq:nbound} in the
large-$m$ regime, but with a sharp constant of $1$.  (Sharp in the
sense that no other separable statistic behaves better under its
Gaussian approximation.)  One could also trade off the false positive
and false negative errors by choosing a different threshold between
the means, getting
\begin{align}
  n = (1 + o(1))\frac{1}{\eps^2} \sqrt{m}\cdot \frac{\sqrt{\log \frac{1}{\delta_+}} + \sqrt{\log \frac{1}{\delta_-}}}{2}.\label{eq:gaussposneg}
\end{align}
for false positive/negative probabilities $\delta_+/\delta_-$.

By contrast, the TV tester has a constant factor worse normalized
variance than the quadratic tester (we shall state this constant
precisely later).  Therefore the Gaussian approximation loses a
constant factor relative to~\eqref{eq:nboundtight}, and it would be
very surprising if the actual statistic avoided this
inefficiency.\footnote{Unsurprisingly, as we show in
  Theorem~\ref{thm:tv}, the inefficiency is real.}  So the Gaussian
approximations predict the actual Figure~\ref{fig:experiment-fig}
behavior.

Unfortunately, the Gaussian approximation does \emph{not} hold in
general for the collisions statistic, so it does not
get~\eqref{eq:nboundtight} or~\eqref{eq:gaussposneg}.  See
Appendix~\ref{app:peebles} for a detailed example, due
to~\citep{peebles}, showing that for exponentially small $\delta$ the
collisions tester does not achieve~\eqref{eq:nbound} for any constant
much less $1 + o(1)$.


\paragraph{Step 2: Massage the tails.}
The Gaussian tail bound implying~\eqref{eq:nboundtight} and~\eqref{eq:gaussposneg} is given by
its moment generating function, so it would suffice to bound the MGF
of $S$.  The problem is that $Y_j$ has roughly exponential tails for every $j$, so
the MGF of $Y_j^2$ does not have a good bound.  To get a good MGF for
$f(Y_j)$, we need to look at an $f$ with at most linear growth.

So this is our situation: the quadratic $f(Y_j) = Y_j^2$ has
near-optimal variance but a very large MGF, while the TV statistic
$f(Y_j) = \abs{Y_j - n/m}$ has suboptimal variance but a pretty good MGF.
Introducing the linear tail with $f(Y_j) = \abs{Y_j - n/m}$ is
how~\citep{DGPP18} achieved the $\sqrt{\log \frac{1}{\delta}}$
dependence, but the worse variance means it inherently performs worse
than the quadratic for large $\delta$ where the Gaussian approximation
holds.

To achieve both good variance and MGF, we should \emph{start} with the
good-variance quadratic statistic, then attenuate the tail behavior to
get good concentration.  We do this with the \emph{Huber} loss
\[
  f(Y_j) = h_{\beta}(Y_j - n/m)
\]
for
\[
  h_{\beta}(x) := \min(x^2, 2\beta \abs{x} - \beta^2).
\]
If we choose a tradeoff point $\beta \gg 1 + \sqrt{n/m}$, most $Y_j$
will lie in the quadratic region and we still get the variance
bound~\eqref{eq:sep}.  But now the MGF is bounded.  We show that, for
a large range of parameters, this leads to the tester that matches a
Gaussian with the optimal variance:
\begin{restatable}[Huber]{theorem}{huber}
\label{thm:huber}
The Huber statistic for appropriate $\beta$
achieves~\eqref{eq:nboundtight} for $n/m \ll 1/\eps^2$,
$\eps, \delta \ll 1$, and $m \geq C\log n$ for sufficiently large
constant $C$.  It
achieves~\eqref{eq:gaussposneg} under the same conditions and $\delta_-, \delta_+ \ll 1$.
\end{restatable}

Combined with Theorem~\ref{thm:quadraticopt}, Theorem~\ref{thm:huber}
shows that the Huber statistic gets the optimal variance over
separable statistics and matches the Gaussian concentration with this
variance.

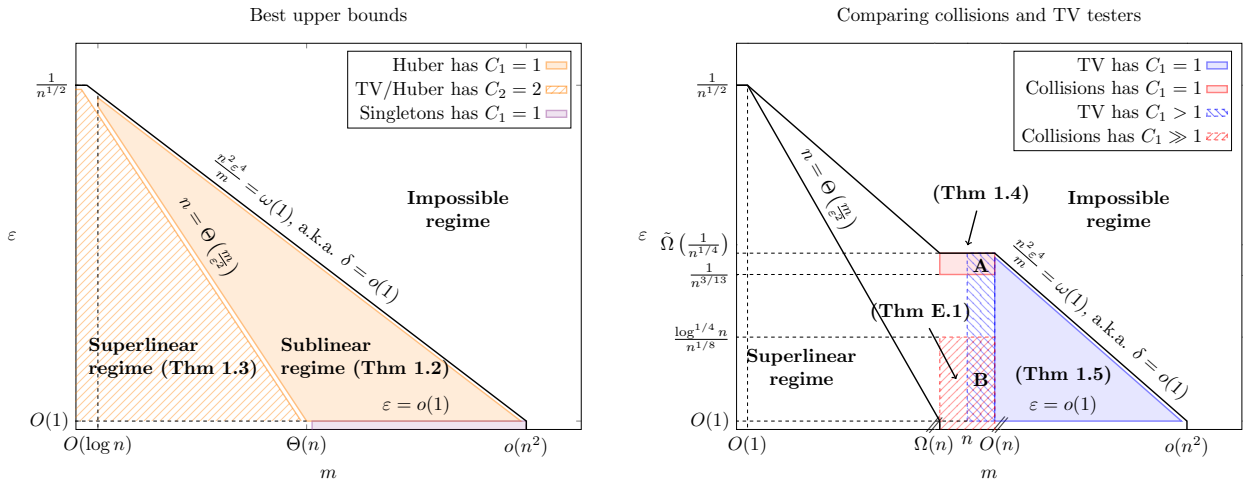
\begin{figure}
    \centering
    \begin{subfigure}[b]{0.48\textwidth}
    \hspace{-6mm}
      \input{best_upper_bounds_picture}
      \caption{The Huber statistic achieves the best constant over almost the entire region of $\eps \ll 1$.  The singletons statistic was previously known to achieve this for $\eps = \Omega(1)$ and $m \gg n$~\cite{DBLP:journals/tit/HuangM13}.}
      \label{fig:huber-bounds}
     \end{subfigure}
    \hfill
    \begin{subfigure}[b]{0.48\textwidth}
    \hspace{-8mm}
    \input{comparing_tv_and_collisions_picture_small}
      \caption{The constant $C_1$ in different $(n, \eps, \delta)$ parameter regimes.  In region \textbf{A}, the collisions tester performs better than the TV tester. In region \textbf{B}, both the collisions and TV tester have $C_1 > 1$.}
      \label{fig:tv-vs-coll-results}
     \end{subfigure}
\caption{Our results in different regimes.}\label{fig:results}
\end{figure}

\begin{figure}[H]
  \centering
  {
    \includegraphics[width=0.5\textwidth, trim= 50 180 0 160, clip]{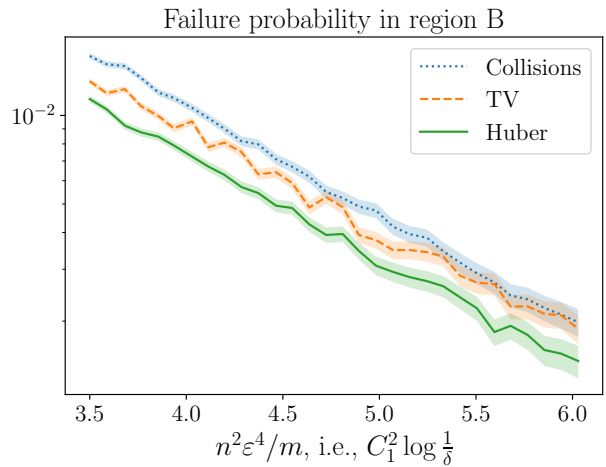}%
    \label{fig:plot}
  }
  \caption{Empirical failure probability of different testers when
    $n = m = (.7/\eps)^{8.1}$, which is in region \textbf{B} of
    Figure~\ref{fig:tv-vs-coll-results}.  The $x$ axis is
    $n^2\eps^4/m$, which should be linear in $\log \frac{1}{\delta}$
    per \eqref{eq:C1C2}.  The shaded region shows two
    standard deviations of uncertainty.}\label{fig:experiment-fig}
\end{figure}

The parameter regime is illustrated in Figure~\ref{fig:huber-bounds}.
The first three asymptotic conditions for Theorem~\ref{thm:huber}
delineate the boundaries of the ``sublinear'' regime, where testing is
possible, nontrivial, and the asymptotic sample
complexity~\eqref{eq:nbound} is dominated by the
$\frac{1}{\eps^2}\sqrt{m \log \frac{1}{\delta}}$ term.  The last
condition, that $m \geq C \log n$, is likely an artifact of our
analysis but is pretty mild.


The rest of our results look at other regimes and other testers, and
we summarize our results in Figure~\ref{fig:results}.  If we express
the sample complexity~\eqref{eq:nbound} as
\begin{align}\label{eq:C1C2}
  n = (C_1 + o(1))\frac{\sqrt{m \log \frac{1}{\delta}}}{\eps^2} + (C_2 + o(1))\frac{\log \frac{1}{\delta}}{\eps^2}
\end{align}
then we can express the constants $C_1, C_2$ in different regimes of
$(n, m, \eps)$.  In the ``sublinear regime'' of
$n/m \ll \frac{1}{\eps^2}$, where we cannot reliably estimate the
distribution, what matters is $C_1$.  For $n = m$, when
$\log n \ll \log \frac{1}{\delta} \ll n^{1/13}$ so that
$\Tilde \Omega(n^{-1/4}) \ll \eps \ll n^{-3/13}$, that is, in regime A
of small $\eps$/large $\delta$, collisions gets $C_1 = 1$ and TV has
$C_1 > 1$; in regime B of large $\eps$/small $\delta$, collisions gets
$C_1 \gg 1$ and TV has $1 < C_1 = O(1)$.  The Huber statistic, by
contrast, gets $C_1 = 1$ for almost the whole regime.  In the
superlinear regime, where the empirical distribution is
$\eps/2$-accurate, a simple union bound shows that the TV statistic
(and hence Huber statistic for $\beta = 0$) gets the optimal $C_2 = 2$:
\begin{restatable}[Superlinear regime]{theorem}{superlinear}\label{thm:tv_superlinear}
  For $n/m \gg 1/\eps^2$ and $\eps \ll 1$, the TV statistic achieves
  \[
    n = (2 + o(1)) \frac{\log \frac{1}{\delta}}{\eps^2}
  \]
  and no other tester can do better.
\end{restatable}

\paragraph{Analysis of collisions.}  While the $\chi^2$/collisions
tester does not match the Gaussian tails to
achieve~\eqref{eq:nboundtight} everywhere, it still is a sum of
mostly-independent variables and so looks like a Gaussian outside the
extreme tails.  Hence the Gaussian
approximation~\eqref{eq:nboundtight} ought to hold when $\delta$ isn't
too small.  Indeed, we show this is true for $n = \Theta(m)$ and
intermediate $\delta$:
\begin{restatable}[Collisions for large $\delta$]{theorem}{collisions}
\label{thm:collisions}
  The quadratic statistic achieves~\eqref{eq:nboundtight} for
  $n/m = \Theta(1)$,
  $\log n \ll \log \frac{1}{\delta} \ll n^{1/13}$ and $\eps \ll 1$.
\end{restatable}

\paragraph{Analysis of TV.}
The TV tester, for $n \leq m$, is equivalent to the tester that counts
empty bins.  We show that this has
\[
  \max_{q: \norm{p-q}_{TV} \geq \eps} \nvar_{p,q}(f) = (1 + o(1))\frac{(e^{n/m}-1-n/m)}{4(n/m)^2} \frac{m}{\eps^4 n^2}.
\]
rather than~\eqref{eq:sep}.  For $n \ll m$ these are equivalent, but
for $n = m$ it is 44\% larger, leading to about $20\%$ more samples.

\begin{restatable}[TV]{theorem}{tv}
\label{thm:tv}
  The TV statistic uses
  \[
    n = (1 + o(1)) \sqrt{\frac{2(e^{n/m}-1-n/m)}{(n/m)^2}} \frac{\sqrt{m \log \frac{1}{\delta}}}{\eps^2} 
  \]
  for $n \leq m$, $n \gg 1$, and $\eps, \delta \ll 1$.
\end{restatable}

Like Theorem~\ref{thm:huber}, both Theorem~\ref{thm:collisions} and
Theorem~\ref{thm:tv} work by showing the Gaussian approximation is
accurate.  Thus one could also trade off false positive/negative
probabilities, with a
$\frac{1}{2}(\sqrt{\log \frac{1}{\delta_-}} + \sqrt{\log
  \frac{1}{\delta_+}})$ dependence.

\paragraph{Experimental performance.}
In Figure~\ref{fig:experiment-fig}, we compare the empirical
performance of the new Huber tester to the existing collisions and TV
testers in a synthetic experiment.  The experiment has $m = n$,
$\eps = .7/n^{1/8.1}$ with alternative distribution
$q = \frac{1 \pm 2\eps}{n}$, and varies $n$ from $200$ to $600$.  This
is in region \textbf{B} of Figure~\ref{fig:tv-vs-coll-results}, and as
predicted we find that the Huber tester has lower failure probability
than the TV or collisions testers.

\subsection{Related work}

The past twenty years have seen a large body of work in distribution
testing; see~\citep{goldreich2017introduction,cannone} for surveys of
the area.  Uniformity testing has been either the basis for, or a
necessary subproblem in, many such results.  Such extensions include
testing
identity~\citep{batufirst,chan2014optimal,goldreich2017introduction,diakonikolas-kane16,
  valiant2017automatic, diakonikolas2020optimal}, testing
independence~\citep{canonne2018testing}, and testing uniformity over
unknown domains~\citep{batu2017generalized,diakonikolas2017sharp}.  One
particularly clean relation is that you can black-box reduce testing
identity to a fixed distribution $p$ to uniformity testing with only a
constant factor loss in parameters~\citep{goldreich2017introduction}.

Most of the above results do not focus on the dependence on $\delta$;
exceptions include~\citep{DGPP18,
  kim2020minimax,diakonikolas2020optimal,DBLP:journals/tit/HuangM13} which give
algorithms within constant factors of optimal for testing uniformity,
identity, and independence.

Lower bounds for uniformity testing started with an $\Omega(\sqrt{m})$
bound in~\citep{goldreich2011testing}, followed by $\Omega(\sqrt{m}/\eps^2)$
in~\citep{paninski2008coincidence} and
$\Omega(\frac{1}{\eps^2}\sqrt{m\log 1/\delta} + \frac{\log
  1/\delta}{\eps^2})$ in~\citep{DGPP18}.

When it comes to constant factors in distribution testing, the
classical regime of $\eps, m$ constant and $n \to \infty$ was analyzed
in~\citep{hoeffding1965multinomial} and the likelihood ratio test was
shown to be optimal.  Alternatively, for $m, \delta$ constant and
$n, 1/\eps \to \infty$, Pearson's $\chi^2$ tester---the quadratic tester---is known to be
asymptotically near optimal for identity testing
(see~\citep{lehmann2005testing}, Chapter 14).

The most closely related work to our paper is Huang and
Meyn~\citep{DBLP:journals/tit/HuangM13}, which (unlike the classical
results) studies constant factors in a regime where all of
$n, m, 1/\delta \to \infty$.  They consider the singletons tester, and
show that $C_1 = 1$ for constant $\eps$ and $n \ll m$.  They also show
that no algorithm can do better in this regime.  However, for
$n = \Theta(m)$ the singletons tester loses constant factors and for
$n \geq O(m \log m)$ it fails with high probability.

\subsection{Future work}
As discussed above, uniformity testing has been the building block for
many other distribution testing problems, such as identity and
independence testing.  Where there are direct reductions (as in
testing identity to a fixed
distribution~\citep{goldreich2017introduction}), these reductions lose
constant factors.  However, these tests still involve statistics that
are the sum of mostly independent random variables.  We believe that
our approach to constructing a test statistic---find a statistic to
optimize performance of the Gaussian approximation, then adjust it to
match the Gaussian tails---could lead to higher performance testers in
these problems as well.

Second, there are some settings of parameters that we have not
analyzed.  Most interesting would be to analyze the intermediate
regime of $\frac{n}{m} = \Theta(\frac{1}{\eps^2})$, where both sample
complexity terms in~\eqref{eq:nbound} are significant.

Third, we could consider constant factors for high probability bounds
in other settings.  For example, it is known by the Cram\'er-Rao bound
\citep{cramer46} that the maximum likelihood estimator (MLE) in
parametric statistics converges to a Gaussian with variance equal to
the inverse of the Fisher information under a broad set of
assumptions; but the tails of this estimator are less well understood,
and could likely be improved by modifying the estimator to be less
sensitive to outliers. Other examples lie in streaming algorithms.
There has been a line of work on understanding the constants in the space
complexity of cardinality estimation in streams
\citep{Flajolet07hyperloglog:the, Ertl2017NewCE, lang17, pettie21}, but
these have focused on the constant $\delta$ regime. We believe our
techniques could lead to optimal high probability bounds on the space
complexity for this problem.  Alternatively, for problems like heavy
hitters~\citep{CCF02,MP14}, the analysis has focused on the
high $\delta$ regime and ignored constant factors; but the underlying
algorithms involve sums of random variables that ought to converge to
Gaussians.

%% file: best_upper_bounds_picture.tex
\usetikzlibrary{patterns}
\begin{tikzpicture}[scale=0.45, every node/.style={scale=1.6}]
    \pgfdeclarepatternformonly{south west lines}{\pgfqpoint{-0pt}{-0pt}}{\pgfqpoint{3pt}{3pt}}{\pgfqpoint{3pt}{3pt}}{
        \pgfsetlinewidth{0.4pt}
        \pgfpathmoveto{\pgfqpoint{0pt}{0pt}}
        \pgfpathlineto{\pgfqpoint{3pt}{3pt}}
        \pgfpathmoveto{\pgfqpoint{2.8pt}{-.2pt}}
        \pgfpathlineto{\pgfqpoint{3.2pt}{.2pt}}
        \pgfpathmoveto{\pgfqpoint{-.2pt}{2.8pt}}
        \pgfpathlineto{\pgfqpoint{.2pt}{3.2pt}}
        \pgfusepath{stroke}}
\pgfdeclarepatternformonly{south east lines}{\pgfqpoint{-0pt}{-0pt}}{\pgfqpoint{3pt}{3pt}}{\pgfqpoint{3pt}{3pt}}{
    \pgfsetlinewidth{0.4pt}
    \pgfpathmoveto{\pgfqpoint{0pt}{3pt}}
    \pgfpathlineto{\pgfqpoint{3pt}{0pt}}
    \pgfpathmoveto{\pgfqpoint{.2pt}{-.2pt}}
    \pgfpathlineto{\pgfqpoint{-.2pt}{.2pt}}
    \pgfpathmoveto{\pgfqpoint{3.2pt}{2.8pt}}
    \pgfpathlineto{\pgfqpoint{2.8pt}{3.2pt}}
    \pgfusepath{stroke}}

\begin{axis}[
    title=Best upper bounds,
    xmin=0.8,
    xmax=10,
    xlabel=$m$,
    ylabel=$\eps$,
    ylabel style={rotate=-90},
    width=6.5in,
    height=13cm,
    legend style={legend cell align=right,legend plot pos=right},
    xtick={1.2, 5, 9},
    xticklabels={$O(\log n)$, $\Theta(n)$, $o(n^2)$},
    ymin=0.8,
    ymax=10,
    ytick={1, 9},
    yticklabels={$O(1)$, $\frac{1}{n^{1/2}}$}]
    \addlegendentry{Huber has $C_1 = 1$}
    \addplot [draw=orange, very thick, fill =orange, fill opacity=0.15, draw opacity=0.5, area legend]coordinates{(5, 1) (1.2, 8.5) (1.2, 8.7) (8.9, 1) (5, 1)};
    \addlegendentry{TV/Huber has $C_2 = 2$}
    \addplot[very thick, pattern=south west lines, pattern color=orange, draw=orange, draw opacity=0.5, fill opacity=0.8, area legend] coordinates{(0, 1) (0, 8.9) (.9, 8.9) (4.9, 1)};
    \addlegendentry{Singletons has $C_1 = 1$}
    \addplot[very thick, fill=Purple, draw=Purple, draw opacity=0.5, fill opacity=0.15, area legend] coordinates{(5.1, 0) (5.1, 1) (9, 1) (9, 0) (5, 0)};
    \addplot[draw=none] coordinates{(1,9) (5,1)} node[pos=0.35, above right, sloped] {$n=\Theta\left(\frac{m}{\eps^2}\right)$};
    \addplot[draw=none] coordinates{(1, 9) (9, 1)} node[pos=0.25, above right, sloped] {$\frac{n^2 \eps^4}{m} = \omega(1)$, a.k.a. $\delta = o(1)$};
    \addplot[draw=none] coordinates{(5, 1) (9, 1)} node[pos=0.5, above] {$\eps = o(1)$};
    \addplot[dashed] coordinates{(1.2, 0) (1.2, 8.8)};
    \addplot[dashed] coordinates{(0, 1) (5, 1)};
    \addplot[dashed] coordinates{(0, 9) (1, 9)};
    
    \addplot[very thick] coordinates {(0, 9) (1, 9) (9, 1) (9, 0)};
    \node[anchor=west, font=\bf] (log_1_over_delta) at (axis cs:.7, 2.5) {
      \begin{tabular}{l}
        Superlinear\\regime (Thm~\ref{thm:tv_superlinear})
      \end{tabular}
    };
    \node[anchor=west, font=\bf] (sublinear) at (axis cs:4.2, 2.5) {
      \begin{tabular}{l}
        Sublinear\\regime (Thm~\ref{thm:huber})
      \end{tabular}
    };
    \node[anchor=west, font=\bf] (impossible_regime) at (axis cs:6.5, 6) {
      \begin{tabular}{c}
        Impossible\\regime
      \end{tabular}
    };
\end{axis}

\end{tikzpicture}

%% file: comparing_tv_and_collisions_picture_small.tex
\usetikzlibrary{patterns}
\begin{tikzpicture}[scale=0.45, every node/.style={scale=1.6}]
\usetikzlibrary{pgfplots.groupplots}


\begin{axis}[
    title=Comparing collisions and TV testers,
    xmin=0.8,
    xmax=10,
    xlabel=$m$,
    ylabel=$\eps$,
    ylabel style={rotate=-90},
    width=6.5in,
    height=13cm,
    legend style={legend cell align=right,legend plot pos=right},
    xtick={1, 4.4, 5, 5.6, 9},
    xticklabels={$O(1)$,  $\Omega(n)$, $n$, $O(n)$, $o(n^2)$},
    extra x ticks ={4.4, 5.6},
    extra x tick style={grid=none, tick label style={xshift=0cm,yshift=.45cm, rotate=-20}},
    extra x tick label={\color{black}{/\!\!/}},
    ymin=0.8,
    ymax=10,
    ytick={1, 3, 4.4692, 5.2, 9},
    yticklabels={$O(1)$, $\frac{\log^{1/4} n}{n^{1/8}}$,  $\frac{1}{n^{3/13}}$, $\Tilde\Omega\left(\frac{1}{n^{1/4}}\right)$,  $\frac{1}{n^{1/2}}$},
    major tick length=0,
    ]
    \addlegendentry{TV has $C_1 = 1$}
    \addplot [very thick, draw=blue, area legend, fill=blue, fill opacity=0.1, draw opacity=0.5]coordinates{(5.5, 1) (5.5, 4.9) (8.9, 1) (5.5, 1)};
    \addlegendentry{Collisions has $C_1 = 1$}
    \addplot[very thick, draw=red, area legend, fill=red, fill opacity=0.1, draw opacity=0.5] coordinates{(4.5, 5) (5.5, 5) (5.5, 4.492) (4.5, 4.492) (4.5, 5) };
    \addlegendentry{TV has $C_1 > 1$ }
    \addplot[dashed, very thick, draw=blue, area legend, pattern=north west lines, pattern color=blue, fill opacity=0.7, draw opacity=0.5] coordinates{(5, 1) (5, 5) (5.5, 5) (5.5, 1) (5, 1)};
    \addlegendentry{Collisions has $C_1 \gg 1$}
    \addplot[dashed, pattern=north east lines, very thick, draw=red, area legend, pattern color=red, fill opacity=0.7, draw opacity=0.5] coordinates{(4.5, 0) (4.5, 3) (5.5, 3) (5.5, 0) (4.5, 0)};
    \addplot[draw=none] coordinates{(1,9) (4.5,1)} node[pos=0.2, above right, sloped] {$n=\Theta\left(\frac{m}{\eps^2}\right)$};
    \addplot[draw=none] coordinates{(5.5, 5) (9, 1)} node[pos=0, above right, sloped] {$\frac{n^2 \eps^4}{m} = \omega(1)$, a.k.a.  $\delta = o(1)$};
    \addplot[draw=none] coordinates{(4.5, 1) (9, 1)} node[pos=0.5, above] {$\eps = o(1)$};
    \addplot[dashed] coordinates{(1, 0) (1, 9)};
    \addplot[dashed] coordinates{(0, 1) (5, 1)};
    \addplot[dashed] coordinates{(0, 5) (4.5, 5)};
    \addplot[dashed] coordinates{(0, 4.492) (4.5, 4.492)};
    \addplot[dashed] coordinates{(0, 3) (4.5, 3)};
    
    \addplot[very thick] coordinates{(0, 9) (1, 9) (4.5, 5) (5.5, 5) (9, 1) (9, 0)};
    
    \addplot[very thick] coordinates {(0, 9) (1, 9) (4.5, 1) (4.5, 0)};
    
    \node[anchor=center, font=\bf](source_A) at (axis cs:5.25, 4.716) {A};
    \node[anchor=center, font=\bf](source_B) at (axis cs:5.25, 2) {B};
    \node[anchor=center, font=\bf](source_peeb) at (axis cs:4.15, 3.6) {(Thm~\ref{thm:peebles})};
    \addplot[draw=none] coordinates{(4.5, 1.7) (9, 1.7)} node[pos=0.5, above, font=\bf] {(Thm~\ref{thm:tv})};
    \node[anchor=center, font=\bf](source_thmcoll) at (axis cs:5.25, 6.4) {(Thm~\ref{thm:collisions})};
    \draw[->, very thick, shorten >= 0.2cm] (source_thmcoll) to (source_A.north west);
    \draw[->, very thick] (source_peeb) to (axis cs:4.8, 2.);
    
    \node[anchor=west, font=\bf] (log_1_over_delta) at (axis cs:.65, 2.25) {
      \begin{tabular}{c}
        Superlinear\\regime
      \end{tabular}
    };
    \node[anchor=west, font=\bf] (impossible_regime) at (axis cs:6.5, 6) {
      \begin{tabular}{c}
        Impossible\\regime
      \end{tabular}
    };
    
\end{axis}

\end{tikzpicture}

%% file: proofoverview.tex
\section{Proof Overview}

\subsection{Variance Optimality}

To show Theorem~\ref{thm:quadraticopt}, we write the optimization problem
\begin{align*}
  m^2 \nvar_{p,q}(S) =\qquad &\min \Var_p[S_f]\\
  \text{s.t.}& \E_q[S_f] - \E_p[S_f] = m
\end{align*}
as a quadratic program in the vector $f = (f_0, \dotsc, f_n)$.  For
$\overline{p}_k = \Pr_p[Y_1 = k]$ and
$\overline{q}_k = \E_{i\in[m]}\Pr_q[Y_i = k]$, the constraint is that
$(\overline{q} - \overline{p}) \cdot f = 1$, and the objective is
$f^T Q f$ for some matrix $Q$.  The KKT condition \citep{Karu39, kuhn1951nonlinear, citeulike:163662} shows that the
optimum is achieved when $Qf = a(\overline{q} - \overline{p})$ for
some scalar $a$.

Solving this exactly requires the pseudoinverse $Q^+$, which would be tricky.  Instead,
we show that the quadratic statistic $f_k = k^2$ satisfies a slightly
different condition
\[
  Qf = a(q' - \overline{p}),
\]
for a different distribution $q' \in \R^{n+1}$ we can write
explicitly.  Therefore the quadratic statistic minimizes the variance
subject to an expectation gap in $q'$ relative to $\overline{p}$.
Moreover, this $q'$ turns out to be precisely the Taylor approximation
in $\eps$ to $\overline{q}$, with order $\eps^4$ error.  All that
remains is to show that this $O(\eps^4)$ distinction between $\overline{q}$ and
$q'$ gives $1 + o(1)$ loss in the program.  That is,
\[
  \abs{\E_{q'} f_k - \E_{\overline{p}} f_k}= (1 + o(1))\abs{\E_{\overline{q}} f_k - \E_{\overline{p}} f_k},
\]
or equivalently
\begin{align}\label{eq:qqp-close-2}
  \abs{\E_{q'} f_k - \E_{\overline{q}} f_k}  \ll \frac{1}{m}\abs{\E_q[S_f] - \E_p[S_f]},
\end{align}
for any statistic that we care about.
We can bound this LHS in terms of the
variance of $f$:
\[
  \abs{\E_{q'} f_k - \E_{\overline{q}} f_k} \lesssim \eps^4 \sqrt{\E_{\overline{p}}[f_k^2]}.
\]
Then we can relate the variance of $f$ to the variance of $S$:
\[
  \E_{\overline{p}}[f_k^2] \lesssim \frac{1}{m} \Var_p[S_f]
\]
using the fact that our statistic is indifferent to constant and
linear terms, so we can assume WLOG $\E[f_k] = \E[k f_k] = 0$.

Combining these results, we get that~\eqref{eq:qqp-close-2}
holds whenever
\[
  \nvar_{p,q}(S) = \frac{\Var_p[S_f]}{(\E_q[S_f] - \E_p[S_f])^2} \ll \frac{1}{\eps^8m}.
\]
Since the quadratic has
$\nvar_{p,q}(S) = \Theta(\frac{m}{\eps^4n^2})$, this holds for both
the quadratic and the statistic of maximal separation
$\nvar_{p,q}(S)$.  Therefore this maximum is within $1 + o(1)$ of the
quadratic.

\subsection{Concentration of Tails}
\paragraph{Setting.}  In this proof overview we will focus on the
Huber statistic in the regime where
$1 \lesssim \frac{n}{m} \ll \frac{1}{\eps^2}$,
as well as $\eps, \delta \ll 1$ (so $\frac{n^2}{m} \eps^4 \gg 1$).

Let $X_1, \dots, X_n$ be the $n$ samples drawn from distribution $\nu$ supported on $[m]$, and let $Y_j^n = \sum_{i=1}^n \1_{\{X_i = j\}}$ be the number of balls that end up in bin $j$.

\paragraph{The Huber statistic.} We consider the Huber statistic
\begin{equation}\label{eq:huber_definition}
  S = \sum_{j=1}^m h_\beta\left(Y_j^n - \frac{n}{m}\right)
\end{equation}
where
\begin{align}
\label{eq:huber_loss_def}
  h_\beta(x) := \left\{
      \begin{array}{cl}
        x^2 & \text{for } \abs{x} < \beta\\
        2\beta \abs{x} - \beta^2 & \text{otherwise}
      \end{array}
\right.
\end{align}
is the Huber loss function, which continuously interpolates between a
quadratic center and linear tails. Note that this is twice the
standard definition, but the statistic's performance is invariant
under affine transformations.


We will set $\beta$ large enough that most bins usually lie in the
quadratic regime in the uniform case.  If we were to set
$\beta=\infty$ (so $S$ is an affine transformation of the
collisions statistic), we would have
$\E_p[S] = n - n/m \approx n$ for the uniform distribution $p$
and
$\E_q[S] \geq n - n/m + 4n(n-1)\eps^2/m \approx n + 4n^2
\eps^2/m$ for any $\eps$-far distribution $q$.  This motivates us to
consider the rescaled statistic:
\begin{equation}
\Tilde{S} = \frac{m}{n^2\eps^2} \left[S - n \right]
\end{equation}
which (for $\beta = \infty$) has $\E[\Tilde{S}]$ being
$o(1)$ or $\geq 4 - o(1)$ in the uniform and far-from-uniform cases,
respectively.

Because $Y_j^n \sim B(n, 1/m)$ in the uniform case, Bernstein's inequality shows that setting
\begin{equation}
\label{eq:beta_second_bound}
\beta = \omega \left( \log\left(\frac{1}{\Delta}\right) + \sqrt{\frac{n}{m} \log\left(\frac{1}{\Delta}\right)} \right)
\end{equation}
gives that each bin lies in the quadratic regime with probability
$1-\Delta^2$, for a parameter $\Delta \ll 1$ that we will constrain
later.  Choosing this $\beta$ leads to smaller $\E[S]$ than
$\beta = \infty$, but the difference is only about
$\beta^2 \Delta^2 m$ because each of the $m$ bins has a $\Delta^2$
chance of lying in the linear region, and most of the differences
happen at the boundary where the Huber statistic is $\Theta(\beta^2)$.
This error is
$O(n \Delta^2 \log^2 \frac{1}{\Delta}) < O(n \Delta^{1.5})$, so
\[
  \E_p[\Tilde{S}] = o(1) + O(\frac{m}{n^2 \eps^2}\Delta^{1.5} n) = o(1)
\]
as long as we have
\begin{align}
  \label{eq:Delta_constraint}
  \Delta = O\left(\frac{n \eps^2}{m} \right)
\end{align}
which is $o(1)$.  Similarly, this implies
\[
  \E_q[\Tilde{S}] \geq 4 - o(1)
\]
for any $\eps$-far distribution $q$.

Finally, we will need some constraint that $\beta$ is not too
large/$\Delta$ too small.  A third moment condition suffices, as we
shall see in a few pages:
\begin{equation}
\label{eq:beta_third_bound}
(\beta^2 \eps^2)^3 = o\left(\Delta^2 \right)
\end{equation}
One can check that $\beta$ and $\Delta$ can be chosen such that the constraints~\eqref{eq:beta_second_bound},
\eqref{eq:Delta_constraint}, and \eqref{eq:beta_third_bound} hold in the
regime we consider here.

  
\paragraph{Analyzing the Huber statistic.} Our tester will pick a
threshold $\tau$, and ``accept'' the distribution as uniform if
$\Tilde{S} \le \tau$.  We therefore need to understand the false
negative probability
\[
  \delta_{-} := \Pr_p[\Tilde{S} \ge \tau]
\]
and similarly, for any $\eps$-far distribution $q$, we need
to bound the false positive probability
\[
  \delta_{+} := \Pr_q[\Tilde{S} \le \tau].
\]

To bound the maximum error $\delta = \max(\delta_{-}, \delta_{+})$, it
suffices to pick $\tau = 2$, halfway between the expectation bounds in
the uniform and $\eps$-far cases.

\paragraph{Completeness.} We start by describing how to analyze
$\delta_-$. The bulk of our analysis here is devoted to analyzing the
moment generating function
$M_{\Tilde{S}, \nu}(t) := \E_\nu[\exp(t \Tilde{S})]$.

A careful analysis (see, e.g., \citep{DGPP19} Lemma 3) shows that when
$\nu$ is the uniform distribution, the number of collisions has
variance $(1 + o(1)) \frac{n^2}{2m}$.  For large enough $\beta$
per~\eqref{eq:beta_second_bound}, this implies
\[
  \Var[\Tilde{S}] = (1 + o(1)) \frac{2m}{n^2 \eps^4}.
\]
Therefore we hope that $\Tilde{S}$ has MGF close to a Gaussian with
this variance.  In Lemma \ref{lem:huber_mgfs} we show that this is in
fact the case: for the uniform distribution $p$,
\begin{align}
\label{eq:huber_mgf_bound}
  M_{\Tilde{S}, p}\left(\frac{n^2 \eps^4}{m} \theta\right) & = \left(1 + O(1/n)\right) \exp\left\{\frac{n^2}{m} \eps^4\left( \theta^2 + o\left(1\right)\right) \right\}
\end{align}

Here, we pulled out $\frac{n^2 \eps^4}{m}$ from the MGF parameter, so
that we will set $\theta$ to be constant at the end. Once we have this, then standard
Chernoff-type arguments imply

\[
  \delta_{-} < \inf_{\theta \ge 0} \frac{M_{\Tilde{S}, p}\left(\frac{n^2 \eps^4}{m} \theta\right)}{e^{\frac{n^2 \eps^4}{m} \theta (E_p[\Tilde{S}] + \tau)}} < \inf_{\theta \ge 0}(1 + O(1/n)) \exp\left\{\frac{n^2}{m}\eps^4\left[ \theta^2 - \tau \theta + o\left(1\right)\right] \right\}
\]
and hence
\[
  \delta_{-} \le (1 + O(1/n)) \exp(-J_- (1 + o(1))\frac{n^2\eps^4}{m} )
\]
for ``error exponent''
\begin{equation}
\label{eq:J_minus_huber}
  J_- := \sup_{\theta \ge 0} \left\{- \frac{m}{n^2\eps^4} \log M_{\Tilde{S}, p}\left(\frac{n^2 \eps^4}{m}\theta\right) +  \theta \tau \right\} \ge \sup_{\theta \ge 0} \{\theta \tau - \theta^2\} = \frac{\tau^2}{4}
\end{equation}

The above is an upper bound on $\delta_-$, but we can also get a lower
bound.  Because the MGF bound~\eqref{eq:huber_mgf_bound} is tightly
that of a Gaussian, with both upper and lower bounds, we can apply the
G\"artner-Ellis theorem (see Appendix~\ref{sec:gartner}) to show that
the tail bound is tightly that of a Gaussian as well:
$\delta_{-} \gtrsim \exp(-(1+o(1))J n^2\eps^4/m)$.

\paragraph{Soundness.}
Because the Huber statistic $S$ is convex, we can apply existing tools
from \citep{DGPP18} to analyze the
statistic for uniformity testing. In particular, it is sufficient to
consider alternate distributions of the form $q$ such that
\begin{equation}
    q_j = 
    \begin{cases}
        1/m + \frac{\eps}{l}, & j \le l\\
        1/m - \frac{\eps}{m-l}, & j > l 
    \end{cases}
\end{equation}
for some $l \in [m]$.  Our discussion of this appears in
Appendix~\ref{app:worst_case_distributions}.  For simplicity of this exposition, suppose $m$ is
even and $l = m/2$. Using a similar procedure as in the case of the
uniform distribution, we show in Lemma \ref{lem:huber_mgfs} that for
this alternate distribution $q$,
\begin{align}
  M_{\Tilde{S}, q}\left(\frac{n^2 \eps^4}{m} \theta\right) = (1 + O(1/n)) \exp\left\{\frac{n^2 \eps^4}{m}\left[\theta^2 + 4 \theta + o(1)\right] \right\}.
\end{align}
That is to say, except for a mean shift of $4 + o(1)$, $\Tilde{S}$
under $q$ concentrates as a Gaussian with the same variance as it did
under $p$.  This gives us that $$\delta_+ \le (1 + O(1/n)) \exp\left(-J_+(1 + o(1)) \frac{n^2 \eps^4}{m} \right)$$

for ``error exponent''
$$J_+ := \sup_{\theta \ge 0} \left\{-\frac{m}{n^2 \eps^4} \log M_{\Tilde{S}, q} \left(- \frac{n^2 \eps^4}{m} \theta \right) - \theta \tau\right\} \ge \sup_{\theta \ge 0} \{- \tau \theta - \theta^2 + 4 \theta\} = \frac{(\tau - 4)^2}{4}$$

Setting $\tau = 2$ so that $J_- = J_+ = 1$ gives us that the error exponent achieved by the Huber tester is $1$ for the uniformity testing problem in this regime.

Alternatively, we could pick a different $\tau \in (0, 4)$ to trade
off $\delta_-$ and $\delta_+$, always getting within $(1+o(1))$ of the
tradeoff given by the Gaussian approximation to $S$.

\paragraph{Analyzing the MGF.} The key question, therefore, is how to
analyze the MGF.  For this, we follow the structure of
\cite{DBLP:journals/tit/HuangM13}, though with different
approximations because of our different regime.

We would like to analyze the MGF $M_{S_n}$ of our test statistic
\[
  S_n = h_\beta(\abs{Y_j^n - n/m}).
\]
If the $Y_j^n$ were independent over $j$, this would be easy: we would
simply bound the MGF of each individual term, and take the product.
For the same reason, it is easy to bound the MGF $A_\lambda(\theta)$
of the \emph{poissonized} test statistic $S_{\Poi(\lambda)}$, where
$\Poi(\lambda)$ balls are drawn rather than $n$.  We can get a Taylor
approximation to $A_\lambda$ that is quite accurate in our regime.

Unfortunately, we cannot just use the Poissonized MGF $A_n$ in place
of the true MGF $M_{S_n}$.  The problem is that Poissonization
inherently increases the variance: the variance of the collisions
statistic is $(1 + o(1))\frac{n^2}{2m}$ before Poissonization but
$(1 + o(1)) (\frac{n^2}{2m} + \frac{n^3}{m^2})$ after Poissonization.
For $n = \Theta(m)$ this is a constant factor we cannot afford to
lose, and for $n \gg m$ it's even worse.  So we need to
``depoissonize'' $A_\lambda$ into $M_{S_n}$.

To depoissonize, we observe that the Poissonized MGF $A_\lambda$ is a
mixture of the non-Poissonized MGFs $M_{S_k}$ for $k \geq 0$, and in
fact $M_{S_n}$ is just (up to scaling) the $\lambda^n$ coefficient in
the Taylor expansion of $A_\lambda$.  We then use Cauchy's theorem to
evaluate this coefficient.




\define{lem:MGF_f_bound}{Lemma}{%
  We have
  \begin{equation}
    \E\left[\exp\left(\eps^2 \theta h_\beta\left(Z_j - \frac{n}{m} \right)\right)\right] = \E\left[f(Z_j)\right] + o\left(\Delta^2\right)
  \end{equation}
  where $h_\beta$ is defined in \eqref{eq:huber_loss_def}, and
  $Z_j \sim \Poi(\lam \nu_j)$, for $\lam = n(1 + O(\eps^2))$ and
  $\nu_j = 1/m + O(\eps/m)$ for all $j$.
    
}

\paragraph{Comparison to Huang-Meyn}
Our proof structure is similar to \citep{DBLP:journals/tit/HuangM13}.
Differences arise from two causes:
first,~\citep{DBLP:journals/tit/HuangM13} consider the simpler
singletons tester $f(k) = 1_{k = 1}$, so the MGF of $f(Y_i)$ can be
written in closed form.  For the Huber statistic, we need to bound the
terms corresponding to the higher moments of the statistic, which is
done in Lemma \ref{lem:MGF_f_bound}.  Second, they use the asymptotic
regime $n/m \ll 1$ rather than $\eps \ll 1$ for their Taylor series
expansions to drop $o(1)$ terms, leading to a number of differences.


Finally, our proof for the alternate distributions is much simpler
than the proof in \citep{DBLP:journals/tit/HuangM13} since we make use
of results from \citep{DGPP18}.

 \define{lem:binomial_anticoncentration}{Lemma}{
   Let $X \sim B(n, 1/2)$, and $\frac{1}{\sqrt{n}} \ll \eps \ll 1$.  Then
   \[
     \Pr[X > \frac{n}{2} + \eps n] = \exp(-2 \eps^2 n (1 + o(1)).
   \]
 }


\subsection{Organization of the Appendix}

Appendix~\ref{section:optimality} shows Theorem~\ref{thm:quadraticopt},
that quadratic statistics have asymptotically optimal variance.  The
next sections show Theorem~\ref{thm:huber},
that the Huber statistic combines this variance with good
concentration, in the main new regime of
$1 \lesssim n/m \ll 1/\eps^2$: some background is given in
Appendix~\ref{section:preliminaries}, the main argument in
Appendix~\ref{section:huber}, and some technical computations are
deferred to Appendix~\ref{section:mgf_computation_lemmas}.

The rest of the appendix includes our analyses of other testers and
other regimes.  Proof of the asymptotically poor performance of the
collisions and singletons testers in some regimes is in
Appendix~\ref{appendix:lower_bounds}. The ``superlinear'' regime of
$n/m \gg 1/\eps^2$ is covered in Appendix~\ref{app:tv_superlinear}.
Analysis of the collisions/quadratic statistic is in
Appendix~\ref{section:squared}, while the TV/empty bins statistic for
$n < m$ is in Appendix~\ref{section:empty_bins}.

\section{Acknowledgments}
Shivam Gupta and Eric Price are supported by NSF awards CCF-2008868, CCF-1751040
(CAREER), and the NSF AI Institute for Foundations of Machine Learning (IFML).

%% file: variance-optimal.tex
\section{Variance Optimality (Theorem~\ref{thm:quadraticopt})}\label{section:optimality}

\paragraph{Setting.} Consider throwing $n$ balls into $m$ bins, for
$\lambda = n/m = O(1)$.  Suppose $m \gtrsim 1/\eps^4$ (as is needed
for constant success probability when $n \lesssim m$).  Let $k, k'$
be the number of balls landing in bins $1$ and $2$, respectively.  For
any $f$, let $\sigma^2 = \E_k[f_k^2]$.

\subsection{Optimality under a different distribution $q'$}
We define
\[
  \overline{p}_k := Bin(n, 1/m, k)
\]
to be the probability any given bin has $k$ balls in it under the
uniform distribution.

\begin{lemma}
For any alternative distribution $\overline{q}$, any statistic
$f$ minimizing the normalized variance
\[
  \frac{\Var_{p}[S_F]}{(\E_{k \sim \overline{q}}[f_k] - \E_{k \sim \overline{p}}[f_k])^2}
\]
satisfies
\[
  (Qf)_k = \alpha (\overline{p}_k - \overline{q}_k)
\]
for some $\alpha$ and all $k$.
\end{lemma}
\begin{proof}
  This is the KKT condition for minimizing the quadratic
  $\Var_{p}[S_f] = \frac{1}{m} f^T Q f$ subject to
  $\sum_k (\overline{p}_k - \overline{q}_k)f_k = 1$.
\end{proof}

Let
$\overline{q}_k := \frac{1}{2}Bin(n, (1+2\eps)/m, k) +
\frac{1}{2}Bin(n, (1-2\eps)/m, k)$.  We would like to show that a
quadratic is $1 - o(1)$-close to maximizing the normalized separation
between $\overline{p}$ and $\overline{q}$.

We have that
\begin{align*}
  \overline{p}_k &= \binom{n}{k} \frac{1}{m^k} (1 - 1/m)^{n-k}\\
  \overline{q}_k &= \binom{n}{k} \frac{1}{m^k} \frac{1}{2}((1+2\eps)^k(1 - (1+2\eps)/m)^{n-k} + (1-2\eps)^k(1 - (1-2\eps)/m)^{n-k})\\
                 &= \overline{p}_k \frac{(1+2\eps)^k(1 - (1+2\eps)/m)^{n-k} + (1-2\eps)^k(1 - (1-2\eps)/m)^{n-k}}{2(1 - 1/m)^{n-k}}\\
                 &= \overline{p}_k \frac{1}{2}((1+2\eps)^k(1 - \frac{2\eps}{m-1})^{n-k} + (1-2\eps)^k(1 + \frac{2\eps}{m-1})^{n-k})\\
\end{align*}

Now, for $\abs{a} \leq 2\eps$,
\[
  (1 + a) = e^{a - \frac{1}{2} a^2 + \frac{1}{3} a^3 + O(\eps^4)},
\]
and
\[
  (1 + \frac{a}{m-1})^{n-k} = e^{a\lambda + O(\abs{k-\lambda}a /m + a^2\lambda/m)} = e^{a \lambda + O(\eps^4)}
\]
for $k \lesssim 1/\eps$ and our setting of
$\lambda = O(1), m \gtrsim 1/\eps^4$.  Thus
\begin{align*}
  \overline{q}_k &=\overline{p}_k\frac{1}{2}( e^{2\eps k - 2\eps^2 k + \frac{8}{3}\eps^3 k + O(\eps^4 k)}e^{-2\eps \lambda + O(\eps^4)} + e^{-2\eps k - 2\eps^2 k - \frac{8}{3}\eps^3 k + O(\eps^4 k)}e^{2\eps \lambda + O(\eps^4)})\\
                 &= \overline{p}_ke^{-2\eps^2k + O(\eps^4 (k + 1))}\cosh(2\eps (k - \lambda) + \frac{8}{3} \eps^3 k)\\
                 &= \overline{p}_k (1 + 2\eps^2((k - \lambda)^2 - k) + O(\eps^4(k^2+ 1 + (k - \lambda)^4))
\end{align*}
as long as the final error term is $o(1)$
We now define
\begin{align}
  \label{eq:4}
  \alpha_k := (k - \lambda)^2 - k  + \lambda/m
\end{align}
so that
\begin{align}
  \label{eq:qalpha}
  \overline{q}_k = \overline{p}_k(1 + 2\eps^2 \alpha_k + O(\eps^4(k^4 + 1)))
\end{align}
under $k \lesssim \frac{1}{\eps}$ and our assumptions.  We make the
following simple observations:
\begin{lemma}\label{lem:alpha}
  \begin{align}
    \label{eq:5}
    \E_{k \sim \overline{p}}[k] = \lambda\\
    \E_{k \sim \overline{p}}[(k-\lambda)^2] = \lambda(1 - 1/m)\\
    \E_{k \sim \overline{p}}[\alpha_k] = 0.
  \end{align}
\end{lemma}
\begin{proof}
  The first two equations are just the mean and variance of a binomial
  random variable, and the third follows trivially.
\end{proof}

Define
\[
  q'_k := \overline{p}_k(1 + 2\eps^2 \alpha_k)
\]
which is also a probability distribution, since
$\E_{\overline{p}}[\alpha_k] = 0$ and $\alpha_k \geq -O(\lambda)$ so
it is positive.  For $q'$, the quadratic statistics are \emph{exactly}
optimal:
\begin{lemma}\label{lem:Qopt}
  Quadratic statistics $f_k = ak^2 + bk + c$ minimize
  \[
    \frac{\Var_{p}[S]}{(\E_{k \sim q'}[mf_k] - \E_{k \sim \overline{p}}[mf_k])^2}
  \]
  over all $f$, attaining value
  \begin{align}
     (1 + o(1))\frac{1}{8\eps^4 \lambda^2m}.
  \end{align}
\end{lemma}

\begin{proof}  \textbf{Value.} We first measure the value obtained by the quadratic statistic.  The quadratic statistic $f(k) = (k - \lambda)^2$
  has four times the variance of the collisions statistic
  $\binom{k}{2}$, so Lemma 3 of~\citep{DGPP19} shows that
  \[
    \Var_p[S] = 4 \binom{n}{2} (\frac{1}{m} - \frac{1}{m^2}) = (1 + o(1)) 2\lambda^2m .
  \]
  We also have, using the moments of a binomial, that
  \begin{align*}
    \E_{k \sim q'}[f_k] - \E_{k \sim \overline{p}}[f_k]
    &= \sum_k 2 \eps^2 \alpha_k \overline{p}_k f_k\\
    &= 2\eps^2 \E_{\overline{p}}[(k - \lambda)^4 - (k - \lambda)^3 - (1-1/m)\lambda(k-\lambda)^2]\\
    &= 2 \eps^2 (1-1/m)(\lambda(1 + \frac{3n-6}{m}(1 - 1/m)) - \lambda(1-2/m)-\lambda^2(1-1/m))\\
    &= 2 \eps^2 (1 + o(1))(\lambda + 3 \lambda^2 - \lambda - \lambda^2)\\
    &= 4 \eps^2 \lambda^2(1 + o(1))
  \end{align*}
  Hence
  \[
    \frac{\Var_p[S]}{(\E_{k \sim q'}[f_k] - \E_{k \sim \overline{p}}[f_k])^2} = (1 + o(1)) \frac{m}{8\eps^4 \lambda^2}.
  \]
  Scaling by $m^2$ gives the result.

  \paragraph{Optimality.}  We now show that it is optimal.  For any statistic $f$, we have that
  \[
    \Var[S_f] = m f^T Q F
  \]
  for a matrix $Q$ defined by
  \[
    Q_{k,k} = \overline{p}_k  + (m-1) \overline{p}_k \overline{p}_{k' \mid k} - m \overline{p}_k^2
  \]
  and
  \[
    Q_{k,k'} = (m-1) \overline{p}_k \overline{p}_{k'\mid k} - m \overline{p}_k \overline{p}_{k'},
  \]
  where $\overline{p}_{k'\mid k} = \Pr[Y_2 = k' \mid Y_1 = k]$.

  For any statistic $S = \sum f(Y_i)$, we have that
  \[
    (Qf)_k = \overline{p}_k (\E[S \mid Y_1 = k] - \E[S]).
  \]
  We also have that
  \[
    \E_{k \sim q'}[f_k] - \E_{k \sim \overline{p}}[f_k] = \sum_k 2 \eps^2 \alpha_k f_k.
  \]

  Therefore we can express the optimization as
  \begin{align}\label{eq:optimize}
    \begin{aligned}
      \min_{f} \quad & f^T Q f\\
      \textrm{s.t.} \quad & 2 \eps^2 \sum_k \alpha_k f_k = 1/m.
    \end{aligned}
  \end{align}
  The KKT condition for optimality is then that $Qf = a \alpha$ for some
  constant $a$.

  Now, the quadratic function $f_k = k^2$ satisfies
  \[
    \E[S] = \frac{n^2}{m} + n(1- \frac{1}{m}).
  \]
  Therefore
  \[
    \E[S \mid Y_1 = k] = k^2 + \frac{(n-k)^2}{m-1} + (n-k)(1- \frac{1}{m-1}).
  \]
  so
  \begin{align*}
    \E[S \mid Y_1 = k] - \E[S] &= \frac{m}{m-1}k^2 - k (\frac{2n}{m-1} + (1 - \frac{1}{m-1})) + (h_1(n, m))\\
                               &=\frac{m}{m-1}\alpha_k + h_2(n, m)
  \end{align*}
  for some functions $h_1, h_2$ of $n$ and $m$ but not $k$.  But since
  the LHS is zero in expectation over $k \sim\overline{p}$, and so is
  $\alpha_k$ by Lemma~\ref{lem:alpha}, we have $h_2 = 0$.  Thus:
  \[
    Qf = \frac{m}{m-1}\alpha.
  \]
  Hence the quadratic satisfies the KKT condition, so it
  optimizes~\eqref{eq:optimize} when scaled appropriately.
\end{proof}

We also note that the error in approximating  $\overline{q}$ by $q'$ has low moments:
\begin{lemma}\label{lem:qqp-moments} In our setting,
  \[
    \E_{\overline{p}}[(\frac{q'_k - \overline{q}_k}{\overline{p}_k})^2] \lesssim \eps^8.
  \]
\end{lemma}
\begin{proof}
  
  For $k \leq 1/\eps$, we have by~\eqref{eq:qalpha} that
  \[
    \abs{\frac{q'_k - \overline{q}_k}{\overline{p}_k}} = \abs{\alpha_k - \frac{\overline{q}_k}{\overline{p}_k}} \lesssim \eps^4 (k^4 + 1)
  \]
  such that
  \[
    \E_{\overline{p}}[(\frac{q'_k - \overline{q}_k}{\overline{p}_k})^2
    1_{k \leq 1/\eps}] \lesssim \E_{\overline{p}}[\eps^8 (k^8 +
    1)] \lesssim \eps^8.
  \]
  On the other hand, for $k > 1/\eps$,
  \[
    \E_{\overline{p}}[(\frac{q'_k - \overline{q}_k}{\overline{p}_k})^2 1_{k > 1/\eps}]
    \lesssim \E_{\overline{p}}[(\frac{\overline{q}_k}{\overline{p}_k})^2 1_{k > 1/\eps}].
  \]
  Now, for $k > 1/\eps$, the $\lambda(1 + \eps)$ part of
  $\overline{q}$ is more likely than the $\lambda(1-\eps)$ part.  Thus
  \[
    \frac{\overline{q}_k}{\overline{p}_k} \leq \frac{\binom{n}{k} ((1+\eps)/m)^k (1 - (1+\eps)/m)^{n-k}}{\binom{n}{k} (1/m)^k (1 - 1/m)^{n-k}} \leq (1 + \eps)^k
  \]
  while
  \[
    \overline{p}_k \leq (\frac{e\lambda}{k})^k = e^{O(k) - k \log k}
  \]
  so
  \[
    \E_{\overline{p}}[(\frac{q'_k - \overline{q}_k}{\overline{p}_k})^2 1_{k > 1/\eps}] \leq \sum_{k \geq 1/\eps} e^{2 \eps k}e^{O(k) - k \log k} \lesssim \eps^{\Omega(1/\eps)} < \eps^8
  \]
  giving the result.
\end{proof}

\subsection{Relating the covariance of one bin to the whole}

Recall that $\sigma^2 = \E[f_k^2]$, for $k \sim \overline{p}$.

\begin{lemma}\label{lem:high-cov}
  For any $B > 2\lambda$, we have
  \[
    \E[f_{k'}f_k 1_{k' > B}] \lesssim \sigma^2 \sqrt{\Pr[k' > B]}.
  \]
\end{lemma}
\begin{proof}
  For any $t > B$, we have
  \[
    \Pr[k  \mid k' = t] \lesssim \Pr[k]
  \]
  for all $k$.  This is trivially true for small $k \leq O(1)$ because
  $\Pr[k] = \Omega(1)$, and for large $k$---since $t$ is above
  average---$\Pr[k \mid k' = t] < \Pr[k]$.

  This implies
  \[
    \E[\abs{f_k}~\mid k' = t] \lesssim \E_k[\abs{f_k}] \leq \sigma.
  \]
  So
  \[
    \E[f_{k'}f_k 1_{k' > B}] = \sum_{k' > B} p_{k'} f_{k'} \E[f_k \mid k'] \lesssim \sigma \sum_{k' > B} p_{k'} \abs{f_{k'}}
  \]
  Of course, by Cauchy-Schwarz,
  \[
    \sum_{k' > B} p_{k'} \abs{f_{k'}} \leq \sqrt{(\sum_{k' > B} p_{k'})(\sum_{k' > B} p_{k'} f_{k'}^2)} \leq \sigma \sqrt{\Pr[k' > B]}
  \]
  and hence
  \[
    \E[f_{k'}f_k 1_{k' > B}] \lesssim \sigma^2 \sqrt{\Pr[k' > B]}.
  \]
\end{proof}

\begin{lemma}\label{lem:cov}
  Let $f_k$ satisfy $\E_p[f_k] = \E_p[k f_k] = 0$.  For sufficiently
  large $n, m$ we have
  \[
    \sigma^2 \lesssim \frac{1}{m} \Var[S_f].
  \]
\end{lemma}
\begin{proof}
  We can expand
  \[
    \Var[S_f] = m\sigma^2 + m(m-1) \E_{k,k'}[f_k f_{k'}].
  \]
  The lemma statement would be implied by
  \begin{align}
    \abs{\E_{k, k'}[f_k f_{k'}]} \leq \frac{1}{2(m-1)} \sigma^2,\label{eq:6}
  \end{align}
  where $k$ is the number of balls in bin $1$ and $k'$ is the number
  in bin $2$.  The probability that $k > B$ is at most
  \[
    2\binom{n}{B}\frac{1}{m^{B}} \leq 2 (\frac{e\lambda}{B})^{B} < \frac{1}{n^2m^4}
  \]
  for $B = O(\log m)$.  By Lemma~\ref{lem:high-cov},
  \[
    \abs{\E_{k, k'}[f_k f_{k'} 1_{k > B \cup k' > B}]} \leq 2\abs{\E_{k, k'}[f_k f_{k'} 1_{k > B}]} \lesssim \sqrt{\frac{1}{m^4}}\sigma^2.
  \]
  Therefore it would suffice to show
  \begin{align}\label{eq:7}
    \abs{\E_{k, k'}[f_k f_{k'} 1_{k < B \cap k' < B}]} \ll \frac{\sigma^2}{m}.
  \end{align}
  Let $\mathcal{B}$ be the event that $k < B \cap k' < B$.

  Let $\lambda' := (n - k')/(m-1) = \lambda(1 + \eps')$ for
  $\eps' = \frac{1}{m-1}\lambda(\lambda - k')$, which under
  $\mathcal{B}$ satisfies $\abs{\eps'} \lesssim m^{-2/3}$.  Then
  $(k \mid k')$ is $b(n-k', 1/(m-1))$, which is well approximated by
  $Poi(\lambda')$.  This Poisson approximation gives
  \begin{align*}
    p'_k &= \frac{(\lambda')^ke^{-\lambda'}}{k!} = p_k(1 + \eps')^k e^{-\lambda \eps'}\\
         &= p_k e^{(k - \lambda) \eps' + O((\eps')^2k)}\\
         &= p_k (1 + (k - \lambda) \eps' + O((k + (k-\lambda)^2)(\eps')^2))
  \end{align*}
  for $k \lesssim 1/\abs{\eps'}$, which holds given $\mathcal{B}$.
  Since $\E_p[f_k] = \E_p[k f_k] = 0$, we have that
  \begin{align*}
    \abs{\sum_{k} p_k(1 + (k - \lambda) \eps') f_k 1_{k \leq B}}
    &= \abs{\sum_{k} p_k(1 + (k - \lambda) \eps') f_k 1_{k > B}}\\
    &\leq \sqrt{(\sum_{k > B} p_k) \E_k (1 + (k - \lambda) \eps')^2 f_k^2}\\
    &\leq \sqrt{\Pr[k > B]} \sigma n \eps' \leq \frac{1}{m^2}\sigma.
  \end{align*}
  
  Therefore, for any $k' \leq B$,
  \begin{align*}
    \abs{\E_{k \mid k'}[f_k1_{k \leq B}]} &\leq \frac{\sigma}{m^2} + \abs{\E_{k} [O(k + (k - \lambda)^2)(\eps')^2 f_k 1_{k \leq B}]} \\
                                          &\lesssim \frac{\sigma}{m^2} +  (\eps')^2\sqrt{\E_{k} [(k + (k - \lambda)^2)^2]\E_{k}[f_k^2]}\\
                                          &\eqsim (\frac{1}{m} + (\eps')^2) \sigma\\
                                          &\lesssim \sigma/m^{4/3}.
  \end{align*}
  Therefore
  \[
    \abs{\E_{k',k}[f_{k'}f_k1_{k,k' \leq B}]} = \E_{k'}[\abs{f_{k'} 1_{k' \leq B}\E_{k \mid k'}[f_{k}1_{k \leq B}]}] \lesssim \E_{k'}[\abs{f_{k'}}1_{k' \leq B}\frac{\sigma}{m^{4/3}}] \lesssim \frac{\sigma^2}{m^{4/3}}
  \]
  which gives~\eqref{eq:7} as needed.
\end{proof}

\subsection{Putting it together}

\quadraticopt*
\begin{proof}
  Because the normalized separation is invariant to adding any linear
  function $ak + b$ to $f_k$, we can add use this degree of freedom to
  WLOG satisfy any two linear constraints.  We require that
  \[
    \E_{\overline{p}}[f_k] = 0
  \]
  and
  \[
    \E_{\overline{p}}[k f_k] = 0.
  \]

  Let $\wh{f}_k = k^2 + ak + b$ be the quadratic test statistic with
  $a$ and $b$ set to satisfy these two constraints..  By
  Lemma~\ref{lem:Qopt}, $\wh{f}$ is optimal under $q'$, so we have
  that
  \begin{align}\label{eq:OPT}
    OPT := (1 + o(1)) \frac{1}{8\eps^4 \lambda^2m} = \frac{\Var[S_{\wh{f}}]}{\E_{q'}[m\wh{f}_k]^2} \geq \frac{\Var[S_{f}]}{\E_{q'}[mf_k]^2}.
  \end{align}
  We have that
  \begin{align*}
    (\E_{q'}[f_k] - \E_{\overline{q}}[f_k])^2
    &= (\sum_k (q'_k - \overline{q}_k) f_k)^2\\
    &= (\E_{\overline{p}}[\frac{q'_k - \overline{q}_k}{\overline{p}_k} f_k])^2\\
    &\leq \E_{\overline{p}}[(\frac{q'_k - \overline{q}_k}{\overline{p}_k})^2] \E_{\overline{p}}[f_k^2]\\
    &\lesssim \eps^8 \sigma_f^2 &\text{(Lemma~\ref{lem:qqp-moments})}\\
    &\lesssim \eps^8 \frac{1}{m} \Var[S_f].&\text{(Lemma~\ref{lem:high-cov})}
  \end{align*}
  The same holds for $\wh{f}$, where we also have by~\eqref{eq:OPT} that
  \[
    \E_{q'}[\wh{f}_k]^2 = (1 + o(1)) \frac{1}{m} \Var[S_{\wh{f}}] \cdot 8\eps^4 \lambda^2
  \]
  so
  \[
    (\E_{q'}[\wh{f}_k] - \E_{\overline{q}}[\wh{f}_k])^2 \lesssim \eps^8
    \frac{1}{m} \Var[S_{\wh{f}}] = \eps^4 \cdot (1 +
    o(1))\frac{1}{8\lambda^2} \E_{q'}[\wh{f}_k]^2 \ll
    \E_{q'}[\wh{f}_k]^2
  \]
  and hence
  \[
    \E_{\overline{q}}[\wh{f}_k]^2 = (1 + o(1))  \E_{q'}[\wh{f}_k]^2
  \]
  so
  \[
    \nvar_{p, q}(S_{\wh{f}})  = \frac{\Var_p[S_{\wh{f}}]}{\E_{\overline{q}}[S_{\wh{f}}]^2} = (1 + o(1))OPT.
  \]

  For any alternative $f$, we split into two cases:

  \paragraph{Reasonably good $f$.}  When
  \[
    \frac{\Var[S_{f}]}{\E_{q'}[mf_k]^2} \leq 100 OPT,
  \]
  we again have
  \[
    \eps^8 \frac{1}{m} \Var[S_f] \ll \E_{q'}[\wh{f}_k]^2
  \]
  so
  \[
    \E_{\overline{q}}[f_k]^2 = (1 + o(1))  \E_{q'}[\wh{f}_k]^2
  \]
  and
  \[
    \nvar_{p, q}(S_f) = (1 + o(1))\frac{\Var[S_{f}]}{\E_{q'}[mf_k]^2} \geq (1 + o(1)) OPT.
  \]
  \paragraph{Bad $f$.} When
  \[
    \frac{\Var[S_{f}]}{\E_{q'}[mf_k]^2} \geq 100 OPT,
  \]
  we use $(a + b)^2 \leq 2a^2 + 2b^2$ to observe that
  \begin{align*}
    \nvar_{p, q}(S_f) &=  \frac{\Var[S_{f}]}{\E_{q}[mf_k]^2}
                        \geq \frac{1}{2}\frac{\Var[S_{f}]}{\E_{q'}[mf_k]^2 +m^2(\E_{q'}[f_k] - \E_{\overline{q}}[f_k])^2}\\
                      &= \frac{1}{2}\frac{\Var[S_{f}]}{\E_{q'}[mf_k]^2 + O(\eps^8m \Var[S_f])}\\
    &\geq \frac{1}{2} \frac{1}{\frac{1}{100 OPT} + \eps^8 m}\\
    &> OPT
  \end{align*}
  Thus, the quadratic tester achieves near-optimal separation for this
  $q$.

  Finally, for arbitrary distributions $q$ $\eps$-far from $p$ in TV,
  we note that the collisions tester satisfies
  \[
    \E_q[S] = \binom{n}{2} \norm{q}_2^2.
  \]
  By convexity the $\eps$-far $q$ minimizing this has its values above
  and below $1/m$ all equal; if there are $k$ values above $1/m$ this
  gives
  \begin{align*}
    \frac{1}{\binom{n}{2}}\E_q[S] &= k (\frac{1}{m} + \frac{\eps}{k})^2 + (m-k) (\frac{1}{m} - \frac{\eps}{m-k})^2\\
    &= \frac{1}{m} + \eps^2 \left(\frac{1}{k} + \frac{1}{m-k}\right)
  \end{align*}
  which is minimized at $k = m-k = m/2$, precisely the $q$ considered
  above.
\end{proof}

%% file: preliminaries.tex
\section{Background for Tester Analysis}\label{section:preliminaries}
\subsection{G\"artner-Ellis Theorem}
\input{gartner_ellis}

\input{worst_distribution}

%% file: gartner_ellis.tex
\label{sec:gartner}
The statements in this section are taken from \cite{dembo1998large}.

Consider a sequence of random variables $Z_n \sim p_n$ and let the logarithmic moment generating function of $Z_n$ be $$\Lambda_n(\theta) := \log \E[e^{\theta Z_n}]$$
\begin{assumption}\label{assumption:gartner_ellis}
Suppose that for each $\theta \in \R$, the logarithmic moment generating function, defined as the limit $$\Lambda(\theta) =  \lim_{n \to \infty}\frac{1}{n} \Lambda_n(n \theta)$$

exists as an extended real number, and that the origin lies in the interior of the set $\mathcal{D}_\Lambda := \{\theta \in \R : \Lambda(\theta) < \infty\}$.
\end{assumption}

Let 
\begin{equation}
\label{eq:fenchel_legendre}
    \Lambda^*(\tau) = \sup_{\theta \ge 0} \{\theta \tau - \Lambda(\theta)\}
\end{equation} be the Fenchel-Legendre transform of $\Lambda$. 

\begin{definition}
$\tau \in \R$ is an exposed point of $\Lambda^*$ if for some $\lambda \in \R$, for every $x \neq y$, $$\lam \tau - \Lambda^*(\tau) > \lam x - \Lambda^*(x)$$

Then, $\lam$ is called an exposing hyperplane.
\end{definition}

\begin{theorem}[G{\"a}rtner--Ellis]
\label{thm:gartner_ellis}
Let Assumption \ref{assumption:gartner_ellis} hold.
\begin{enumerate}[(a)]
\item 
  For any closed set $F$, $$\limsup_{n \to \infty} \frac{1}{n} \log p_n(F) \le - \inf_{x \in F}\Lambda^*(x)$$
\item For any open set $G$, $$\liminf_{n \to \infty} \frac{1}{n} \log p_n(G) \ge - \inf_{x \in G \cap F} \Lambda^*(x)$$
  where $F$ is the set of exposed points of $\Lambda^*$ whose exposing hyperplane belongs to $D^o_\Lambda$, where $D^{o}_\Lambda$ is the interior of $D_\Lambda$.
\end{enumerate}
\end{theorem}

%% file: worst_distribution.tex
\subsection{Worst Case Distributions for Uniformity Testing}\label{app:worst_case_distributions}

In this section, we study the worst-case $\eps$-far distributions for
test statistics that are convex symmetric functions of the histogram
(i.e., the number of times each domain element is sampled) of an
arbitrary random variable $Y$.  This is an extension of the results
in~\citep{DGPP18}, which we recap below.

\paragraph{Prior work.}

We start with the following definition:
\begin{definition}\label{def:majorization}
Let $p=(p_1,\dots,p_n),q=(q_1,\dots,q_n)$ be probability distributions
and $p^{\downarrow},q^{\downarrow}$ denote the vectors with the same values as $p$ and $q$ respectively, 
but sorted in non-increasing order. We say that $p$ \emph{majorizes} $q$ (denoted by $p\succ q$) if 
\begin{equation}\label{eq:condition}
\forall k: \sum_{i=1}^k p^{\downarrow}_i \geq \sum_{i=1}^k q^{\downarrow}_i \;.
\end{equation}
\end{definition} 

A proof of the following simple fact can be found in~\citep{DGPP18}:

\begin{fact}\label{lem:averaging-implies-majorization}
Let $p$ be a probability distribution over $[n]$ and $S\subseteq [n]$. 
Let $q$ be the distribution which is identical to $p$ on $[n]\setminus S$, 
and for every $i \in S$ we have $q_i=\frac{{p}(S)}{\vert S\vert}$, 
where $\vert S\vert$ denotes the cardinality of $S$. 
Then, we have that $p\succ q$.    
\end{fact}

We also use the following standard terminology: we say that a real
random variable $A$ \emph{stochastically dominates} a real random
variable $B$ if for all $x \in \R$ it holds
$\Pr[A > x] \geq \Pr[B > x]$.

We say that a test statistic $S$ is ``convex symmetric'' if it is a
convex function of the histogram $(Y_1, \dots, Y_m)$ and invariant
under permutation of the $Y_i$.  A ``test'' is given by a test
statistic $S$ and threshold $\tau$, and outputs ``uniform'' if
$S \leq \tau$ and ``non-uniform'' otherwise.

It was shown in~\citep{DGPP18} that if $p$ majorizes $q$, then a
convex symmetric test statistic of $p$ stochastically dominates one
from $q$:

\begin{lemma}[Lemma 19 of~\citep{DGPP18}]\label{lm:general domination}
  Let $f: \R^n \to \R$ be a symmetric convex function and $p$ be a
  distribution over $[n]$.  Suppose that we draw $m$ samples from $p$,
  and let $X_i$ denote the number of times we sample element $i$.  Let
  $g(p)$ be the random variable $f(X_1,X_2,\ldots,X_n)$.  Then, for
  any distribution $q$ over $[n]$ such that $p\succ q$, we have that
  $g(p)$ stochastically dominates $g(q)$.
\end{lemma}


\paragraph{New claims.}  We will show that it suffices to consider
distributions that are ``flat'', meaning that $p_i$ takes only two
values:

\begin{definition}
  We say a probability distribution $p$ over $[n]$ is an \emph{$\gamma$-skewed
    flat distribution} if it takes the form:
  \[
    p_i = \left\{
      \begin{array}{cl}
        a & i \in T\\
        b & i \notin T
      \end{array}
    \right.
  \]
  for some reals $a, b$ and set $T \subseteq [n]$ with
  $\abs{T} \in [\gamma n, (1-\gamma)n]$.
\end{definition}

We make the following generalization of Lemma~21 in~\citep{DGPP18}
(which is the $\gamma=1/2$ case):
 
\begin{lemma}\label{lem:averaging-preserves-tv}
  Let $p$ be a probability distribution.  For any $0 < \gamma < 1/2$,
  there exists an $\gamma$-skewed flat distribution $p'$ such that
  $p \succ p'$ and
  \[
    (1-\gamma)\cdot \norm{p-U_n}_{TV}\leq \norm{p^\prime-U_n}_{TV} \leq \norm{p-U_n}_{TV}.
  \] 
\end{lemma}
\begin{proof}
  Let $T = \{i : p_i > 1/n\}$, so
  \begin{align}
    \norm{p-U_n}_{TV} = \sum_{i \in T} (p_i - 1/n) = \sum_{i \in [n] \setminus T} (1/n - p_i).
  \end{align}
  If $\abs{T} \in [\gamma n, (1-\gamma)n]$, we can simply choose $p'$
  to be $p$ averaged over $T$, and $p$ averaged over
  $[n]\setminus T$---this is $\gamma$-skewed and flat, has
  $p \succ p'$ by Lemma~\ref{lem:averaging-implies-majorization}, and
  has $\norm{p' - U_n}_{TV} = \norm{p - U_n}_{TV}$.  The only
  remaining cases have $\abs{T} \notin [\gamma n, (1-\gamma)n]$, since
  this approach would not be $\gamma$-skewed.

  Let $T' \subset [n]$ contain the largest either $\gamma n$ or
  $(1-\gamma)n$ coordinates of $p$, depending on whether
  $\abs{T} < \gamma n$ or $\abs{T} > (1-\gamma)n$, and let $p'$
  average $p$ over $T'$ and over $[n]\setminus T$.  This is
  $\gamma$-skewed and flat, and has $p \succ p'$ by
  Lemma~\ref{lem:averaging-implies-majorization}, so the only question
  is the TV bound.

  We have that
  \[
    \norm{p' - U_n}_{TV} = \sum_{i \in T'} (p'_i - 1/n) = \sum_{i \in T'} (p_i - 1/n)
    = \norm{p - U_n}_{TV} - \sum_{i \in T \setminus T'} (p_i - 1/n) - \sum_{i \in T' \setminus T} (1/n - p_i).
  \]
  Every term in the right two sums is nonnegative, so
  $\norm{p' - U_n}_{TV} \leq \norm{p - U_n}_{TV}$.

  Now, if $\abs{T} < \gamma n$, then $T \setminus T'$ is empty and,
  since $T'$ takes the largest coordinates in $p$, $p_i$ is larger on
  average for $i \in T' \setminus T$ than for $i \in [n] \setminus T$:
  \[
    \sum_{i \in T' \setminus T} (1/n - p_i) \leq \frac{\abs{T'\setminus T}}{\abs{[n] \setminus T}}\sum_{i \in [n] \setminus T} (1/n - p_i) = \frac{\gamma n - \abs{T}}{n - \abs{T}}\norm{p - U_n}_{TV} \leq \gamma \norm{p - U_n}_{TV}
  \]
  so
  \[
    \norm{p' - U_n}_{TV} \geq (1-\gamma)\norm{p - U_n}_{TV}.
  \]
  Similarly, if $\abs{T} > (1-\gamma) n$, then $T' \setminus T$ is empty and
  \[
    \sum_{i \in T \setminus T'} (p_i - 1/n) \leq \frac{\abs{T \setminus T'}}{\abs{T}}\sum_{i \in T} (p_i - 1/n) = \frac{\abs{T} - (1-\gamma)n}{\abs{T}} \norm{p - U_n}_{TV} \leq \gamma \norm{p-U_n}_{TV},
  \]
  again giving 
  \[
    \norm{p' - U_n}_{TV} \geq (1-\gamma)\norm{p - U_n}_{TV}
  \]
  as desired.
\end{proof}

The above results mean that it suffices to prove that our algorithm
can distinguish the uniform distribution from $\gamma$-skewed flat
distributions.  The inefficiency from not considering extremely skewed
distributions is only $1 + O(\gamma)$:

\begin{lemma}\label{lem:gamma-error}
  Suppose a convex symmetric test statistic $S$ and threshold has the
  property that, when applied to any $\eps$-far $\gamma$-skewed flat
  distribution $p$, the false negative rate is at most $\delta$.  Then
  the same statistic and threshold, when applied to any
  $\frac{1}{1-\gamma}\eps$-far distribution $p'$, also has false
  negative rate at most $\delta$.
\end{lemma}
\begin{proof}
  For any such $p'$, Lemma~\ref{lem:averaging-preserves-tv} states
  that there exists a $p$ that is $\gamma$-skewed, $\eps$-far from
  $U_n$ in TV, and with $p' \succ p$.  Lemma~\ref{lm:general
    domination} then states that $S$ on $p'$ stochastically dominates
  $S$ on $p$, so the chance of falling below the threshold is smaller
  for $p'$ than for $p$---and the latter is $\delta$ by assumption.
\end{proof}

\paragraph{Implication for Error Exponents.}
Let $\eps = \eps(n)$, $m = m(n)$, and $\tau = \tau(n)$ be functions of
$n$.  Let $p$ be uniform on $[m]$.  The false positive error exponent
$c_{+} = c_{+}(\eps, m)$ of a test $(S, \tau)$ is
\[
  c_{+} = \lim_{n \to \infty} -\frac{m}{n^2 \eps^4} \log \Pr_p[S > \tau].
\]
For a \emph{particular} family of distributions $q$, the false
negative error exponent $c_{-}^{(q)} = c_{-}^{(q)}(\eps, m)$ is
\[
  c_{-}^{(q)} = \lim_{n \to \infty} -\frac{m}{n^2 \eps^4} \log \Pr_q[S \leq \tau].
\]
The false negative error exponent $c_{-}$ is the worst such exponent
over all $\eps$-far distributions $q$:
\[
  c_{-} = \inf_{q: \norm{p - q}_{TV} \geq \eps} c_{-}^{(q)}.
\]
Varying $\tau$ allows for a tradeoff between false negatives and false
positives.  Balancing the two gives us the \emph{error exponent}
$c = c(\eps, m)$ for a test statistic $S$:
\[
  c = \sup_{\tau} \min(c_{+}, c_{-}).
\]
If a test statistic has error exponent $c$, it can distinguish the
uniform distribution from any non-uniform distribution with
probability $1 - \exp(-(1 + o(1))c \eps^4 n^2/m)$.  Equivalently, it
gets error probability $\delta$ where
\[
  n = \frac{1 + o(1)}{\sqrt{c}} \cdot \frac{\sqrt{m \log \frac{1}{\delta}}}{\eps^2}.
\]



We define $\overline{c} = \overline{c}(\eps, m, \gamma)$ to denote an alternative to
$c$ where we only consider $\eps$-far distributions
$q$ that are $\gamma$-skewed and flat.

\begin{lemma}\label{lem:Csandwich}
  For any functions $\eps, m, \gamma$,
  \[
    (1 - 4\gamma) \cdot \overline{c}((1 - \gamma)\eps, m, \gamma) \leq c(\eps, m) \leq \overline{c}(\eps, m, \gamma).
  \]
\end{lemma}
\begin{proof}
  The upper bound on $c$ is trivial: as an infimum over a larger set
  of $q$, $c_{-} \leq \overline{c}_{-}$, so $c \leq \overline{c}$.

  For the lower bound on $c$, we note by Lemma~\ref{lem:gamma-error}
  that for any $q$ with $\norm{p - q}_{TV} \geq \eps$ that there
  exists a $\gamma$-skewed flat distribution $q'$ with $\norm{p - q'}_{TV} \geq (1-\gamma)\eps$ such that
  \begin{align}
    \Pr_q[S \leq \tau] \leq  \Pr_{q'}[S \leq \tau].\label{eq:qcomparison}
  \end{align}
  This implies that
  \begin{align*}
    \frac{m}{n^2 \eps^4}\log \Pr_q[S \leq \tau] \leq  (1 - \gamma)^4 \frac{m}{n^2 (1-\gamma)^4\eps^4}\log \Pr_{q'}[S \leq \tau].
  \end{align*}
  so
  \[
    c_{-}^{(q)}(\eps, m) \geq (1 - \gamma)^4 c_{-}^{(q')}((1-\gamma)\eps, m),
  \]
  and hence $c \geq (1 - 4 \gamma)\overline{c}((1 - \gamma)\eps, m)$.
\end{proof}

\begin{lemma}
\label{lem:final_worst_case_lemma}
  Let $S$ by a convex symmetric test statistic.  Consider any family
  of parameters $(n, \eps, m)$.  Suppose that there exists a constant
  $\gamma'$ such that, for any $\gamma = \Omega(1) > \gamma'$ and $\eps'$ that uniformly
  satisfies $(1 - \gamma') \eps(n) \leq \eps'(n) \leq \eps(n)$,
  \[
    \overline{c}(\eps', m, \gamma) = c^*
  \]
  for a fixed value $c^*$ [that depends on the family $(n, \eps, m)$
  but not on the value of $n$ or $\gamma, \gamma'$].

  Then
  \[
    c(\eps, m) = c^*.
  \]
\end{lemma}

\begin{proof}
  By Lemma~\ref{lem:Csandwich},
  \[
    c(\eps, m) \leq \overline{c}(\eps, m, \gamma) = c^*.
  \]
  Moreover, for any $C$ we have
  \[
    c(\eps, m) \geq (1 - 4\gamma)c^*.
  \]
  where $c$ is a limit as $n \to \infty$ independent of $C$.  But this
  means that $c(\eps, m) = c^*$, because it is larger than any fixed
  value less than $c^*$.
\end{proof}

%% file: huber_statistic.tex
\section{Huber Statistic in Sublinear Regime}\label{section:huber}
\subsection{Regime}
The Huber statistic is given by 
\begin{equation}
  S = \sum_{j=1}^m h_\beta\left(Y_j^n - \frac{n}{m}\right)\tag{\ref{eq:huber_definition}}
\end{equation}
where
\begin{align}
  h_\beta(x) := \left\{
      \begin{array}{cl}
        x^2 & \text{for } \abs{x} < \beta\\
        2\beta \abs{x} - \beta^2 & \text{otherwise}
      \end{array}
\right. \tag{\ref{eq:huber_loss_def}}
\end{align}
is the Huber loss function. Here $Y_j^n = \sum_{i=1}^n \1_{\{X_i = j\}}$ and $X_1, \dots, X_n$ are the $n$ samples drawn from distribution $\nu$ supported on $[m]$.

\begin{assumption}\label{assumption:huber}
 $n = \Omega(m), n/m \ll \frac{1}{\eps^2}$, $\eps \ll 1$, $\frac{n^2}{m} \eps^4 \gg 1$, and $m \ge C \log n$ for sufficiently large constant $C$. In addition, we have the following constraints on $\beta$, the Huber parameter, and $\Delta$.
\begin{equation}
\beta = \omega\left( \log\left(\frac{1}{\Delta}\right) + \sqrt{\frac{n}{m} \log\left(\frac{1}{\Delta}\right)} \right) \tag{\ref{eq:beta_second_bound}}
\end{equation}
\begin{align}
  \Delta = O\left(\frac{n \eps^2}{m} \right)\tag{\ref{eq:Delta_constraint}}
\end{align}
\begin{equation}
(\beta^2 \eps^2)^3 = o\left(\Delta^2 \right)\tag{\ref{eq:beta_third_bound}}
\end{equation}
 
\end{assumption}
We will assume that Assumption~\ref{assumption:huber} holds throughout this section.


%
%
%

Note that since $\Delta = o(1)$, \eqref{eq:beta_third_bound} implies that 
\begin{equation}
\label{eq:beta_eps^2_little_o_1}
\beta^2 \eps^2 = o(1)
\end{equation}





%


Our goal is to compute an upper bound on the asymptotic expansion of the cumulant generating function (also called the logarithmic moment generating function) of this statistic.

For ease, instead of analyzing $S$ directly, we will analyze the statistic 
\begin{equation}
\label{eq:huber_Tilde_S_definition}
\Tilde{S} = \frac{m}{n^2 \eps^2 }\left[S - n\right]
\end{equation}
Note that this has the same error probability as $S$ since it simply applies a translation and scaling to $S$.

Consider the moment generating function (MGF) of $\Tilde{S}$ with respect to distribution $\nu$, given by $$M_{\Tilde{S}, \nu}(\theta) = \E_\nu \left[\exp(\theta \Tilde{S})\right]$$
The logarithmic moment generating function of $\Tilde{S}$ with respect to distribution $\nu$ is given by 
\begin{equation}
\label{eq:huber_log_MGF_definition}
\Lambda_{n, \nu}(\theta) := \log\left(M_{\Tilde{S}, \nu}(\theta) \right)
\end{equation}
We will compute an asymptotic expansion of the limiting logarithmic moment generating function of $\Tilde{S}$, given by $$\Lambda_\nu(\theta) = \lim_{n \to \infty}\frac{m}{n^2 \eps^4} \Lambda_{n, \nu}\left(\frac{n^2 \eps^4}{m} \theta\right)$$
For ease of exposition, we define a centering function 
\begin{equation}
\label{eq:centering_def}
\phi(k):= \left|k - \frac{n}{m} \right|
\end{equation}

\subsection{Poissonization}

Define $\Tilde{S}_{Poi(\lam)}$ to be the Poissonized statistic, that is the statistic $\Tilde{S}$ when the number of balls is chosen according to the Poisson distribution with mean $\lam$.

We begin by computing the MGF of $\Tilde{S}_{Poi(\lam)}$ with MGF parameter $\frac{n^2 \eps^4}{m} \theta$. That is, let

\begin{equation}
\label{eq:huber_A_lam(theta)}
A_\lam(\theta) := \E\left[\exp\left(\frac{n^2 \eps^4}{m} \theta \Tilde{S}_{Poi(\lam)}\right) \right] = \exp(- \eps^2 \theta n) \E\left[\exp\left(\eps^2 \theta \sum_{j=1}^m h_\beta\left(Z_j - \frac{n}{m} \right) \right)\right]
\end{equation}

where $Z_j \sim Poi(\lam \nu_j)$ and are independent. Due to this independence,
\begin{equation*}
A_\lam(\theta) = \exp(-\eps^2 \theta n) \prod_{j=1}^m \E\left[\exp\left(\eps^2 \theta h_\beta \left(Z_j - \frac{n}{m} \right)\right) \right]
\end{equation*}

Define 
\begin{equation}
    \label{eq:huber_f_def}
    f(k) := 1 + \eps^2 \theta \phi(k)^2 + \frac{\eps^4 \theta^2}{2} \phi(k)^4
\end{equation}

We will first show the following

\begin{lemma}
\label{lem:satisfying_huber_assumptions}
$$\eps^2 \theta \E\left[ h_\beta \left(Z_j - \frac{n}{m} \right)\right] = \eps^2 \theta \E[\phi(Z_j)^2] + o(\Delta^2)$$
$$\frac{\eps^4 \theta^2}{2} \E\left[ h_\beta \left(Z_j - \frac{n}{m} \right)^2\right] = \frac{\eps^4 \theta^2}{2} \E[\phi(Z_j)^4] + o(\Delta^2)$$
$$\sum_{l=3}^\infty \frac{(\eps^2 \theta)^l}{l!} \E\left[h_\beta\left(Z_j - \frac{n}{m}\right)^l\right] = o(\Delta^2)$$
where $h_\beta$ is defined in \eqref{eq:huber_loss_def}, and $Z_j \sim Poi(\lam \nu_j)$, for $\lam = n(1 + O(\eps^2))$ and $\nu_j = 1/m + O(\eps/m)$ for all $j$.
\end{lemma}
\begin{proof}
$$\eps^2 \theta \E\left[h_\beta\left(Z_j - \frac{n}{m}\right) \right] = \eps^2 \theta\left\{ \E\left[\1_{\{\phi(Z_j) \le \beta\}} \phi(Z_j)^2\right] + \E\left[\1_{\{\phi(Z_j) > \beta\}} \beta (2 \phi(Z_j) - \beta) \right]\right\}$$
$$ = \eps^2 \theta\ E[\phi(Z_j)^2] - \eps^2 \theta \E[\1_{\{\phi(Z_j) > \beta\}} \phi(Z_j)^2] + \eps^2 \theta \E[\1_{\{\phi(Z_j) > \beta\}} \beta (2 \phi(Z_j) - \beta)]$$
By Lemma~\ref{lem:huber_tail_moment_bound}, the second term is $o(\Delta^2)$. For the third term,  $$\eps^2 \theta \E[\1_{\{\phi(Z_j) > \beta\}} \beta(2 \phi(Z_j) - \beta)] \le \E[\1_{\{\phi(Z_j) > \beta\}} \exp(\eps^2 |\theta| \beta(2\phi(Z_j) - \beta))]$$
By Lemma~\ref{lem:huber_linear_terms_bound}, this is $o(\Delta)^2$. So, we have the first claim. The second claim can be proved in a similar way. For the third claim,
\begin{align*}
\sum_{l=3}^\infty \frac{(\eps^2 \theta)^l}{l!} \E\left[h_\beta\left(Z_j - \frac{n}{m}\right)^l\right] = \sum_{l=3}^\infty \frac{(\eps^2\theta)^l}{l!} \E\left[\1_{\{\phi(Z_j) \le \beta\}} \phi(Z_j)^{2l} \right]\\
+ \sum_{l=3}^\infty \frac{(\eps^2 \theta)^l}{l!} \E\left[\1_{\{\phi(Z_j) > \beta\}} (\beta(2 \phi(Z_j) - \beta))^l \right]
\end{align*}
By Lemma~\ref{lem:quadratic_terms_bound}, the first term is $o(\Delta^2)$. For the second term, in a similar fashion as before,
$$\sum_{l=3}^\infty \frac{(\eps^2 \theta)^l}{l!} \E\left[\1_{\{\phi(Z_j) > \beta\}} (\beta(2 \phi(Z_j) - \beta))^l\right] \le \E\left[\1_{\{\phi(Z_j) > \beta\}} \exp(\eps^2 |\theta| \beta(2 \phi(Z_j) - \beta)) \right]$$
By Lemma~\ref{lem:huber_linear_terms_bound}, this is $o(\Delta^2)$.
\end{proof}

\state{lem:MGF_f_bound}
\begin{proof}
Follows from Lemma \ref{lem:satisfying_huber_assumptions}.
\end{proof}

\subsection{Depoissonization}\label{huber_depoissonization}

First, we will show that $A_\lam(\theta)$ is analytic in $\lam$.
\begin{lemma}
\label{lem:huber_analytic}
$A_\lam(\theta)$ is analytic in $\lam$.
\end{lemma}
\begin{proof}
We will show that $\E[\exp\left(\eps^2 \theta h_\beta \left(Z_j - \frac{n}{m} \right) \right)]$ can be written as a finite sum of analytic functions in $\lambda$. Since the sum and product of analytic functions also analytic, this will show that $A_\lam(\theta)$ is analytic. Let

\begin{equation*}
A := \sum_{k : \phi(k) \le \beta} \left[\frac{(\lam \nu_j)^k}{k!} e^{-\lam \nu_j} \exp\left(\eps^2 \theta \phi(k)^2\right)\right]\\
\end{equation*}

$$B := \E\left[\exp\left\{\eps^2 \theta \beta\left(2 \left(Z_j - \frac{n}{m}\right) - \beta\right) \right\}\right] $$

$$C := \sum_{k : k < \frac{n}{m} - \beta} \left[ \frac{(\lam \nu_j)^k}{k!} e^{- \lam \nu_j} \left[\exp\left\{\eps^2 \theta \beta \left(2 \left(\frac{n}{m} - k\right) - \beta\right)\right\} - \exp\left\{\eps^2 \theta \beta \left(2 \left(k - \frac{n}{m}\right) - \beta\right)\right\}\right]\right]$$

Then, the expectations in $B$ and $C$ can be expressed in terms of the moments and MGF of the Poisson distribution, and so, are analytic. Each of $A, B, C$ is a finite sum of analytic functions, and so, is analytic. It is easy to verify that $$\E\left[\exp\left(\eps^2 \theta h_\beta \left(Z_j - \frac{n}{m} \right) \right)\right] = A + B + C$$
is thus analytic.

\end{proof}

Now, we want depoissonize $A_\lam(\theta)$. For ease of exposition, we will prove a more general result. First we assume the following. 
\begin{assumption}
\label{assumption:xi}
Suppose $\xi$ is a function such that $$A_\lam(\theta) = \exp(-\eps^2 \theta n) \prod_{j=1}^m \E[\exp(\eps^2 \theta \xi(Z_j))]$$ where $A_\lam(\theta)$ is analytic in $\lam$, and $Z_j \sim Poi(\lam \nu_j)$.

We assume that, for $\lam = n(1 + O(\eps^2))$ and $\nu_j = 1/m + O(\eps/m)$ for all $j$, we have
\begin{equation}
\label{eq:xi_first_moment_assumption}
\eps^2 \theta \E[\xi(Z_j)] = \eps^2 \theta \E[\phi(Z_j)^2] + o(\Delta^2)
\end{equation}

\begin{equation}
\label{eq:xi_second_moment_assumption}
\frac{\eps^4 \theta^2}{2} \E[\xi(Z_j)^2] = \frac{\eps^4 \theta^2}{2} \E[\phi(Z_j)^4] + o(\Delta^2)
\end{equation}

\begin{equation}
\label{eq:xi_higher_moment_assumption}
\sum_{l=3}^\infty \frac{(\eps^2 \theta)^l}{l!} \E[\xi(Z_j)^l] = o(\Delta^2)
\end{equation}
so that
$$\E[\exp(\eps^2 \theta \xi(Z_j))] = \E[f(Z_j)] + o\left(\Delta^2 \right)$$

 where $f$ is defined in \eqref{eq:huber_f_def}.

Let $Y_j^n = \sum_{i=1}^n \1_{\{X_i = j\}}$ and $X_1, \dots, X_n$ be $n$ samples drawn from distribution $\nu$ supported on $[m]$.
\end{assumption}
We will show the following:

\begin{lemma}
\label{lem:main_generic_mgf_lemma}
Suppose Assumption \ref{assumption:xi} holds.
Then, if $\nu$ is the uniform distribution such that $\nu_j = 1/m$ for all $j$, we have  

$$\exp(-\eps^2 \theta n) \E\left[\exp\left(\eps^2 \theta \sum_{j=1}^m \xi(Y_j^n)\right)\right] = (1 + O(1/n)) \exp\left\{\frac{n^2 \eps^4}{m} (\theta^2 + o(1)) \right\}$$

If $\nu$ is an alternate distribution such that $\nu_j = \frac{1}{m} + \frac{\eps}{\gamma m}$ for $j \le \gamma m$, and $\nu_j = \frac{1}{m} - \frac{\eps}{(1-\gamma) m}$ for $j > \gamma m$, for $\gamma = \Theta(1), 1-\gamma = \Theta(1)$, we have

\begin{dmath*}
\exp(-\eps^2 \theta n) \E\left[\exp\left(\eps^2 \theta \sum_{j=1}^m \xi(Y_j^n)\right)\right] = (1 + O(1/n)) \exp\left\{\frac{n^2 \eps^4}{m} \left( \theta^2 + \theta \frac{1}{\gamma(1-\gamma)} + o(1)\right) \right\}
\end{dmath*}

\end{lemma}


First, we have that our expression stated can be written as an integral.
\begin{lemma}
\label{lem:general_depoissonization_integral}
Consider any function $f : \R \to \R$. If we draw $Z_j \sim Poi(\lam \nu_j)$ for $j \in [m]$, and $$\prod_{j=1}^m \E[f(Z_j)]$$ is analytic in $\lam$, we have
$$\E\left[\prod_{j=1}^m f(Y_j^n)\right] = \frac{n!}{2 \pi i} \oint \frac{e^\lam}{\lam^{n+1}} \prod_{j=1}^m \E[f(Z_j)] d \lam$$
where $Y_1^n, \dots, Y_{m}^n$ are $n$ samples drawn according to $\nu$.
\end{lemma}

\begin{proof}
Conditioning on $\sum_{j=1}^m Z_j = k$, we have
$$\prod_{j=1}^m \E\left[f(Z_j)\right] = \E\left[\prod_{j=1}^m f(Z_j)\right] = \sum_{k = 0}^\infty \Pr\left[\sum_{j=1}^m Z_j = k\right] \E\left[\prod_{j=1}^m f(Z_j) \Big| \sum_{j=1}^m Z_j = k\right]$$
$$ = \sum_{k=0}^\infty \frac{\lam^k}{k!} e^{-\lam} \E\left[\prod_{j=1}^m f(Y_j^n)\right]$$

Now, for any analytic function $\phi(\lam)$ with power series expansion given by
\[
  \phi(\lam) = \sum_{k=0}^\infty a_k \lam^k
\]
we have by Cauchy's theorem that
\[
  a_n = \frac{1}{2 \pi i} \oint \phi(\lam) \frac{1}{\lam^{n+1}} d\lam.
\]

By assumption, $\prod_{j=1}^m \E[f(Z_j)]$ is analytic in $\lam$. Therefore,
$$\E\left[\prod_{j=1}^m f(Y_j^n)\right] = \frac{n!}{2 \pi i} \oint \frac{e^\lam}{\lam^{n+1}} \prod_{j=1}^m \E[f(Z_j)] d \lam$$
which is the desired bound.

\end{proof}
\begin{corollary}
Under Assumption \ref{assumption:xi},
\begin{equation}
\label{eq:depoissonization_integral}
\exp(-\eps^2 \theta n) \E\left[\exp\left(\eps^2 \theta \sum_{j=1}^m \xi(Y_j^n)\right)\right] = \frac{n!}{2 \pi i} \oint e^{\lam} A_\lam(\theta) \frac{d \lam}{\lam^{n+1}}
\end{equation}
\end{corollary}

We will choose a contour passing through a particular $\lam_0$, and this will make it easy to evaluate the integral. 
We carry out the integration along the contour given by $\lam = \lam_0 e^{i \psi}$, where
$$\lam_0 = n(1 - \eps^2 \theta)$$

We substitute $\lam_0 e^{i \psi}$ into \eqref{eq:depoissonization_integral} to get that

\begin{equation}
\label{eq:huber_mgf_integral}
\exp(-\eps^2 \theta n) \E\left[\exp\left(\eps^2 \theta \sum_{j=1}^m \xi(Y_j^n)\right)\right] = e^{-\eps^2 \theta n} \frac{n!}{2 \pi} \lam_0^{-n} Re\left[ \int_{-\pi}^\pi g(\psi) d \psi\right]
\end{equation}
with 
\begin{equation}
\label{eq:huber_h_def}
g(\psi) := e^{-i n \psi}  \prod_{j=1}^m\left\{\sum_{k=0}^\infty \frac{(\lam_0 \nu_j e^{i \psi})^k}{k!}\exp(\eps^2 \theta  \xi(k)) \right\}
\end{equation}

%





We will split this integral into $3$ parts. Let 
\begin{equation}
    \label{eq:huber_split_integrals}
    \begin{gathered}
        I_1 = Re\left[\int_{-\pi/3}^{\pi/3} g(\psi) d \psi \right]\\
        I_2 = Re\left[\int_{-\pi}^{-\pi/3} g(\psi) d \psi \right]\\
        I_3 = Re\left[\int_{\pi/3}^{\pi} g(\psi) d \psi \right]
    \end{gathered}
\end{equation}

We will show that $I_1$ dominates. We show this by bounding $g(\psi)$ in the region $\psi \in [-\pi, -\pi/3]\cup [\pi/3, \pi]$ as follows.
\begin{lemma}
\label{lem:huber_small_integrals_bound}
Under Assumption \ref{assumption:xi}, and $m \geq C \log n$ for sufficiently large constant $C$, for $\psi \in [-\pi, -\pi/3] \cup [\pi/3, \pi]$, $$|g(\psi)| \le O\left(\frac{e^n}{n}\right)$$
\end{lemma}

\begin{proof}
By definition of $g$ from \eqref{eq:huber_h_def}, and using the assumption on $\xi$ from Assumption \ref{assumption:xi}, we have that,
\begin{align*}
|g(\psi)| &=\left| e^{-i n \psi}  \prod_{j=1}^m\left\{\sum_{k=0}^\infty \frac{(\lam_0 \nu_j e^{i \psi})^k}{k!} \exp(\eps^2 \theta \xi(k)) \right\}\right| \\
&\le \left|\prod_{j=1}^m \left\{ \sum_{k=0}^\infty \frac{(\lam_0 \nu_j e^{i \psi})^k}{k!}\right\} \right| + \left| \prod_{j=1}^m \left\{ \sum_{k=0}^\infty \frac{(\lam_0 \nu_j e^{i \psi})^k}{k!} \left(\sum_{l=1}^\infty \frac{(\eps^2 \theta)^l}{l!} \xi(k)^l \right)\right\} \right|  
\end{align*}
Now, for choice of $\lam_0 = n(1 - \eps^2 \theta)$, and $\psi \in [-\pi, -\pi/3] \cup [\pi/3, \pi]$, $$ \left|\prod_{j=1}^m \left\{ \sum_{k=0}^\infty \frac{(\lam_0 \nu_j e^{i \psi})^k}{k!}\right\} \right| = \left| \prod_{j=1}^m e^{\lam_0 \nu_j e^{i \psi}} \right| = \left|e^{\lam_0 e^{i \psi}} \right| = |e^{n(1 - \eps^2 \theta) e^{i \psi}}| \le O(e^{0.5n})$$
For the second term, by Assumption \ref{assumption:xi}, for our $\lam_0 = n(1 - \eps^2 \theta)$,  and $\nu_j = 1/m + O(\eps/m)$ for all $m$, this is 
\begin{align*}
e^{\lam_0} \prod_{j=1}^m \left\{\sum_{l=1}^\infty \frac{(\eps^2 \theta)^l}{l!} \E[\xi(Z_j)^l] \right\} &= e^{n (1 - \eps^2\theta)} \prod_{j=1}^m \left\{\eps^2 \theta \E[\phi(Z_j)^2] + \frac{\eps^4}{2} \theta^2 \E[\phi(Z_j)^4] + o(\Delta^2) \right\}
\end{align*}
By Lemma \ref{lem:theta_phi(Z_j)_o(1)}, and since $n = o(m/\eps^2)$, this is 
$$e^{n(1 - \eps^2 \theta)} \prod_{j=1}^m O\left(\frac{n \eps^2}{m}\right)\le e^{n(1 - \eps^2 \theta) - \Omega(m)}$$

Since $m \ge C \log n$ for sufficiently large constant $C$, the claim follows.
\end{proof}

Note that this implies that for the integrals defined in \eqref{eq:huber_split_integrals} that 
\begin{equation}
\label{eq:huber_small_integrals}
I_2 + I_3 = O\left(\frac{e^{n}}{n}\right)
\end{equation}

Now, we will compute $I_1$. Define $G(\psi) := \log(g(\psi))$. Then, by definition of $g$ in \eqref{eq:huber_h_def},

\begin{equation}
\label{eq:huber_H(psi)}
G(\psi) = - i n\psi  + \sum_{j=1}^m \log\left\{ \sum_{k=0}^\infty \frac{(\lam_0 \nu_j e^{i \psi})^k}{k!} \exp(\eps^2 \theta \xi(k)) \right\}
\end{equation}

Note that \begin{equation}
\label{eq:Huber_G(0)_imaginary_0}
    \Im(G(0)) = 0
\end{equation}

Then, applying Lemma \ref{lem:squared_huber_H_derivatives},
\begin{equation}
\label{eq:huber_G'(0)_0}
\Re(G'(0)) = 0 
\end{equation}
Computing the asymptotic expansion of $G''(\psi)$ by Lemma \ref{lem:G_second_derivative}, we have
\begin{equation}
\label{eq:huber_H''(psi)_asymptotic}
G''(\psi) = - n e^{i \psi} + O\left(\frac{n^2 \eps^2}{m} \right) + o(1)
\end{equation}

Now, by Taylor's theorem, for any $\psi \in [- \pi/3, \pi/3]$ there exists $\tilde{\psi} \in (0, \psi)$ such that
\begin{equation}
\label{eq:huber_taylors_theorem}
G(\psi) = G(0) + G'(0) \psi + \frac{G''(\tilde{\psi})}{2} \psi^2
\end{equation}

But, by \eqref{eq:huber_H''(psi)_asymptotic}, $Re[G''(\psi)] \le -0.4 n$ for any $\psi \in [-\pi/3, \pi/3]$. So, for $\psi \in [-\pi/3, \pi/3]$, 
\begin{equation}
\label{eq:Huber_G(psi)_at_most_G(0)_with_residual}
Re(G(\psi)) \le G(0) - 0.2 n \psi^2
\end{equation}

Now, we have the following upper bound on $I_1$.

\begin{lemma}
\label{lem:huber_I_1_upper_bound}
$$I_1 \le e^{G(0)} \frac{\sqrt{\pi}}{\sqrt{0.2n}}$$
\end{lemma}

\begin{proof}
\begin{equation}
I_1 = Re\left[\int_{-\pi/3}^{\pi/3} e^{G(\psi)} d \psi\right] \le \int_{- \pi /3}^{\pi/3} e^{Re(G(\psi))} d \psi \le e^{G(0)} \int_{-\pi/3}^{\pi/3} e^{-0.2 n \psi^2} d \psi
\end{equation}

\end{proof}

$$ \le e^{G(0)} \int_{-\infty}^\infty e^{-0.2 n \psi^2}d \psi = e^{G(0)} \frac{\sqrt{\pi}}{\sqrt{0.2n}}$$

The next lemma shows that $I_1$ is also lower bounded by the above quantity (up to constants).

\begin{lemma}
\label{lem:huber_I_1_lower_bound}
$$I_1 \ge e^{G(0)} \frac{0.5 \sqrt{\pi}}{\sqrt{1.1n}} (1 + o(1))$$
where $I_1$ is defined in \eqref{eq:huber_split_integrals}
\end{lemma}

\begin{proof}
By \eqref{eq:huber_H''(psi)_asymptotic},  $\Im(G''(\psi)) = - n \sin(\psi) + O\left(\frac{n^2 \eps^2}{m}\right)$. So, for large enough $n$, since $|\sin(\psi)| \le |\psi|$, for any $\psi \in [-\pi/3, \pi/3]$, $|\Im(G''(\psi))| \le 1.1 n |\psi|$. So, by \eqref{eq:huber_taylors_theorem}, \eqref{eq:Huber_G(0)_imaginary_0} and \eqref{eq:huber_G'(0)_0}, we have that for constant $c > 0 $ and $\psi \in [-\pi/3, \pi/3]$, 
$$|\Im(G(\psi))| \le 1.1 n |\psi|^3 + c n \eps^2 \psi^2$$

Also, $\Re(G''(\psi)) \ge -1.1 n$ by a similar argument. So, by \eqref{eq:huber_taylors_theorem} and \eqref{eq:huber_G'(0)_0}, for $\psi \in [-\pi/3, \pi/3]$,
$$\Re(G(\psi)) \ge G(0) - 1.1 n \psi^2$$

Now, for $t_n = 0.1 \min\{n^{-1/3}, \frac{1}{\eps\sqrt{c n} }\}$, we have that for $\psi \in [-t_n, t_n]$, $\cos(\Im(G(\psi))) \ge 0.5$ so that $\Re(e^{G(\psi)}) \ge 0.5 e^{\Re(G(\psi)}$. We can split $I_1$ further into $3$ parts:

$$I_1 = \Re\left[ \int_{-\pi/3}^{-t_n} e^{G(\psi)} d \psi\right] + \Re\left[\int_{t_n}^{\pi/3} e^{G(\psi)} d \psi \right] + \Re\left[\int_{-t_n}^{t_n} e^{G(\psi)} d \psi \right]$$

Now, by \eqref{eq:Huber_G(psi)_at_most_G(0)_with_residual}, $$\left| \int_{-\pi/3}^{-t_n}e^{G(\psi)} d \psi \right| \le e^{G(0)} \int_{- \infty}^{-t_n} e^{-0.2 n \psi^2} d \psi = t_n e^{G(0)} \int_{-\infty}^{-1} e^{-0.2 n t_n^2 \bar{\psi}^2} d \bar \psi$$
$$ \le t_n e^{G(0)} \int_{-\infty}^{-1} e^{-0.2 n t_n^2 |\bar \psi|} d \bar \psi = e^{G(0)} O\left(\frac{1}{n t_n} \right) = e^{G(0)} o\left(\frac{1}{\sqrt{n}} \right)$$

In a similar way, we can bound the second term. For the third term, we have $$\Re\left[ \int_{-t_n}^{t_n} e^{G(\psi)} d \psi \right] \ge \int_{-t_n}^{t_n} 0.5 e^{\Re(G(\psi))} d \psi \ge 0.5 e^{G(0)} \int_{-t_n}^{t_n} e^{-1.1 n \psi^2} d \psi$$

$$\ge 0.5 e^{G(0)} \left[ \int_{-\infty}^\infty e^{-1.1 n \psi^2} d \psi - 2 \int_{-\infty}^{- t_n} e^{-1.1 n \psi^2} d \psi \right]$$
$$\ge 0.5 e^{G(0)} \left(\frac{\sqrt{\pi}}{\sqrt{1.1 n}} + O\left(\frac{1}{n t_n}\right) \right) = 0.5 e^{G(0)} \frac{\sqrt{\pi}}{\sqrt{1.1 n}} (1 + o(1))$$

Combining the bounds, we get that $$I_1 \ge e^{G(0)} \frac{0.5 \sqrt{\pi}}{\sqrt{1.1 n}} (1 + o(1))$$

\end{proof}

Combining the upper bound on $I_1$ from Lemma \ref{lem:huber_I_1_upper_bound} and the lower bound from Lemma \ref{lem:huber_I_1_lower_bound}, we have

$$I_1 = e^{G(0)} \frac{1}{\sqrt{n}} e^{O(1)}$$

So, by \eqref{eq:huber_mgf_integral} and \eqref{eq:huber_small_integrals}, 
\begin{equation}
\label{eq:huber_mgf_H(0)}
\exp(-\eps^2 \theta n) \E_\nu\left[\xi(Y_j^n) \right] = e^{- \eps^2 \theta n} \frac{n!}{2 \pi} \lam_0^{-n} e^{G(0)} \frac{\sqrt{\pi}}{\sqrt{0.2n}}(1+o(1))
\end{equation}
So, it remains to compute $G(0)$.
\begin{lemma}
Under Assumption~\ref{assumption:xi},
\label{lem:huber_G(0)_computation}
\begin{dmath}
\label{eq:huber_G(0)}
G(0) = \lam_0 + \sum_{j=1}^m \left\{ \eps^2 \theta \left[(\lam_0 \nu_j)^2 + \lam_0 \nu_j - 2 \lam_0 \nu_j \frac{n}{m} + \frac{n^2}{m^2} \right] + \frac{\eps^4\theta^2}{2} \left[4 (\lam_0 \nu_j)^3 + 6 (\lam_0 \nu_j)^2 + (\lam_0 \nu_j) - 8(\lam_0 \nu_j)^2 \frac{n}{m} - 4\frac{n}{m}(\lam_0 \nu_j) + 4 \frac{n^2}{m^2}(\lam_0 \nu_j) \right] + o\left(\Delta^2 \right) \right\}
\end{dmath}
\end{lemma}

\begin{proof}
Using equation \eqref{eq:huber_H(psi)}, we have
$$G(0) = \lam_0 + \sum_{j=1}^m \log\left\{\E\left[\exp(\eps^2 \theta\xi(Z_j))\right]\right\} $$
where $Z_j \sim Poi(\lam_0 \nu_j)$.
By Assumption~\ref{assumption:xi}, we have that this is
$$\lam_0 + \sum_{j=1}^m \log \left\{\E\left[f(Z_j)\right] + o\left(\Delta^2\right)\right\}$$

Using the definition of $f$ from \eqref{eq:huber_f_def}, this is
$$\lam_0 + \sum_{j=1}^m \log \left\{ 1 + \eps^2 \theta \E[\phi(Z_j)^2] +  \frac{\eps^4\theta^2}{2} \E[\phi(Z_j)^4] + o\left(\Delta^2 \right)\right\}$$

Since by Lemma \ref{lem:theta_phi(Z_j)_o(1)}, $\eps^2 \theta \E[\phi(Z_j)^2]$ and $\eps^4 \frac{\theta^2}{2} \E[\phi(Z_j)^4]$ are $o(1)$, using the fact that $\log(1 + x) = x - x^2/2 + O(x^3)$ for $x \to 0$, we have $$G(0) = \lam_0 + \sum_{j=1}^m \left\{ \eps^2 \theta \E[\phi(Z_j)^2] + \frac{\eps^4 \theta^2}{2} \left( \E[\phi(Z_j)^4] - \E[\phi(Z_j)^2]^2 \right) + o\left(\Delta^2 \right) \right\} $$

Now, using Lemma \ref{lem:phi(Z_j)_expansions}, we have that 
\begin{dmath*}
G(0) = \lam_0 + \sum_{j=1}^m \left\{ \eps^2 \theta \left[(\lam_0 \nu_j)^2 + \lam_0 \nu_j - 2 \lam_0 \nu_j \frac{n}{m} + \frac{n^2}{m^2} \right] + \frac{\eps^4\theta^2}{2} \left[4 (\lam_0 \nu_j)^3 + 6 (\lam_0 \nu_j)^2 + (\lam_0 \nu_j) - 8(\lam_0 \nu_j)^2 \frac{n}{m} - 4\frac{n}{m}(\lam_0 \nu_j) + 4 \frac{n^2}{m^2}(\lam_0 \nu_j) \right] + o\left(\Delta^2 \right) \right\}
\end{dmath*}

\end{proof}

\begin{lemma}
\label{lem:huber_G(0)_uniform_computation}
For the uniform distribution so that $\nu_j = 1/m$ for all $j$, and $\lam_0=n(1 - \eps^2 \theta)$,
$$ G(0) = \lam_0 + \eps^2 \theta n + \eps^4 \theta^2 \left(-\frac{n}{2} + \frac{n^2}{m} \right) + o\left(\frac{n^2 \eps^4}{m} \right)$$
\end{lemma}
\begin{proof}
Substituting $\nu_j = 1/m$ for all $j$, and $\lam_0 = n(1 - \eps^2 \theta)$, we have $$\sum_{j=1}^m \left[(\lam_0 \nu_j)^2 + \lam_0 \nu_j - 2 \lam_0 \nu_j \frac{n}{m} + \frac{n^2}{m^2}\right] =  n - n \eps^2\theta + \frac{n^2 \eps^4}{m} \theta^2$$

\begin{dmath*}
\sum_{j=1}^m \left[ 4 (\lam_0 \nu_j)^3 + 6 (\lam_0 \nu_j)^2 + (\lam_0 \nu_j) - 8 (\lam_0 \nu_j)^2 \frac{n}{m} - 4 \frac{n}{m} (\lam_0 \nu_j) + 4 \frac{n^2}{m^2} (\lam_0 \nu_j) \right]= n + 2 \frac{n^2}{m} + O\left(\frac{n^2 \eps^2}{m}\right)
\end{dmath*}

So, by Lemma \ref{lem:huber_G(0)_computation}, we have
$$G(0) = \lam_0 +  \eps^2 \theta n + \eps^4 \theta^2 \left(\frac{- n}{2} + \frac{n^2}{m} \right) + o\left(\frac{n^2 \eps^4}{m} \right)$$

\end{proof}

\begin{lemma}
\label{lem:Huber_G(0)_alternate_distributions}
For alternate distributions such that $\nu_j = \frac{1}{m} + \frac{\eps}{\gamma m}$ for $j \le \gamma m$, and $\nu_j = \frac{1}{m} - \frac{\eps}{(1-\gamma) m}$ for $j > \gamma$, for $\gamma = \Theta(1), 1-\gamma = \Theta(1)$, and $\lam_0 = n(1 - \eps^2 \theta)$, we have
$$G(0) = \lam_0  + \eps^2 \theta n + \eps^4 \theta^2 \left(\frac{n}{2} + \frac{n^2}{m}\right) + \eps^4 \theta \frac{n(\gamma m^2 \theta - \gamma^2 m^2 \theta - mn)}{\gamma(\gamma - 1) m^2 } + o\left(\frac{n^2 \eps^4}{m} \right)$$
\end{lemma}

\begin{proof}
We have $$\sum_{j=1}^m \left[ (\lam_0 \nu_j)^2 + \lam_0 \nu_j - 2 \lam_0 \nu_j \frac{n}{m} + \frac{n^2}{m^2} \right] = n - \frac{n(m n + \gamma^2 m^2 \theta - \gamma m^2 \theta) \eps^2}{\gamma(\gamma-1)m^2} + o\left(\frac{n^2 \eps^2}{m}\right)$$

\begin{dmath*}
\sum_{j=1}^m \left[4 (\lam_0 \nu_j)^3 + 6 (\lam_0 \nu_j)^2 + (\lam_0 \nu_j) - 8 (\lam_0 \nu_j)^2 \frac{n}{m} - 4 \frac{n}{m} (\lam_0 \nu_j) + 4 \frac{n^2}{m^2} (\lam_0 \nu_j) \right] = n + 2\frac{n^2}{m} + O\left(\frac{n^2 \eps^2}{m}  \right)
\end{dmath*}

Thus, for this $\nu$, from Lemma \ref{lem:huber_G(0)_computation},
$$G(0) = \lam_0  + \eps^2 \theta n + \eps^4 \theta^2 \left(\frac{n}{2} + \frac{n^2}{m}\right) + \eps^4 \theta \frac{n(\gamma m^2 \theta - \gamma^2 m^2 \theta - mn)}{\gamma (\gamma - 1) m^2} + o\left(\frac{n^2 \eps^4}{m} \right)$$
\end{proof}


Finally, we prove the main lemma.
\begin{proof}[Proof of Lemma \ref{lem:main_generic_mgf_lemma}]
By \eqref{eq:huber_mgf_H(0)}, substituting $\lam_0 = n(1 - \eps^2 \theta)$ and $G(0)$ from Lemma \ref{lem:huber_G(0)_uniform_computation}, we have for uniform $\nu$,
\begin{dmath*}
\exp(-\eps^2 \theta n) \E_\nu\left[\xi(Y_j^n) \right] = e^{-\eps^2 \theta n} \frac{e^n n!}{\sqrt{2 \pi n}} (n(1 - \eps^2 \theta))^{-n} \exp \left\{ - n \eps^2 \theta + n \eps^2 \theta + \eps^4 \theta^2\left(- \frac{n}{2} + \frac{n^2}{m}  \right) + o\left(\frac{n^2 \eps^4}{m} \right) \right\}
\end{dmath*}

$$ =e^{-\eps^2 \theta n} \frac{e^n n!}{n^n \sqrt{2 \pi n}} \exp\left\{-n\left(-\eps^2\theta - \frac{\eps^4 \theta^2}{2}\right) \right\}\exp\left\{\eps^4 \theta^2 \left(-\frac{n}{2} + \frac{n^2}{m} \right) + o\left(\frac{n^2 \eps^4}{m} \right) \right\}$$

$$ = \frac{e^n n!}{n^n \sqrt{2 \pi n}} \exp\left\{\eps^4 \theta^2 \left(\frac{n^2}{m} \right) + o\left(\frac{n^2 \eps^4}{m} \right) \right\}$$
$$ = (1 + O(1/n)) \exp\left\{ \frac{n^2 \eps^4}{m} \left(\theta^2 + o(1) \right) \right\}$$

By \eqref{eq:huber_mgf_H(0)}, substituting $\lam_0 = n(1-\eps^2 \theta)$ and $G(0)$ from Lemma \ref{lem:Huber_G(0)_alternate_distributions}, for $\nu$ such that $\nu_j = \frac{1}{m} + \frac{\eps}{\gamma m}$ for $j \le \gamma m$ and $\nu_j = \frac{1}{m} - \frac{\eps}{(1-\gamma) m}$ for $j > (1-\gamma) m$, with $\gamma = \Theta(1), (1-\gamma) = \Theta(1)$, we have

\begin{dmath*}
\exp(-\eps^2 \theta n) \E_\nu[\xi(Y_j^n)] = (1 + O(1/n)) \exp\left\{\frac{n^2 \eps^4}{m} \left( \theta^2 + \theta \frac{1}{\gamma(1-\gamma)} + o(1)\right) \right\}
 \end{dmath*}
\end{proof}

Finally, this gives us the MGF of the Huber statistic.

\begin{lemma}
\label{lem:huber_mgfs}
We have that for uniform $\nu$ such that $\nu_j = 1/m$ for all $j$, $$\E\left[\exp\left(\frac{n^2 \eps^4}{m} \theta \Tilde{S}\right) \right] = \exp(-\eps^2 \theta n) \E\left[\exp\left(\eps^2 \theta \sum_{j=1}^m h_\beta\left(Y_j^n - \frac{n}{m} \right) \right)\right]$$
$$= (1 + O(1/n)) \exp\left\{\frac{n^2 \eps^4}{m} (\theta^2 + o(1)) \right\}$$

and for alternate $\nu$ such that $\nu_j = \frac{1}{m} + \frac{\eps}{\gamma m}$ for $j \le \gamma m$ and $\nu_j = \frac{1}{m} - \frac{\eps}{(1-\gamma) m}$ for $j > (1-\gamma) m$, for $\gamma = \Theta(1), 1-\gamma = \Theta(1)$,
$$\E\left[\exp\left(\frac{n^2 \eps^4}{m} \theta \Tilde{S}\right) \right] = \exp(-\eps^2 \theta n) \E\left[\exp\left(\eps^2 \theta \sum_{j=1}^m h_\beta\left(Y_j^n - \frac{n}{m} \right) \right)\right]$$
\begin{dmath*}
 = (1 + O(1/n)) \exp\left\{\frac{n^2 \eps^4}{m} \left( \theta^2 + \theta \frac{1}{\gamma(1-\gamma)} + o(1)\right) \right\}
\end{dmath*}
\end{lemma}
\begin{proof}
Note that Assumption \ref{assumption:xi} holds for $A_\lam(\theta)$ as defined in \eqref{eq:huber_A_lam(theta)} due to Lemma~\ref{lem:satisfying_huber_assumptions}. So, by Lemma \ref{lem:main_generic_mgf_lemma}, the claim holds.
\end{proof}

\subsection{Application of the G{\"a}rtner-Ellis Theorem}

In this section, we apply the G{\"a}rtner-Ellis Theorem to obtain the probability that our statistic crosses a threshold, under the uniform distribution, and under one of the worst-case $\eps$-far distributions.

\begin{lemma}
\label{lem:gartner_ellis_application_huber}
Under the uniform distribution $p$, we have that for $\tau > 0$, $$\lim_{n \to \infty} - \frac{m}{n^2 \eps^4} \log\left(\Pr_p\left[\Tilde{S} \ge \tau \right] \right) = \frac{\tau^2}{4}$$


Under an $\eps$-far distribution $q$ of the form $q_j = \frac{1}{m} + \frac{\eps}{\gamma m}$ for $j \le l$ and $q_j = \frac{1}{m} - \frac{\eps}{(1-\gamma)m}$ for $j > l$, and $\gamma = \Theta(1)$, $1-\gamma = \Theta(1)$, for $\tau < \frac{1}{\gamma(1-\gamma)}$,

$$\lim_{n \to \infty} -\frac{m}{n^2 \eps^4} \log \left(\Pr_q\left[\Tilde{S} \le \tau \right] \right) = \frac{\left(\tau \gamma(\gamma - 1) + 1\right)^2}{4 \gamma^2 (\gamma - 1)^2 }$$

\end{lemma}
\begin{proof}
Note that by Lemma \ref{lem:huber_mgfs}, the limiting logarithmic moment generating function with respect to the uniform distribution $p$ is given by $$\Lambda_p(\theta) = \lim_{n \to \infty} \frac{m}{n^2 \eps^4} \log\left(\E_p\left[\exp\left(\frac{n^2 \eps^4}{m} \theta\Tilde{S}\right) \right]\right) = \theta^2$$

Thus, Assumption \ref{assumption:gartner_ellis} holds for $\mathcal{D}_{\Lambda_p} = \R$. Furthermore, the Fenchel-Legendre Transform (defined in equation \ref{eq:fenchel_legendre}) of $\Lambda_p$ is given by $$\Lambda_p^*(\tau) = \sup_{\theta } \{\theta \tau - \theta^2\} = \frac{\tau^2}{4}$$

This is a strongly convex function of $\tau$, so the set of exposed points of $\Lambda_p^*$ whose exposing hyperplane belongs to $\mathcal{D}_{\Lambda_p}^o$ is all of $\R$. Thus, by the  Theorem \ref{thm:gartner_ellis} (G{\"a}rtner-Ellis), for $\tau > 0$,

$$ \lim_{n \to \infty} -\frac{m}{n^2 \eps^4} \log\left(\Pr_p\left[\Tilde{S} \ge \tau\right]\right) = \inf_{x \ge \tau} \Lambda_p^*(x) = \frac{\tau^2}{4}$$

%

Similarly, the limiting logarithmic moment generating function with respect to an alternate distribution $q$ is given by $$\Lambda_q(\theta) = \lim_{n \to \infty} - \frac{m}{n^2 \eps^4} \log \left( \E_q\left[\exp\left(\frac{n^2 \eps^4}{m} \theta \Tilde{S} \right) \right] \right) = \theta^2 + \frac{1}{\gamma(1-\gamma)} \theta$$

The Fenchel-Legendre transform is given by $$\Lambda_q^*(\tau) = \sup_{\theta} \left\{\theta \tau - \theta^2 - \frac{1}{\gamma(1-\gamma)} \theta\right\} = \frac{\left(\tau \gamma(\gamma-1) + 1\right)^2}{4 \gamma^2 (\gamma-1)^2}$$

Again, applying the G{\"a}rtner-Ellis Theorem gives, for $\tau < \frac{1}{\gamma(1-\gamma)}$,
$$\lim_{n \to \infty} - \frac{m}{n^2 \eps^4} \log\left(\Pr_q\left[\Tilde{S} _n^* \le \tau \right] \right) = \inf_{x \le \tau} \Lambda^*_q(x) = \frac{\left(\tau \gamma(\gamma-1) + 1 \right)^2}{4 \gamma^2 (\gamma-1)^2}$$


\end{proof}

\subsection{Setting the threshold}

We need to set our threshold $\tau$ so that the minimum of the error probability under the uniform distribution $p$, and any $\eps$-far distribution $q$ is maximized. Note that by Lemma \ref{lem:gartner_ellis_application_huber}, it is sufficient to consider a threshold $\tau$ such that $0 < \tau < \frac{1}{\gamma(1-\gamma)}$, since otherwise, the error probability in one of the two cases is at least constant.  To set our threshold, we will first observe that for any $\tau$ in this range, the ``error exponent'' under $\eps$-far distributions is minimized for a particular $\eps$-far distribution. Then, we will set the threshold to maximize the minimum of the error exponent under the uniform distribution, and under this $\eps$-far distribution.

\begin{lemma}
\label{lem:threshold_huber}
Setting the threshold $\tau = 2$, we have for the uniform distribution $p$, $$\lim_{n \to \infty} -\frac{m}{n^2 \eps^4} \log \left( \Pr_p\left[\Tilde{S} \ge \tau \right] \right) = 1$$
and for any $\eps$-far distribution $q$ such that $q_j = \frac{1}{m} + \frac{\eps}{\gamma m}$ for $j \le \gamma m$ and $q_j = \frac{1}{m} - \frac{\eps}{(1-\gamma)m}$ for $j > \gamma m$ and $\gamma = \Theta(1), 1-\gamma = \Theta(1)$,$$\lim_{n \to \infty} - \frac{m}{n^2 \eps^4} \log \left( \Pr_q \left[ \Tilde{S} \le \tau \right]\right) \ge 1$$
 with equality for $q$ such that $q_j = \frac{1}{m} + \frac{2 \eps}{m}$ for $j \le m/2$ and $q_j = \frac{1}{m} - \frac{2 \eps}{m}$ for $j > m/2$.
\end{lemma}

\begin{proof}
By Lemma \ref{lem:gartner_ellis_application_huber}, for $0 < \tau < \frac{1}{\gamma(1-\gamma)}$, $$\lim_{n \to \infty} \frac{m}{n^2 \eps^4} \log \left( \Pr_q \left[\Tilde{S} \le \tau \right]\right) = \frac{\left(\tau \gamma(\gamma-1) + 1\right)^2}{4 \gamma^2 (\gamma-1)^2}$$

Now, the numerator of the right hand side is minimized when $\gamma=1/2$, and the denominator is maximized when $\gamma=1/2$. Thus, the right hand side is minimized when $\gamma=1/2$. So, we have that,
$$\lim_{n \to \infty} -\frac{m}{n^2 \eps^4} \log\left(\Pr_{q}\left[ \Tilde{S} \le \tau \right] \right) \ge \frac{1}{4} (\tau - 4)^2$$
 with equality for distribution $q$ such that $q_j = 1/m + 2 \eps/m$ for $j \le m/2$ and $q_j = 1/m - 2 \eps/m$ for $j > m/2$.
 
Then, the claim follows by substituting in $\tau = 2$ in the expression for the uniform distribution in Lemma \ref{lem:gartner_ellis_application_huber} and in the above expression.
\end{proof}

Recall that our target sample complexity is 
\begin{align}\tag{\ref{eq:nboundtight}}
  n = (1 + o(1)) \sqrt{m \log \frac{1}{\delta}}/\eps^2
\end{align}
We have our result for the Huber tester.
\begin{theorem}[Huber with $n/m \gtrsim 1$]
\label{thm:huber_n_Omega_m}
The Huber statistic for appropriate $\beta$ achieves \eqref{eq:nboundtight} for $1 \lesssim n/m \ll 1/\eps^2$, $\eps, \delta \ll 1$ and $m \ge C \log n$ for sufficiently large constant $C$.
\end{theorem}
\begin{proof}
First we need to show that for every $(n, m, \eps)$ that satisfy our conditions, there is a $\beta, \Delta$ that satisfies \eqref{eq:Delta_constraint}, \eqref{eq:beta_second_bound} and \eqref{eq:beta_third_bound}. We will set $$\Delta = \frac{n \eps^2}{m}$$ so that \eqref{eq:Delta_constraint} is satisfied. Then, observe that \eqref{eq:beta_second_bound} and \eqref{eq:beta_third_bound} can be satisfied as long as $$\log\left(\frac{1}{\Delta}\right) + \sqrt{\frac{n}{m} \log \left(\frac{1}{\Delta}\right)} = o\left(\frac{\Delta^{1/3}}{\eps}\right)$$ 
Now, since $n = \Omega(m)$, 
$$\log\left(\frac{1}{\Delta}\right) = \log \left(\frac{m}{n \eps^2}\right) = o\left(\log\left(\frac{1}{\eps}\right)\right) = o\left(\frac{1}{\eps^{1/3}}\right) = o\left(\frac{n^{1/3}}{m^{1/3} \eps^{1/3}} \right) = o\left(\frac{\Delta^{1/3}}{\eps}\right)$$
and since $n = o\left(\frac{ m}{\eps^2\log^3\left(\frac{m}{n \eps^2}\right)}\right)$,
$$\sqrt{\frac{n}{m} \log \left(\frac{1}{\Delta}\right)}= \sqrt{\frac{n}{m} \log \left(\frac{m}{n \eps^2} \right)} = o\left(\frac{n^{1/3}}{m^{1/3} \eps^{1/3}}\right) = o\left(\frac{\Delta^{1/3}}{\eps}\right) $$

By Lemma \ref{lem:threshold_huber}, we have that $\bar c(\eps, m, C) = 1$ for every $\eps$ that satisfies our assumptions, and every $C > 2$. In particular, any $\eps'$ such that $\left(1 - \frac{1}{C'} \right) \eps(n) \le \eps'(n) \le \eps(n)$ has $\bar{c}(\eps', m, C) = 1$ for every $C > 2$. Thus, by Lemma \ref{lem:final_worst_case_lemma}, we have that $c(\eps, m) = 1$ for every $\eps$ that satisfies our assumptions. The claim follows.
\end{proof}

Since the Huber statistic for $\beta = 0$ is equivalent to the TV statistic, Theorem~\ref{thm:huber_n_Omega_m} and Theorem~\ref{thm:tv} together give us the main result.
\huber*

%% file: mgf_computation_lemmas.tex
\section{MGF computation Lemmas}\label{section:mgf_computation_lemmas}
\subsection{Huber Lemmas}
\begin{lemma}
\label{lem:theta_phi(Z_j)_o(1)}
Suppose Assumption~\ref{assumption:huber} holds. For $Z_j \sim Poi(\lam_0 \nu_j)$ and $\lam_0 = n(1 - \eps^2 \theta)$, $\nu_j = \frac{1}{m} + O\left(\frac{\eps}{m}\right)$, and for $n = \Omega(m)$
$$\eps^2\theta \E[ \phi(Z_j + c)^2] = O\left(\frac{n \eps^2}{m} \right)$$
$$\frac{\eps^4\theta^2}{2} \E[\phi(Z_j + c)^4] = O\left(\left( \frac{n \eps^2}{m} \right)^2\right)$$
for any $0 \le c \le 4$
\end{lemma}

\begin{proof}

Note that $\lam_0 \nu_j = \frac{n}{m} + O(\frac{n \eps}{m})$. Then, using Lemma \ref{lem:phi(Z_j)_expansions},
$$\eps^2 \theta \E[\phi(Z_j)^2] = \eps^2 \theta\left( \frac{n}{m} + O\left(\frac{n\eps}{m}\right) + O\left(\frac{n^2 \eps^2}{m^2}\right)\right) = O\left(\frac{n \eps^2}{m}\right) $$
since $n = o(m/\eps^2)$.

Similarly, 
$$\eps^4 \frac{\theta^2}{2} \E[\phi(Z_j)^4] =\eps^4 \frac{\theta^2}{2} \left[\frac{n}{m} + O(\frac{n^2}{m^2}) + O(\frac{n^3 \eps^2}{m^3}) + O(\frac{n^4 \eps^4}{m^4})\right ] = O\left(\left(\frac{n \eps^2}{m}\right)^2 \right) $$

Now, $$\eps^2 \theta \E[\phi(Z_j + c)^2] = \eps^2 \theta \E\left[\left(Z_j + c - \frac{n}{m}\right)^2 \right] = \eps^2 \theta \E\left[\phi(Z_j)^2 + 2 c \left(Z_j - \frac{n}{m} \right) + c^2\right]$$
$$ = O\left(\frac{n \eps^2}{m} \right) + O(\eps^2) = O\left(\frac{n \eps^2}{m} \right)$$

since $\eps^2 = O\left( \frac{n \eps^2}{m} \right)$.

Similarly, $$\frac{\eps^4 \theta^2}{2} \E[\phi(Z_j + c)^4] = \frac{\eps^4 \theta^2}{2} \E\left[\phi(Z_j)^4 + 4 c \left(Z_j - \frac{n}{m} \right)^3 + 6 c^2 \phi(Z_j)^2 + 4 c^3 \left(Z_j - \frac{n}{m} \right) + c^4 \right]$$

Now, $$\E\left[4c \left(Z_j - \frac{n}{m} \right)^3\right] = 4 \E\left[Z_j^3 - 3 Z_j^2 \frac{n}{m} + 3 Z_j \frac{n^2}{m^2} - \frac{n^3}{m^3}\right]$$

By Lemma \ref{lem:poisson_moments}, this is $$4 \left[(\lam_0 \nu_j)^3 + 3 (\lam_0 \nu_j)^2 + \lam_0 \nu_j - 3 ((\lam_0\nu_j)^2 + \lam_0 \nu_j) \frac{n}{m} + 3 \lam_0 \nu_j \frac{n^2}{m^2} - \frac{n^3}{m^3}\right]$$

Now, since $\lam_0 \nu_j = \frac{n}{m} + O\left(\frac{n\eps}{m} \right)$, this is
$$ \frac{n}{m} + O\left(\frac{n^2\eps}{m^2} \right) + O\left(\frac{n^3 \eps^3}{m^3}\right)$$

so that $$\frac{\eps^4 \theta^2}{2} \E\left[4 c \left(Z_j - \frac{n}{m}\right)^3 \right] = O\left(\left(\frac{n \eps^2}{m}\right)^2 \right) = O\left(\left(\frac{n \eps^2}{m} \right)^2\right)$$

So, finally $$\frac{\eps^4 \theta^2}{2} \E[\phi(Z_j + c)^4] = O\left(\left( \frac{n \eps^2}{m} \right)^2\right)$$ as required.
\end{proof}
\begin{lemma}
\label{lem:squared_huber_h_bound}
Suppose Assumption~\ref{assumption:huber} holds. For $Z_j \sim Poi(\lam_0 \nu_j)$, and $\lam_0 = n(1 - \eps^2\theta)$, $\nu_j = \frac{1}{m} + O(\frac{\eps}{m})$ for all $j$,$$\prod_{j=1}^m\left\{ \E\left[f(Z_j)\right] +  o\left(\Delta^2 \right) \right\} = \exp\{O(n \eps^2) \}$$

where $f$ is defined in \eqref{eq:huber_f_def}
\end{lemma}

\begin{proof}

\begin{align*}
\prod_{j=1}^m\left\{ \E\left[f(Z_j) \right] + o\left(\Delta^2 \right)\right\} =
\exp\left\{\sum_{j=1}^m \log\left[\E\left[f(Z_j)\right] + o\left( \Delta^2 \right)\right] \right\}
\end{align*}

Note that due to Lemma \ref{lem:theta_phi(Z_j)_o(1)}, $\Delta = O\left(\frac{n \eps^2}{m} \right)$, and the fact that $n = o(m/\eps^2)$, the above is $$\exp\left\{\sum_{j=1}^m \log\left[1 + O\left(\frac{n \eps^2}{m}\right) \right] \right\}$$

We can Taylor expand the $\log$ since it is of form $\log(1+o(1))$. The above is then

$$\exp\left\{\sum_{j=1}^m \left[O\left(\frac{n \eps^2}{m}\right) \right]\right\} = \exp\{ O(n \eps^2)\}$$

Setting $\eta \ge 2$, this is $$\exp(O(n \eps^2))$$
\end{proof}
\begin{lemma}
\label{lem:huber_second_derivative_helper_lemmas}
Suppose Assumption~\ref{assumption:huber} holds. For $\lam_0 = n(1 - \eps^2 \theta)$ and $\nu_j = \frac{1}{m} + O\left(\frac{\eps}{m}\right)$ for all $j$, and for $\xi$ that satisfies Assumption~\ref{assumption:xi},
where $Z_j \sim Poi(\lam_0 \nu_j)$, we have 
\begin{align*}
&e^{-2\lam_0\nu_j}\left|\left(\sum_{k=0}^\infty \frac{(\lam_0 \nu_j e^{i\psi})^k}{k!} \exp(\eps^2 \theta \xi(k+1))\right)^2 \right. \\
&\left.- \left(\sum_{k=0}^\infty \frac{(\lam_0 \nu_j e^{i \psi})^k}{k!} \exp(\eps^2 \theta \xi(k)) \right) \left(\sum_{k=0}^\infty  \frac{(\lam_0 \nu_j e^{i \psi})^k}{k!} \exp(\eps^2 \theta \xi(k+2))\right)\right| = O(\eps^2) + O\left(\left(\frac{n \eps^2}{m} \right)^2 \right)
\end{align*}

and $$e^{-2\lam_0\nu_j}\left|\left(\sum_{k=0}^\infty \frac{(\lam_0 \nu_j e^{i\psi})^k}{k!} \exp(\eps^2 \theta \xi(k)) \right) \left(\sum_{k=0}^\infty \frac{(\lam_0 \nu_j e^{i\psi})^k}{k!} \exp(\eps^2 \theta \xi(k+1)) \right)\right| = 1 + O\left(\frac{n \eps^2}{m}\right)$$

\end{lemma}

\begin{proof}
\textbf{Notation:} For simplicity, let $\PE[f(W_j)] = \sum_{k=0}^\infty \frac{(\lam_0 \nu_j e^{i\psi})^k}{k!} f(k) $
Note that $$\PE[f(W_j) + g(W_j)] = \PE[f(W_j)] + \PE[g(W_j)]$$
Also, note that $\left|e^{-\lam_0\nu_j} \PE[f(W_j)]\right| = \E[f(Z_j)]$ where $Z_j \sim Poi(\lam_0 \nu_j)$.
We have that,
\begin{align*}
    &\PE[\exp(\eps^2 \theta \xi(W_j+1))]^2 = \PE\left[1 + \eps^2 \theta \xi(W_j + 1) + \sum_{l=2}^\infty \frac{(\eps^2 \theta \xi(W_j+1))^l}{l!} \right]^2\\
    &= \PE\left[1 + \eps^2 \theta \xi(W_j+1)\right]^2 + 2 \PE\left[1 + \eps^2 \theta \xi(W_j+1)\right] \PE\left[\sum_{l=2}^\infty \frac{(\eps^2 \theta \xi(W_j+1))^l}{l!} \right] + \PE\left[\sum_{l=2}^\infty \frac{(\eps^2 \theta \xi(W_j+1))^l}{l!} \right]^2
\end{align*}
Similarly,
\begin{align*}
    \PE[\exp(\eps^2 \theta \xi(W_j))]\PE[\exp(\eps^2 \theta \xi(W_j + 2))] = \PE[1 + \eps^2 \theta \xi(W_j)]\PE[1 + \eps^2 \theta \xi(W_j + 2)]\\
    + \PE[1 + \eps^2 \theta \xi(W_j)]\PE\left[\sum_{l=2}^\infty \frac{(\eps^2 \theta \xi(W_j+2))^l}{l!} \right]\\
    + \PE\left[\sum_{l=2}^\infty \frac{(\eps^2 \theta \xi(W_j))^l}{l!} \right]\PE[1 + \eps^2 \theta \xi(W_j+1)] + \PE\left[\sum_{l=2}^\infty \frac{(\eps^2 \theta \xi(W_j))^l}{l!} \right]\PE\left[\sum_{l=2}^\infty \frac{(\eps^2 \theta \xi(W_j+2))^l}{l!} \right] 
\end{align*}

So, using the properties of $\xi$ from Assumption~\ref{assumption:xi}, this first expression is
\begin{align*}
&e^{-2 \lam_0 \nu_j}\left| \PE[\exp(\eps^2 \theta \xi(W_j + 1))]^2 - \PE[\exp(\eps^2 \theta \xi(W_j))]\PE[\exp(\eps^2 \theta \xi(W_j + 2))] \right|\\
&\le e^{-2\lam_0 \nu_j}\left| \PE[1 + \eps^2 \theta \xi(W_j + 1)]^2 - \PE[1 + \eps^2 \theta \xi(W_j)] \PE[1 + \eps^2 \theta \xi(W_j + 2)] \right| + o(\Delta^2)\\
&= e^{-2 \lam_0 \nu_j} \left| \eps^2 \theta\PE[1] \left\{2\PE[\xi(W_j + 1)] - \PE[\xi(W_j)] - \PE[\xi(W_j + 2)] \right\}\right.\\
&\left.+ \eps^4\theta^2 \left\{\PE[\xi(W_j +1)]^2 - \PE[\xi(W_j)]\PE[\xi(W_j + 2)]\right\} \right| + o(\Delta^2)\\
&= \eps^2 \theta \left\{2\E[\xi(Z_j + 1)] - \E[\xi(Z_j)] - \E[\xi(Z_j + 2)] \right\} + \eps^4 \theta^2 \left\{\E[\xi(Z_j + 1)]^2 - \E[\xi(Z_j)] \E[\xi(Z_j + 2)] \right\} + o(\Delta^2)
\end{align*}
Using properties of $\xi$, this is 
\begin{align*}
    &\eps^2 \theta \left\{ 2 \E[\phi(Z_j + 1)^2] - \E[\phi(Z_j)^2] - \E[\phi(Z_j + 2)^2]\right\}\\
    &+ \eps^4 \theta^2 \left\{ \E[\phi(Z_j + 1)^2]^2 - \E[\phi(Z_j)^2]\E[\phi(Z_j + 2)^2]\right\} + o(\Delta^2)
\end{align*}
Simplifying using the definition of $\phi$, and applying Lemma~\ref{lem:theta_phi(Z_j)_o(1)}, this is $$O(\eps^2) + O\left(\left(\frac{n \eps^2}{m} \right)^2 \right)$$
This gives us the first claim. For the second claim, we have the expression 
\begin{align*}
    e^{-2\lam_0 \nu_j} |\PE[\exp(\eps^2 \theta \xi(W_j))]\PE[\exp(\eps^2 \theta \xi(W_j + 1))]| = \E[\exp(\eps^2 \theta \xi(Z_j))]\E[\exp(\eps^2 \theta \xi(Z_j + 1))]
\end{align*}
By properties of $\xi$ from Assumption~\ref{assumption:xi}, and using Lemma~\ref{lem:theta_phi(Z_j)_o(1)}, this is $$1 + O\left(\frac{n \eps^2}{m} \right)$$

\end{proof}

\begin{lemma}
\label{lem:G_second_derivative}
Suppose Assumption~\ref{assumption:huber} holds. For $\lam_0 = n(1 - \eps^2 \theta)$, $\nu_j = \frac{1}{m} + O\left(\frac{\eps}{m} \right)$, and $\xi$ that satisfies Assumption~\ref{assumption:xi}, 

and $$G(\psi) = - i n\psi  + \sum_{j=1}^m \log\left\{ \sum_{k=0}^\infty \frac{(\lam_0 \nu_j e^{i \psi})^k}{k!} \xi(k) \right\}$$

we have

$$G''(\psi) = -n e^{i \psi} + O\left(\frac{n^2 \eps^2}{m} \right)$$

\end{lemma}

\begin{proof}
By Lemmas \ref{lem:theta_phi(Z_j)_o(1)}, \ref{lem:huber_second_derivative_helper_lemmas} and \ref{lem:squared_huber_H_derivatives} we have $$G''(\psi) = \sum_{j=1}^m \frac{(\lam_0 \nu_j e^{i \psi})^2 \left\{O(\eps^2) + O\left(\left(\frac{n \eps^2}{m}\right)^2\right) \right\} - (\lam_0 \nu_j e^{i \psi}) (1 + O(\frac{ n \eps^2}{m}))}{(1 + O(\frac{n \eps^2}{m})) }$$

By Lemma \ref{lem:theta_phi(Z_j)_o(1)}, this is,
$$ \sum_{j=1}^m \left\{ O\left(\frac{n^2 \eps^2}{m^2} \right) - (\lam_0 \nu_j e^{i \psi}) \left(1 + O\left(\frac{n \eps^2}{m}\right)\right)\right\} \left(1 + O\left(\frac{n \eps^2}{m}\right)\right)$$

$$ = - (\lam_0 e^{i \psi}) \left(1  + O\left(\frac{n \eps^2}{m} \right)\right) + O\left(\frac{n^2 \eps^2}{m}\right) = -n e^{i \psi} + O\left(\frac{n^2 \eps^2}{m}\right)$$
\end{proof}

\input{higher_moment_bounds_huber}

%% file: higher_moment_bounds_huber.tex
\begin{lemma}
\label{lem:quadratic_terms_bound}

Suppose Assumption~\ref{assumption:huber} holds. For $\lam_0 = n(1 + O(\eps^2)$ and $\nu$ such that $\nu_j = 1/m + O(\eps/m)$ for all $j$, and $\theta = O(1)$, we have
$$\left|\E \left[ \1_{\{\phi(Z_j + c) \le \beta\}} \left\{ \sum_{l=3}^\infty \frac{(\eps^2 \theta \phi(Z_j + c)^2)^l}{l!}\right \} \right]\right| = o\left(\Delta^2 \right)$$
where $Z_j \sim Poi(\lam_0 \nu_j)$

\end{lemma}

\begin{proof}
We have that 
\begin{align*}
\left|\E \left[ \1_{\{\phi(Z_j + c) \le \beta\}} \left\{ \sum_{l=3}^\infty \frac{(\eps^2 \theta \phi(Z_j + c)^2)^l}{l!}\right \} \right]\right| &\le \left| \E\left[\sum_{l=3}^\infty \frac{(\eps^2 \theta \beta^2)^l}{l!} \right]\right|\\
&\le \sum_{l=3}^\infty \frac{(\eps^2 \theta \beta^2)^l}{l!}\\
&= O\left((\eps^2 \theta \beta^2)^3 \sum_{l=0}^\infty \frac{(\eps^2 \theta \beta^2)^l}{l!}\right)
\end{align*}

Also, $$\sum_{l=0}^\infty \frac{(\eps^2 \theta \beta^2)^l}{l!} = \exp (\eps^2 \theta \beta^2)$$

But, by \eqref{eq:beta_eps^2_little_o_1}, $$\eps^2 \theta \beta^2 = o(1)$$

Thus, we have that $$\exp(\eps^2 \theta \beta^2) = e^{o(1)} = O(1)$$

Putting the above together gives us the claim.
\end{proof}

\begin{lemma}
\label{lem:Poisson_tail_generic}
Suppose Assumption~\ref{assumption:huber} holds. For $\lam = n(1 + O(\eps^2) + O(\frac{n}{m} \eps^2))$ and $\nu$ such that $\nu_j = 1/m + O(\eps/m)$ for all $j$, for any $\beta$,
$$\Pr[\phi(Z_j + c) > \beta] \le 2\exp\left\{- \frac{\Omega(\beta^2)}{O(\frac{n}{m})} \right\} + 2 \exp\left\{- \Omega(\beta) \right\}$$

for integer $0 \le c \le 3$ and any constant $\eta > 0$.
\end{lemma}

\begin{proof}
Note that for the conditions given, $$\lam \nu_j = \frac{n}{m} + O(\frac{n \eps}{m})$$
Now, since $Z_j \sim Poi(\lam \nu_j)$ and $\lam =  n(1 + O(\eps^2) + O(\frac{n}{m} \eps^2))$,
$$\E[\1_{\phi(Z_j + c) > \beta}] = \Pr[\phi(Z_j + c) > \beta] = \Pr\left[\left |Z_j + c - \frac{n}{m} \right| > \beta\right] $$
$$\le \Pr\left[\left |Z_j - \lam \nu_j\right| > \beta + \left |\lam\nu_j + c - \frac{n}{m}\right|\right] \le \Pr\left[\left| Z_j - \lam \nu_j \right| > \beta + O(\frac{n \eps}{m})\right]$$

Using Poisson concentration bounds \citep{Poisson-bounds}, this is at most $$2\exp\left\{- \frac{\left(\beta + O(\frac{n \eps}{m})\right)^2}{\lam \nu_j + \left(\beta + O(\frac{n \eps}{m})\right)}\right\} = 2\exp\left\{- \frac{\left(\beta + O(\frac{n \eps}{m})\right)^2}{\frac{n}{m} + \beta + O(\frac{n \eps}{m})}\right\}$$

Now, if $\beta = O(\frac{n}{m})$, this is $$2 \exp\left \{- \frac{(\beta + O(\frac{n \eps}{m}))^2}{O(\frac{n}{m}) + O(\frac{n \eps}{m})}\right\} = 2 \exp\left\{-\frac{\Omega(\beta^2)}{O(\frac{n}{m})}\right\} $$


Similarly, if instead $\frac{n}{m} = O(\beta)$, the bound is $$2 \exp\left\{-\frac{(\beta + O(\frac{n \eps}{m}))^2}{O(\beta) + O(\frac{n \eps}{m})}\right \} = 2 \exp\left\{- \Omega(\beta)\right \}$$

The claim follows.


\end{proof}

\begin{lemma}
 \label{lem:huber_indicator_phi_greater_than_beta_O(eps_12)}
Suppose Assumption~\ref{assumption:huber} holds. For $\lam = n(1 + O(\eps^2) + O(\frac{n}{m} \eps^2))$ and $\nu$ such that $\nu_j = 1/m + O(\eps/m)$ for all $j$, for $\beta$ such that \eqref{eq:beta_second_bound} is satisfied,
$$\E[\1_{\phi(Z_j + c) > \beta}] =O\left(\Delta^{2 \eta}\right)$$

for integer $0 \le c \le 3$ and any constant $\eta > 0$.
\end{lemma}

\begin{proof}

By Lemma \ref{lem:Poisson_tail_generic}, $$\E[\1_{\{\phi(Z_j + c) > \beta \}}] \le 2 \exp\left\{- \frac{\Omega(\beta^2)}{O(\frac{n}{m})}\right\} + 2 \exp \left\{-\Omega(\beta) \right\}$$

But by \eqref{eq:beta_second_bound}, this is $$4 \exp\left\{- \omega \left(\log \left(\frac{1}{\Delta}\right) \right) \right\} = O\left(\Delta^{2 \eta}\right)$$

\end{proof}

\begin{lemma}
\label{lem:huber_tail_moment_bound}
Suppose Assumption~\ref{assumption:huber} holds. For $\lam = n(1 + O(\eps^2))$, $\nu_j = 1/m + O(\eps/m)$ for all $j$, we have

$$\left| \E\left [ \1_{\{\phi(Z_j + c) > \beta\}} \eps^2 \theta \phi(Z_j+c)^2\right]\right| = o\left(\Delta^2 \right)$$

$$\left| \E\left [ \1_{\{\phi(Z_j + c) > \beta\}} \frac{\eps^4 \theta^2}{2} \phi(Z_j+c)^4 \right]\right| = o\left(\Delta^2 \right)$$

for integer $0 < c \le 3$ and $Z_j \sim \Poi(\lam\nu_j)$
\end{lemma}

\begin{proof}

We have $$\left|\E\left [ \1_{\{\phi(Z_j + c) > \beta\}} \eps^2 \theta \phi(Z_j+c)^2\right]\right| \le \sum_{l=0}^\infty \left| \Pr\left[ \phi(Z_j + c) > 2^l \beta\right] \eps^2 \theta(2^{l+1} \beta)^2 \right|$$

By Lemma \ref{lem:Poisson_tail_generic}, this is $$\sum_{l=0}^\infty \left| \left(2 \exp\left\{- \frac{\Omega(2^{2l}\beta^2)}{O(\frac{n}{m})} \right\} + 2 \exp\left\{- \Omega(2^l \beta) \right\}\right) \eps^2 \theta (2^{l+1} \beta)^2 \right|$$

By \eqref{eq:beta_second_bound}, this is $$\sum_{l=0}^\infty \left|O\left(\Delta^{2^l \eta}\right) \eps^2 \theta (2^{l+1} \beta)^2 \right|$$

Note that by \eqref{eq:beta_third_bound}, $\eps^2 \beta^2 = o(1)$. Thus, this is $$\sum_{l=0}^\infty \left|O\left(\Delta^{2^l \eta}\right) o(1) 2^{2(l+1)}\right|$$

Setting $\eta \ge 3$, since $2^{2(l+1)} \le O(2^{2^l})$ we have that this is this is $$O\left(\Delta^3 \right) \sum_{l=0}^\infty \left|o(1)^{2^l}\right| = O\left(\Delta^3 \right)$$ 

The first claim follows. The second claim can be proved in a similar way.

\end{proof}

\begin{lemma}
\label{lem:huber_linear_mgf_bound}
Suppose Assumption~\ref{assumption:huber} holds. For $\lam = n(1 + O(\eps^2) + O(\frac{n}{m} \eps^2))$ and $\nu$ such that $\nu_j = 1/m + O(\eps/m)$ for all $j$,
$$E\left[ \exp( \eps^2 \theta \beta (2\phi(Z_j + c) - \beta))\right] = O(1)$$
for integer $0 \le c \le 3$
\end{lemma}

\begin{proof}
$$\exp(\eps^2 \theta \beta (2\phi(k) - \beta)) = \exp\left\{\eps^2 \theta \beta \left(2\left|k - \frac{n}{m} \right| - \beta \right) \right\} $$

$$= \exp\left \{\eps^2 \theta \beta \left( \1_{\{k \le \frac{n}{m}\}} 2 \left(\frac{n}{m} - k\right) + \1_{\{k > \frac{n}{m} \}} 2 \left(k - \frac{n}{m} \right) - \beta \right) \right\}  $$
$$ = \1_{\{k \le \frac{n}{m}\}}\exp\left\{\eps^2 \theta \beta\left( 2 \left( \frac{n}{m} - k\right) - \beta \right) \right\} + \1_{\{k > \frac{n}{m}\}} \exp\{\left(2 \theta \beta \left( 2 \left(k - \frac{n}{m} \right) - \beta \right) \right\}$$
Since $\exp(x) \ge 0$ for all $x$, this is at most
$$\exp\left\{\eps^2 \theta \beta\left( 2\left( \frac{n}{m} - k\right) - \beta \right) \right\} + \exp\left\{\eps^2 \theta \beta \left( 2\left(k - \frac{n}{m} \right) - \beta \right) \right\}$$

Thus, 
\begin{dmath*}
\E\left[ \exp(\eps^2 \theta \beta (2\phi(Z_j + c) - \beta))\right] \le \E\left[\exp\left\{\eps^2 \theta \beta\left( 2 \left( \frac{n}{m} - Z_j - c\right) - \beta \right) \right\}\right] + \E\left[\exp\left\{\eps^2 \theta \beta \left(2 \left(Z_j + c - \frac{n}{m} \right) - \beta \right) \right\}\right]\label{eq:huber_lemma_exp_split_bound}
\end{dmath*}

Now, 
\begin{equation}
E\left[\exp\left\{\eps^2 \theta \beta\left( 2\left( \frac{n}{m} - Z_j - c\right) - \beta \right) \right\}\right] = \exp\left\{\eps^2 \theta \beta \left(\frac{2n}{m} - 2c - \beta\right) \right\} \E\left[\exp\left\{-2 \eps^2 \theta \beta Z_j\right\}\right]\label{eq:huber_lemma_exp(2thetabeta)}
\end{equation}

But, since $Z_j \sim Poi(\lam \nu_j)$, we have $$\E[\exp\{-2 \eps^2 \theta \beta Z_j\}] = \exp\left\{\lam \nu_j \left(e^{-2 \eps^2 \theta \beta} - 1\right) \right\}$$

Note that $$\lam \nu_j = \frac{n}{m} + O(\frac{n \eps}{m})$$
So, $$\E[\exp\{-2\eps^2 \theta \beta Z_j\}] = \exp\left\{\frac{n}{m}(1+O(\eps)) \left(e^{-2 \eps^2 \theta \beta} - 1\right) \right\}$$

Using the fact that $e^x \le 1 + x$ for all $x$, the above is at most $$\exp\left\{\frac{n}{m} (1 + O(\eps)) (-2 \eps^2 \theta \beta)\right\} = \exp\left\{-2 \eps^2 \theta \beta \frac{n}{m} + O(\frac{n}{m}\eps \theta \beta) \right \}$$

Since $n = o(m/\eps^2)$, and $\theta = \Theta(\eps^2)$,
$$O\left(\frac{n}{m} \eps \theta \beta \right) = o\left(\beta \eps\right)$$

Using \eqref{eq:beta_eps^2_little_o_1}, this is $o(1)$.

Thus, $$\E[\exp\{-2 \eps^2 \theta \beta Z_j\}] = \exp\left\{-2 \eps^2 \theta \beta \frac{n}{m} + o(1)\right\}$$

Thus, using this in \eqref{eq:huber_lemma_exp(2thetabeta)}, we have that $$\E\left[\exp\left\{\eps^2 \theta \beta\left( 2\left( \frac{n}{m} - Z_j - c\right) - \beta \right) \right\}\right] = \exp\left\{- \eps^2 \theta \beta^2 -2c \eps^2 \theta \beta + o(1)\right\}$$

Equation \eqref{eq:beta_eps^2_little_o_1} tells us that $$\eps^2 \theta \beta^2 = o(1)$$

Similarly, $$\eps^2\theta \beta = o(\eps)$$

So, $$\E\left[\exp\left\{\eps^2 \theta \beta\left( 2\left( \frac{n}{m} - Z_j - c\right) - \beta \right) \right\}\right] = \exp\left\{o(1)\right\} = O(1)$$

Using a very similar argument, we can show that $$\E\left[\exp\left\{\eps^2 \theta \beta \left(2 \left(Z_j + c - \frac{n}{m} \right) - \beta \right) \right\}\right] = O(1)$$.

So, by the above and \eqref{eq:huber_lemma_exp_split_bound}, 
$$\E\left[ \exp(\eps^2 \theta \beta (2\phi(Z_j+c) - \beta))\right] = O(1)$$
\end{proof}

\begin{lemma}
\label{lem:huber_linear_terms_bound}
Suppose Assumption~\ref{assumption:huber} holds. For $\lam = n(1 + O(\eps^2) + O(\frac{n}{m} \eps^2))$ and $\nu$ such that $\nu_j = 1/m + O(\eps/m)$ for all $j$,
$$\E\left[\1_{\{\phi(Z_j+c) > \beta\}} \exp \left\{\eps^2 \theta \beta(2 \phi(Z_j+c) - \beta)\right\}\right] = O\left(\Delta^\eta\right)$$
with integer $0 \le c \le 3$ and any constant $\eta > 0$.
\end{lemma}

\begin{proof}
By the Cauchy-Schwarz inequality,
\begin{dmath*}
\E\left[\1_{\{\phi(Z_j+c) > \beta\}} \exp \left\{ \eps^2 \theta \beta(2 \phi(Z_j+c) - \beta)\right\}\right] \le \sqrt{\E[\1_{\{\phi(Z_j+c) > \beta \}}] \E\left[\exp \left\{\eps^2 \theta \beta(2 \phi(Z_j+c) - \beta)\right\}\right]}
\end{dmath*}

By Lemmas \ref{lem:huber_indicator_phi_greater_than_beta_O(eps_12)} and \ref{lem:huber_linear_mgf_bound}, this is $O\left(\Delta^\eta\right)$.
\end{proof}

%% file: lemmas.tex
\subsection{General Lemmas}
\begin{lemma}
\label{lem:phi(Z_j)_expansions}
For $Z_j \sim Poi(\lam \nu_j)$, and $\phi$ defined in \eqref{eq:centering_def}
\begin{align*}
\E[\phi(Z_j)^2] &= (\lam \nu_j)^2 + \lam \nu_j - 2 \lam \nu_j \frac{n}{m} + \frac{n^2}{m^2}\\
\E[\phi(Z_j)^4] &= (\lam \nu_j)^4 + 6 (\lam \nu_j)^3 + 7 (\lam \nu_j)^2 + \lam \nu_j - 4 \frac{n}{m}\left [(\lam \nu_j)^3 + 3(\lam \nu_j)^2 + \lam \nu_j\right]\\
&+ 6 \left(\frac{n}{m} \right)^2 \left[(\lam \nu_j)^2 + \lam \nu_j\right] - 4 \left(\frac{n}{m}\right)^3 \lam \nu_j + \left(\frac{n}{m} \right)^4
\end{align*}

\end{lemma}

\begin{proof}
$$\E[\phi(Z_j)^2] = \E\left[\left( Z_j - \frac{n}{m}\right)^2\right] = \E\left[Z_j^2 - 2 Z_j \frac{n}{m} + \frac{n^2}{m^2}\right]$$
$$=(\lam \nu_j)^2 + \lam \nu_j - 2 \lam \nu_j \frac{n}{m} + \frac{n^2}{m^2}$$

\begin{align*}
\E[\phi(Z_j)^4] &= \E\left[\left(Z_j - \frac{n}{m} \right)^4\right] = \E\left[Z_j^4 - 4 \frac{n}{m} Z_j^3 + 6 \left(\frac{n}{m}\right)^2 Z_j^2 - 4\left(\frac{n}{m}\right)^3 Z_j + \left(\frac{n}{m} \right)^4 \right]\\
&= (\lam \nu_j)^4 + 6 (\lam \nu_j)^3 + 7 (\lam \nu_j)^2 + \lam \nu_j - 4 \frac{n}{m}\left [(\lam \nu_j)^3 + 3(\lam \nu_j)^2 + \lam \nu_j\right]\\
&+ 6 \left(\frac{n}{m} \right)^2 \left[(\lam \nu_j)^2 + \lam \nu_j\right] - 4 \left(\frac{n}{m}\right)^3 \lam \nu_j + \left(\frac{n}{m} \right)^4
\end{align*}
\end{proof}

\begin{lemma}
For $Z_j \sim Poi(\lam \nu_j)$, and $\phi$ defined in \eqref{eq:centering_def}

\begin{align*}
\E\left[\phi(Z_j + 1)^2 - \phi(Z_j)^2\right]
&= 2 \lam \nu_j + 1 - \frac{2n}{m}\\
\E\left[\phi(Z_j+1)^4 - \phi(Z_j)^4\right] &= 4(\lam \nu_j)^3 + 18(\lam \nu_j)^2 + 14(\lam \nu_j) + 1 \\
&- \frac{4n}{m}(3(\lam \nu_j)^2 + 6\lam \nu_j + 1) + \frac{6 n^2}{m^2} (2 \lam \nu_j + 1) - \frac{4 n^3}{m^3}
\end{align*}
\end{lemma}
\begin{proof}
\begin{align*}
\E[\phi(Z_j+1)^2 - \phi(Z_j)^2] = \E\left[2 Z_j + 1 - \frac{2n}{m} \right] =  2 \lam \nu_j + 1 - \frac{2n}{m}
\end{align*}

\begin{align*}
&\E[\phi(Z_j + 1)^4 - \phi(Z_j)^4]\\
&= \E\left[(4 Z_j^3 + 6 Z_j^2 + 4 Z_j + 1) - \frac{4n}{m} (3 Z_J^2 + 3 Z_j + 1) + 6 \frac{n^2}{m^2} (2Z_j + 1) - 4 \frac{n^3}{m^3}\right]\\
&= 4(\lam \nu_j)^3 + 18(\lam \nu_j)^2 + 14(\lam \nu_j) + 1 - \frac{4n}{m}(3(\lam \nu_j)^2 + 6\lam \nu_j + 1) + \frac{6 n^2}{m^2} (2 \lam \nu_j + 1) - \frac{4 n^3}{m^3}
\end{align*}
\end{proof}

\begin{lemma}
\label{lem:poisson_moments}
The first four moments of the Poisson distribution are given by
$$\E[X] = \lam$$
$$\E[X^2] = \lam^2 + \lam$$
$$\E[X^3] = \lam^3 + 3 \lam^2 + \lam$$
$$\E[X^4] = \lam^4 + 6 \lam^3 + 7 \lam^2 + \lam$$

for $X \sim Poi(\lam)$
\end{lemma}
\begin{proof}
  Computation of moments.
\end{proof}

%

\begin{lemma}
\label{lem:squared_huber_H_derivatives}
For any function $f$ and
$$G(\psi) = - i n\psi  + \sum_{j=1}^m \log\left\{ \sum_{k=0}^\infty \left[ \frac{(\lam_0 \nu_j e^{i \psi})^k}{k!} f(k) \right]\right\}$$

we have

$$ G'(\psi) = - i n + i \sum_{j=1}^m (\lam_0 \nu_j e^{i \psi}) \frac{\sum_{k=0}^\infty \frac{(\lam_0 \nu_j e^{i \psi})^k}{k!} f(k+1)}{\sum_{k=0}^\infty \frac{(\lam_0 \nu_j e^{i \psi})^k}{k!} f(k)}$$

so that $Re(G'(0)) = 0$, and 
\begin{align*}
G''(\psi) = &\sum_{j=1}^m\left\{ \frac{1}{\left(\sum_{k=0}^\infty \frac{(\lam_0 \nu_j e^{i \psi})^k}{k!} f(k) \right)^2}\left[ (\lam_0 \nu_j e^{i \psi})^2 \left\{\left(\frac{ \sum_{k=0}^\infty  (\lam_0 \nu_j e^{i \psi})^k}{k!} f(k+1)\right)^2 \right. \right. \right.\\
&\left. \left. \left. - \left(\sum_{k=0}^\infty \frac{(\lam_0 \nu_j e^{i \psi})^k}{k!} f(k) \right) \left( \sum_{k=0}^\infty \frac{(\lam_0 \nu_j e^{i \psi})^k}{k!} f(k+2)\right)\right\}\right. \right. \\
&\left. \left. - (\lam_0 \nu_j e^{i \psi}) \left( \sum_{k=0}^\infty \frac{(\lam_0 \nu_j e^{i \psi})^k}{k!} f(k) \right) \left(\sum_{k=0}^\infty \frac{(\lam_0 \nu_j e^{i \psi})^k}{k!} f(k+1) \right) \right]\right\}
\end{align*}
\end{lemma}

\begin{proof}
Follows from taking derivatives.
\end{proof}

%% file: lower_bounds.tex
\section{Lower bounds}\label{appendix:lower_bounds}
\input{counterexample.tex}
\input{paninski_lower_bound}

%% file: counterexample.tex
\subsection{The collisions tester is asymptotically bad when $n = \Theta(m)$ and $\eps = \omega \left( \frac{\log^{1/4} n}{n^{1/8}} \right)$}\label{app:peebles}

The following lower bound showing the collisions tester is
asymptotically suboptimal is based on a note of Peebles~\citep{peebles}.

\begin{theorem}\label{thm:peebles}
When $n = \Theta(m)$ and $\eps = \omega\left(\frac{\log^{1/4}n}{n^{1/8}}\right)$, the collisions tester has error probability $\exp\left(-o\left(\frac{n^2 \eps^4}{m}\right) \right)$, and so, takes $\omega(\sqrt{m\log(1/\delta)} / \eps^2)$ samples to distinguish between the uniform distribution and an $\eps$-far distribution with error probability $\delta$.
\end{theorem}

\begin{restatable}[Due to Peebles.]{thm}{peebles}
When $n = \Theta(m)$ and $\eps = \omega\left(\frac{\log^{1/4}n}{n^{1/8}}\right)$, the collisions tester has error probability $\exp\left(-o\left(\frac{n^2 \eps^4}{m}\right) \right)$, and so, takes $\omega(\sqrt{m\log(1/\delta)} / \eps^2)$ samples to distinguish between the uniform distribution and an $\eps$-far distribution with error probability $\delta$.
\end{restatable}
\begin{proof}
First, let $X_1, \dots, X_n$ be the elements sampled from our distribution. Let $E_{i, j}$ be the event that $X_i = X_j$. Under the uniform distribution $p$, we have that the probability that $X_i = X_j$ is 
$$ \Pr_p[E_{i, j}] = \sum_{j=1}^m p_j^2 = \frac{1}{m}$$

Thus, the expected number of collisions under the uniform distribution is $$\E_p\left[\sum_{i < j} E_{i, j}\right] = \sum_{i < j} \E_p[E_{i, j}] = \binom{n}{2}/m$$

Now, consider the $\eps$-far distribution $q$ such that $q_j = \frac{1}{m} + \frac{2 \eps}{m}$ for $j \le m/2$ and $q_j = \frac{1}{m} - \frac{2 \eps}{m}$ for $j > m/2$. We have that the probability that $X_ i = X_j$ under $q$ is $$\Pr_q[E_{i, j}] = \sum_{j=1}^m q_j^2 = \frac{1}{m} + \sum_{j=1}^m \left( q_j - \frac{1}{m}\right)^2 =\frac{1 + 4 \eps^2}{m}$$

Thus, the expected number of collisions under $q$ is $$\E_q\left[\sum_{i < j} E_{i, j}\right] = \sum_{i < j}\E[E_{i, j}] = \binom{n}{2} \frac{1 + 4 \eps^2}{m}$$

Now, under $p$, if we sample the first element at least $4n/\sqrt{m}$ times, then we will have at least $\binom{4n/\sqrt{m}}{2} > \binom{n}{2} \frac{1 + 4 \eps^2}{m}$ collisions for large enough $n, m$. In this case, since the number of collisions under $p$ is more than the expected number of collisions under $q$, and our threshold will be less than the expected number of collisions under $q$, we will output $q$, and make a mistake. This happens with probability at least $$\frac{1}{n^{4n/\sqrt{m}}} = \exp\left(-\frac{4n}{\sqrt{m}} \log n\right)$$

This is bigger than $\exp(-\Omega(\frac{n^2 \eps^4}{m}))$, the error probability of the TV tester, as long as $$\frac{4n}{\sqrt{m}} \log n = o\left( \frac{n^2 \eps^4}{m}\right)$$

Since $n = \Theta(m)$, this happens as long as $$\eps = \omega\left(\frac{\log^{1/4}n}{n^{1/8}} \right)$$

Thus, for error probability $\delta$, we require $\omega(\sqrt{m \log(1/\delta)}/\eps^2)$ samples in the regime stated.
\end{proof}

%% file: paninski_lower_bound.tex
\subsection{Paninski tester is asymptotically bad when $n \ge \Theta(m \log m)$}
\begin{restatable}{thm}{paninski}\label{thm:paninski}
When $n \ge 48 m \log m$, the Paninski tester fails to distinguish between the uniform distribution on $m$ and an $\eps$-far distribution with error probability $\Omega(1)$ for $\eps < 1/3$.
\end{restatable}

\begin{proof}
Recall that the Paninski tester counts the number of bins that see exactly one sample. Let $E_j = \1_{\{Y_j = 1\}}$ be the event that the bin $j$ has exactly $1$ sample. Now, for $p$ the uniform distribution on $[m]$, the expected number of samples that land in the $j^{th}$ bin is $$\E_p[Y_j] = \frac{n}{m} = 48\log m$$
Thus, by the Bernstein's inequality,
$$\Pr_p[Y_j \le \log m] = \Pr_p\left[Y_j  \le \left(1 - \frac{47}{48}\right) \E_p[Y_j]\right] \le e^{-\left(\frac{47}{48}\right)^2 \times 12 \log m} = e^{-3 \log m} \le \frac{1}{m^3}$$

So, by union bound, $$\Pr_p [\exists j, Y_j \le  \log m] \le \sum_{j = 1}^{m} \Pr_p\left[Y_j \le  \log m\right] \le \frac{1}{m^2}$$

So, with probability $1 - 1/m^2$, every bin has at least $ \log m$ balls, which means that the Paninski statistic is $0$ with probability $1 - 1/m^2$.

Now, under $\eps$-far distribution $q$ such that $q_j = \frac{1}{m} + \frac{2 \eps}{m}$ for $j \le m/2$ and $q_j = \frac{1}{m} - \frac{2 \eps}{m}$ for $j > m/2$, we have, for $j \le m/2$

$$\E_q[Y_j] = \frac{n}{m} = (1 + 2 \eps) 48\log m$$

and for $j > m/2$,

$$\E_q[Y_j] = \frac{n}{m} = (1 - 2 \eps) 48 \log m$$

Then, for $j \le m/2$, since $\eps > 0$, $$\Pr_q[Y_j \le \log m] = \Pr_q\left[Y_j \le \left(1 - \frac{47 + 96 \eps}{48 + 96\eps}\right) \E_q[Y_j] \right] \le \Pr_q\left[Y_j \le \left(1 - \frac{47}{48}\right) \E_q[Y_j] \right] \le \frac{1}{m^3}$$

On the other hand, for $j > m/2$
$$\Pr_q[Y_j \le \log m] = \Pr_q[Y_j \le \left(1 - \frac{47 - 96 \eps}{48(1 - 2 \eps)} \E_q[Y_j] \right) \le e^{-\left(\frac{47 - 96 \eps}{48(1 - 2\eps)}\right)^2 \times (1 - 2 \eps) 12 \log m}$$

Since $\eps < 1/3$, this is at most $$e^{-\frac{15^2 \times 3 }{48^2} \times 12 \log m} \le \frac{1}{m^3}$$

Thus, we have that $$\Pr_q[\exists j, Y_j \le \log m] \le \sum_{j=1}^m \Pr_q[Y_j \le \log m] \le \frac{1}{m^2}$$

So again, with probability $1 - 1/m^2$, every bin has at least $\log m$ balls under $q$, so that the Paninski statistic is $0$ with $1 - \frac{1}{m^2}$. Putting the above together gives us that we will fail with probability $\Omega(1)$.

\end{proof}

%% file: tv_tester_superlinear.tex
\section{TV Tester in the Superlinear Regime}\label{app:tv_superlinear}

\begin{lemma}
\label{lem:tv_tester_optimal_n=omega(m_over_eps_sq)}
The TV tester can distinguish between the uniform distribution on $[m]$ and an $\eps$-far distribution with failure probability $e^{- \frac{1}{2} \eps^2 n(1+o(1))}$ when $\eps = o(1)$ and $n = \omega(m/\eps^2)$.
\end{lemma}

\begin{proof}
Let $Y$ be the empirical distribution when $n$ samples are drawn from the distribution.
The TV tester compares the empirical distribution $Y$ to the uniform distribution and outputs uniform if and only if $$\|Y - p\|_{TV} < \frac{\eps}{2}$$
where $p$ is the uniform distribution on $[m]$. 

Since by the definition of TV distance, $$\|Y - p\|_{TV} = \max_{S \subseteq [m]} \left| Y_S - p_S \right|$$
we have that under the uniform distribution $p$, the probability of failure is $$\Pr_p[\|Y - p\|_{TV} \ge \tau^*] = \Pr_p\left[\max_{S \subseteq[m]} |Y_S - p_S| \ge \frac{\eps}{2} \right]$$

$$\le \sum_{S \subseteq[m]} \Pr_p[|Y_S - p_S| \ge \frac{\eps}{2}]$$

Note that the summand is the probability that the empirical mean of $n$ samples from a coin with probability of heads $\frac{|S|}{m}$ deviates from its expectation by at least $\frac{\eps}{2}$. By the Chernoff bound, this is at most $$e^{-D\left( \frac{|S|}{m} + \frac{\eps}{2} \| \frac{|S|}{m} \right) n} \le e^{- \frac{1}{2} \eps^2 n}$$

since for any $r, \tau$, $D(r + \tau \| r) \ge 2 \tau^2$.

Thus, $$\Pr_p\left[\|Y - p \|_{TV} \ge \frac{\eps}{2} \right] \le 2^m e^{-\frac{1}{2} \eps^2 n} = e^{- \frac{1}{2} \eps^2 n(1 + o(1))}$$ since $n = \omega(m/\eps^2)$.

Now, under $\eps$-far distribution $q$, since $\|p - q\|_{TV} \ge \eps$, the set $S = \{j | q_j > p_j\}$ has $|q_S - p_S| \ge \eps$ so that $\sum_{j \in S} q_j \ge \frac{|S|}{m} + \eps$. Now, since $\|Y - p\|_{TV} \ge |Y_S - p_S|$, we have $$\Pr_q\left[\|Y - p\|_{TV} \le \frac{\eps}{2}\right] \le  \Pr_q\left[|Y_S - p_S| \le \frac{\eps}{2} \right]$$

Now, the RHS of the above is at most the probability that the empirical mean of $n$ samples from a coin with probability of heads at least $\frac{1}{2} + \eps$ is less than $\frac{1}{2} + \frac{\eps}{2}$. By the Chernoff bound, this is at most $$e^{-D(\frac{1}{2} + \frac{\eps}{2} \| \frac{1}{2} + \eps) n} \le e^{-\frac{1}{2} n \eps^2}$$

Thus, the TV tester fails with probability at most $e^{-\frac{1}{2} n \eps^2 (1 + o(1))}$ as required.
\end{proof}

\begin{lemma}
\label{lem:n=omega(m_over_eps_sq)_lower_bound}
When $n = \omega\left(\frac{m}{\eps^2}\right)$ and $\eps = o(1)$, any tester that distinguishes between the uniform distribution on $[m]$ and an $\eps$-far distribution fails with probability at least $e^{-\frac{1}{2} n \eps^2(1+o(1))}$.
\end{lemma}

\begin{proof}
Let $p$ be the uniform distribution on $[m]$, and let $q$ be the $\eps$-far distribution such that $q_j = \frac{1}{m} + \frac{2\eps}{m}$ for $j \le m/2$ and $q_j = \frac{1}{m} - \frac{2 \eps}{m}$ for $j > m/2$. Let $Y$ be the empirical distribution from the samples drawn, and let $B = \{\sum_{j=1}^{m/2} Y_j \ge \frac{1}{2} + \frac{\eps}{2} \}$. Under the uniform distribution, by Lemma \ref{lem:binomial_anticoncentration},

\begin{equation}\label{eq:method_of_types_lower_bound}
\Pr_p[B] \ge e^{-\frac{1}{2} \eps^2 n(1 + o(1))}
\end{equation}

Now, the likelihood ratio of $Y \in B$ is given by $$\frac{q}{p}(Y) \ge (1 + 2 \eps)^{(\frac{1}{2} + \frac{\eps}{2})n} (1 - 2\eps)^{(\frac{1}{2} -\frac{\eps}{2})n}$$
$$ = e^{\left(\frac{1}{2} + \frac{\eps}{2} \right) \log\left(1 + 2 \eps\right)n + \left(\frac{1}{2} - \frac{\eps}{2} \right) \log\left(1 - 2\eps\right)n}$$

Now, 
\begin{align*}
  \left(\frac{1}{2} + \frac{\eps}{2} \right) \log\left(1 + 2 \eps\right) + \left(\frac{1}{2} - \frac{\eps}{2} \right) \log\left(1 - 2\eps\right)
  &= \left(\frac{1}{2} + \frac{\eps}{2} \right) \left\{ \log\left(\frac{\frac{1}{2} + \frac{\eps}{2}}{\frac{1}{2}} \right) - \log \left(\frac{\frac{1}{2} + \frac{\eps}{2}}{\frac{1}{2} + \eps} \right)\right\}\\
  &\qquad+ \left(\frac{1}{2} - \frac{\eps}{2}\right) \left\{ \log \left(\frac{\frac{1}{2} - \frac{\eps}{2}}{\frac{1}{2}} \right) - \log \left(\frac{\frac{1}{2} - \frac{\eps}{2}}{\frac{1}{2} - \eps}\right) \right\}\\
  &= D\left(\frac{1}{2} + \frac{\eps}{2} \| \frac{1}{2} \right) - D\left(\frac{1}{2} + \frac{\eps}{2}\| \frac{1}{2} + \eps \right)\\
    &= o(\eps^2)
\end{align*}
since $D(\frac{1}{2} + \frac{\eps}{2} \| \frac{1}{2}) = \frac{\eps^2}{2}(1 + o(1))$ and $D(\frac{1}{2} + \frac{\eps}{2} \| \frac{1}{2} + \eps) = \frac{\eps^2}{2} (1 + o(1))$.
Thus,
\begin{equation}\label{eq:likelihood_ratio_testing}
\frac{q}{p}(X) \ge  e^{o(\eps^2)}
\end{equation}


Now, suppose there is a test $\phi$ that uses $n$ samples and distinguishes between the two cases with failure probability $e^{-\left(\frac{1}{2} + \xi\right) n \eps^2}$ for $\xi > 0$, so that $\phi = 0$ denotes that the test outputs uniform, and $\phi=1$ denotes that the test outputs far from uniform.

By assumption, $$\Pr_p[\phi = 1] \le e^{-\left(\frac{1}{2} + \xi \right)n \eps^2}$$

Now, $$\Pr_q[\phi = 0] \ge \Pr_q[\{\phi = 0\} \cap B] \ge \frac{q}{p} (\{\phi = 0\} \cap B) \Pr_p[\{\phi = 0\} \cap B]$$

But by \eqref{eq:method_of_types_lower_bound} and our assumption, $$\Pr_p[\{\phi = 0\} \cap B] \ge \Pr_p[B] - \Pr_p[\phi = 1] \ge e^{-\frac{1}{2} \eps^2 n (1 + o(1))}$$

so that, using \eqref{eq:likelihood_ratio_testing} $$\Pr_q[\phi = 0] \ge e^{-\frac{1}{2} \eps^2 n (1 + o(1)) }$$

Thus, $\phi$ has error probability at least $e^{-\frac{1}{2} \eps^2 n(1 + o(1))}$, and we have a contradiction, so that any test fails with probability at least $e^{-\frac{1}{2} n \eps^2 (1 + o(1))}$.
\end{proof}

\superlinear*

\begin{proof}
By Lemmas \ref{lem:tv_tester_optimal_n=omega(m_over_eps_sq)} and \ref{lem:n=omega(m_over_eps_sq)_lower_bound}, the claim follows.
\end{proof}

\subsection{Anticoncentration of a Binomial Variable}\label{subsection:anticoncentration}

In this section we show that the Chernoff bound has sharp constants
for binomial random variables $B(n, \frac{1}{2})$.

We use the following lower bound on binomial coefficients, found in
Lemma 4.7.1 of~\citep{AshInfoTheory}:
\begin{lemma}\label{lem:binomialvalue}
  If $n$ and $np$ are integers, then
  \[
    \binom{n}{p} \geq \frac{1}{\sqrt{8np(1-p)}} 2^{n h(p)}
  \]
  where $h(p) := -p \log_2 p - (1-p) \log_2 (1-p)$ is the binary entropy function.
\end{lemma}

\state{lem:binomial_anticoncentration}

\begin{proof}
  The upper bound is the standard Chernoff bound, so we focus on the
  lower bound.  For any $\eps' \in [\eps, \eps + 3/\sqrt{n}]$ with
  $\frac{n}{2}+\eps'n$ integral we have by Lemma~\ref{lem:binomialvalue} that
  \begin{align*}
    \Pr[X = \frac{n}{2} + \eps' n] &= 2^{-n} \binom{n}{(1/2 + \eps') n}\\
    &\geq \frac{1}{\sqrt{2n(1+o(1))}} 2^{-n(1-h(\frac{1}{2} + \eps'))}.
  \end{align*}
  Now, the binary entropy function has Taylor series
  \begin{align*}
    h(\frac{1}{2} + \eps') &= 1 - \frac{2}{\ln 2} (\eps')^2 + O((\eps')^4).
  \end{align*}
  Since $\eps' = \eps(1 + o(1))$ by construction, this means
  \[
    \Pr[X = \frac{n}{2} + \eps' n] \geq \frac{1}{\sqrt{2n(1+o(1))}}  e^{-2n \eps^2 (1 + o(1))}.
  \]
  Summing over the $3\sqrt{n}$ such $\eps'$, we have
  \[
    \Pr[X \geq \frac{n}{2} + \eps n] \geq e^{-2n \eps^2 (1 + o(1))}
  \]
  as desired.
\end{proof}

%% file: squared_statistic.tex
\section{Squared Statistic in Sublinear Regime}\label{section:squared}
Define a centering function 
\begin{equation}
\phi(k):= \left|k - \frac{n}{m} \right|\tag{\ref{eq:centering_def}}
\end{equation}
We will analyze the squared statistic $S$ given by $$S = \sum_{j=1}^m \phi(Y_j^n)^2$$
where $Y_j^n = \sum_{i=1}^n 1_{X_i = j}$ and $X_1, \dots, X_n$ are the $n$ samples drawn from distribution $\nu$ supported on $[m]$. Note that this is equivalent to the collisions statistic, since it simply applies  a translation and scaling.
We will set 
\begin{equation}
\label{eq:beta_setting_squared}
\beta := \kappa \frac{n^2 \eps^4}{m}
\end{equation}
for constant $\kappa > 0$ to be set later. We also define a parameter $\Delta$.
\begin{assumption}\label{assumption:squared}
 $1 \lesssim n/m \ll \frac{1}{\eps^2}$, $\eps \ll \frac{1}{n^{3/13}}$, $\frac{n^2}{m} \eps^4 \gg \log m$ and $m \ge C \log n$ for sufficiently large constant $C$. In addition, we have the following constraints on $\beta$ and $\Delta$.
\begin{equation}
\beta = \omega\left( \log\left(\frac{1}{\Delta}\right) + \sqrt{\frac{n}{m} \log\left(\frac{1}{\Delta}\right)} \right) \tag{\ref{eq:beta_second_bound}}
\end{equation}
\begin{align}
  \Delta = O\left(\frac{n \eps^2}{m} \right)\tag{\ref{eq:Delta_constraint}}
\end{align}
\begin{equation}
(\beta^2 \eps^2)^3 = o\left(\Delta^2 \right)\tag{\ref{eq:beta_third_bound}}
\end{equation}
\end{assumption}
We assume that Assumption~\ref{assumption:squared} holds throughout this section. Observe that Assumption~\ref{assumption:squared} implies Assumption~\ref{assumption:huber}. Note that since $n = \Theta(m)$ and $\frac{n^2 \eps^4}{m} = \omega(\log m)$, we have that 
\begin{equation}
\label{eq:squared_eps_larger_than}
\eps = \omega\left(\frac{\log m}{n^{1/4}}\right)
\end{equation}

For ease of exposition, we will analyze
\begin{equation}
    \Tilde{S} = \frac{m}{n^2 \eps^2}[S - n]
\end{equation}
We will study the log MGF of this statistic conditioned on all $\phi(Y_j^n)$'s at most $\beta$, given by
\begin{equation}
\Lambda_{n, \nu} := \log\left(\E_\nu\left[\exp(\theta \Tilde{S}) | \forall j, \phi(Y_j^n) \le \beta\right]\right)
\end{equation}
We will compute an asymptotic expansion of the limiting log MGF conditioned on all $\phi(Y_j^n)$ at most $\beta$ given by 
\begin{equation}
\Lambda_\nu(\theta) := \lim_{n \to \infty} \frac{m}{n^2 \eps^4} \Lambda_{n, \nu}\left(\frac{n^2 \eps^4}{m} \theta\right)
\end{equation}

\subsection{Poissonization}
Let $\Tilde{S}_{Poi(\lam)}$ be the Poissonized statistic. We compute the conditional MGF of $\Tilde{S}_{Poi(\lam)}$ with MGF parameter $\frac{n^2 \eps^4}{m} \theta$ conditioned on it being the case that $\phi(Z_j) \le \beta$ for all $j$. That is,
\begin{equation}
\label{eq:squared_A_lam(theta)}
A_\lam(\theta) := \exp(-\eps^2 \theta n) \E\left[ \1_{\{\forall j, \phi(Z_j) \le \beta \}}\exp\left(\eps^2 \theta \sum_{j=1}^m \phi(Z_j)^2 \right) \right]
\end{equation}

where $Z_j \sim Poi(\lam \nu_j)$ and are independent. Due to this independence, $$A_\lam(\theta) = \exp(-\eps^2 \theta n) \prod_{j=1}^m \E\left[\1_{\{\phi(Z_j) \le \beta \} } \exp\left(\eps^2 \theta \phi(Z_j)^2 \right) \right]$$
\begin{lemma}
\label{lem:satisfying_squared_assumption}
Suppose Assumption~\ref{assumption:squared} holds. We have,
\begin{align*}
    &\eps^2 \theta \E[\1_{\{\phi(Z_j) \le \beta\}} \phi(Z_j)^2] = \eps^2 \theta \E[\phi(Z_j)^2] + o(\Delta^2)\\
    &\frac{\eps^4\theta^2}{2} \E\left[\1_{\{\phi(Z_j) \le \beta\}} \phi(Z_j)^4 \right] = \frac{\eps^4 \theta^2}{2} \E[\phi(Z_j)^4] + o(\Delta^2)\\
    &\sum_{l=3}^\infty \frac{(\eps^2 \theta)^l}{l!} \E\left[\1_{\{\phi(Z_j) \le \beta\}} \phi(Z_j)^{2l} \right] = o(\Delta^2)
\end{align*}
    when $Z_j \sim Poi(\lam \nu_j)$ for $\lam = n(1 + O(\eps^2))$ and $\nu_j = 1/m + O(\eps/m)$ for all $j$.
\end{lemma}
\begin{proof}
For the first claim, $$\eps^2 \theta \E[\1_{\{\phi(Z_j) \le \beta\}} \phi(Z_j)^2] = \eps^2 \theta \E[\phi(Z_j)^2] - \eps^2 \theta \E[\1_{\{\phi(Z_j) > \beta\}} \phi(Z_j)^2] $$
By Lemma~\ref{lem:huber_tail_moment_bound}, the second term is $o(\Delta^2)$. Thus, we have the first claim. The second claim can be proved in a similar way.
The third claim follows by Lemma~\ref{lem:quadratic_terms_bound}.
\end{proof}
%
%
%

\subsection{Depoissonization}

As before, we will first show that $A_\lam(\theta)$ is analytic in $\lam$.

\begin{lemma}
\label{lem:squared_analytic}
$A_\lam(\theta)$ is analytic in $\lam$.
\end{lemma}

\begin{proof}
We will show that $\E[\1_{\{\phi(Z_j) \le \beta \}} \exp(\eps^2 \theta \phi(Z_j)^2)]$ is analytic. We have that it can be written as 
$$\sum_{k : \phi(k) \le \beta} \left[ \frac{(\lam \nu_j)^k}{k!} e^{-\lam \nu_j} \exp(\eps^2 \theta \phi(k)^2)\right]$$

which is a finite sum of analytic functions, and is thus analytic. The claim follows.
\end{proof}

Note that for $A_\lam(\theta)$ defined in \eqref{eq:squared_A_lam(theta)}, by Lemmas \ref{lem:satisfying_squared_assumption} and \ref{lem:squared_analytic}, Assumption \ref{assumption:xi} holds for $$\xi(k) = \1_{\{\phi(k) \le \beta\}} \exp(\eps^2 \theta \phi(k)^2)$$

Then, we have the following

\begin{lemma}
\label{lem:squared_main_depoissonization_lemma}
We have that for uniform distribution $p$ such that $p_j = 1/m$ for all $j$, $$\E_p\left[\exp\left(\frac{n^2 \eps^4}{m} \theta \Tilde{S}\right) | \forall j, \phi(Y_j^n) \le \beta \right] = \exp(-\eps^2 \theta n) \E_p\left[\exp\left(\eps^2 \theta \sum_{j=1}^m \1_{\{\phi(Y_j^n) \le \beta\}} \phi(Y_j^n)^2 \right) \right]$$
$$ = (1 + O(1/n)) \exp\left\{\frac{n^2 \eps^4}{m} (\theta^2 + o(1)) \right\}$$

and for alternate distributions $q$ such that $q_j = \frac{1}{m} + \frac{\eps}{\gamma m}$ for $j \le l$ and $\nu_j = \frac{1}{m} - \frac{\eps}{(1-\gamma) m}$ for $j  >l$, 
$$\E_q\left[\exp\left(\frac{n^2 \eps^4}{m} \theta \Tilde{S}\right) | \forall j, \phi(Y_j^n) \le \beta \right] = \exp(-\eps^2 \theta n) \E_q\left[\exp\left(\eps^2 \theta \sum_{j=1}^m \1_{\{\phi(Y_j^n) \le \beta\}} \phi(Y_j^n)^2 \right) \right]$$

$$ = (1 + O(1/n)) \exp\left\{\frac{n^2 \eps^4}{m} \left[\theta^2 + \theta \frac{1}{\gamma(1-\gamma)} + o(1) \right] \right\}$$
\end{lemma}

\begin{proof}
By Lemma \ref{lem:main_generic_mgf_lemma}, the claim holds.
\end{proof}

\subsection{Application of the G{\"a}rtner-Ellis Theorem}

In this section, we apply the G{\"a}rtner-Ellis Theorem to obtain the probability that our statistic crosses a threshold, under the uniform distribution, and under one of the worst-case $\eps$-far distributions.

\begin{lemma}
\label{lem:gartner_ellis_application_squared}
Under the uniform distribution $p$, we have that for $\tau > 0$, $$\lim_{n \to \infty} - \frac{m}{n^2 \eps^4} \log\left(\Pr_p\left[\Tilde{S} \ge \tau | \forall j, \phi(Y_j^n) \le \beta \right] \right) = \frac{\tau^2}{4}$$


Under an $\eps$-far distribution $q$ of the form $q_j = \frac{1}{m} + \frac{\eps}{\gamma m}$ for $j \le l$ and $q_j = \frac{1}{m} - \frac{\eps}{(1-\gamma)m}$ for $j > l$, and $\gamma = \Theta(1)$, $1-\gamma = \Theta(1)$, for $\tau < \frac{1}{\gamma(1-\gamma)}$,

$$\lim_{n \to \infty} -\frac{m}{n^2 \eps^4} \log \left(\Pr_q\left[\Tilde{S} \le \tau | \forall j, \phi(Y_j^n) \le \beta \right] \right) = \frac{\left(\tau \gamma (\gamma - 1) + 1\right)^2}{4 \gamma^2 (\gamma - 1)^2}$$

\end{lemma}
\begin{proof}
Note that by Lemma \ref{lem:squared_main_depoissonization_lemma}, the limiting logarithmic moment generating function with respect to the uniform distribution $p$ is given by $$\Lambda_p(\theta) = \lim_{n \to \infty} \frac{m}{n^2 \eps^4} \log\left(\E_p\left[\exp\left(\frac{n^2 \eps^4}{m} \theta\Tilde{S}\right) | \forall j, \phi(Y_j^n) \le \beta \right]\right) = \theta^2$$

Thus, Assumption \ref{assumption:gartner_ellis} holds for $\mathcal{D}_{\Lambda_p} = \R$. Furthermore, the Fenchel-Legendre Transform (defined in equation \ref{eq:fenchel_legendre}) of $\Lambda_p$ is given by $$\Lambda_p^*(\tau) = \sup_{\theta } \{\theta \tau - \theta^2\} = \frac{\tau^2}{4}$$

This is a strongly convex function of $\tau$, so the set of exposed points of $\Lambda_p^*$ whose exposing hyperplane belongs to $\mathcal{D}_{\Lambda_p}^o$ is all of $\R$. Thus, by the  Theorem \ref{thm:gartner_ellis} (G{\"a}rtner-Ellis), for $\tau > 0$,

$$ \lim_{n \to \infty} -\frac{m}{n^2 \eps^4} \log\left(\Pr_p\left[\Tilde{S} \ge \tau | \forall j, \phi(Y_j^n) \le \beta \right]\right) = \inf_{x \ge \tau} \Lambda_p^*(x) = \frac{\tau^2}{4}$$

%

Similarly, the limiting logarithmic moment generating function with respect to an alternate distribution $q$ is given by $$\Lambda_q(\theta) = \lim_{n \to \infty} - \frac{m}{n^2 \eps^4} \log \left( \E_q\left[\exp\left(\frac{n^2 \eps^4}{m} \theta \Tilde{S} \right) | \forall j, \phi(Y_j^n) \le \beta \right] \right) = \theta^2 + \frac{1}{\gamma(1-\gamma)} \theta$$

The Fenchel-Legendre transform is given by $$\Lambda_q^*(\tau) = \sup_{\theta} \left\{\theta \tau - \theta^2 - \frac{1}{\gamma(1-\gamma)} \theta\right\} = \frac{\left(\tau \gamma (\gamma-1) + 1\right)^2}{4 \gamma^2 (\gamma-1)^2}$$

Again, applying the G{\"a}rtner-Ellis Theorem gives, for $\tau < \frac{1}{\gamma(1-\gamma)}$,
$$\lim_{n \to \infty} - \frac{m}{n^2 \eps^4} \log\left(\Pr_q\left[\Tilde{S} _n^* \le \tau | \forall j, \phi(Y_j^n) \le \beta \right] \right) = \inf_{x \le \tau} \Lambda^*_q(x) = \frac{\left(\tau \gamma (\gamma-1) + 1\right)^2}{4 \gamma^2 (\gamma-1)^2}$$


\end{proof}

\subsection{Removing the conditioning}\label{subsection:squared_remove_conditioning}
So far, we have analyzed the probability that $\Tilde{S}$ crosses a threshold $\tau$ under the uniform distribution, and under a family of $\eps$-far distributions conditioned on the event that $\phi(Y_j^n) \le \beta$ for all $j$. We will now remove this conditioning.

\begin{lemma}
\label{lem:squared_removed_conditioning_probs}
Under the uniform distribution $p$, we have that for $\tau > 0$,
$$\lim_{n \to \infty} -\frac{m}{n^2 \eps^4} \log \left(\Pr_p\left[\Tilde{S} \ge \tau \right] \right) = \frac{\tau^2}{4}$$
Under an $\eps$-far distribution $q$ for the form $q_j = \frac{1}{m} + \frac{\eps}{\gamma m}$ for $j \le l$ and $q_j = \frac{1}{m} - \frac{\eps}{(1-\gamma) m}$ for $j > l$, and $\gamma = \Theta(1)$, $1-\gamma = \Theta(1)$, for $\tau < \frac{1}{\gamma(1-\gamma)}$,
$$\lim_{n \to \infty} - \frac{m}{n^2 \eps^4} \log \left( \Pr_q\left[\Tilde{S} \le \tau \right] \right) = \frac{(\tau \gamma (\gamma-1) + 1)^2}{4 \gamma^2 (\gamma-1)^2}$$
\end{lemma}

\begin{proof}
For the uniform distribution we will upper and lower bound $\Pr_p [\Tilde S \ge \tau]$ by $$e^{\frac{n^2 \eps^4}{m}\frac{\tau^2}{4}(1 + o(1))}$$
The claim for the uniform distribution will follow.

Under any distribution $\nu$, 
$$\Pr_\nu\left[\Tilde{S} \ge \tau\right] = \Pr_\nu\left[\Tilde{S} \ge \tau | \forall j, \phi(Y_j^n) \le \beta\right] \Pr_\nu[\forall j, \phi(Y_j^n) \le \beta]$$
$$+ \Pr_\nu\left[\Tilde{S} \ge \tau | \exists j, \phi(Y_j^n) > \beta \right] \Pr_\nu[\exists j, \phi(Y_j^n) > \beta]$$

Now, by Bernstein's inequality, since $n \eps^4 = \omega(1)$ by \eqref{eq:squared_eps_larger_than} and $\beta = \kappa \frac{n^2 \eps^4}{m}$, we have, for the uniform distribution $p$, for every $j$, $$\Pr_p[\phi(Y_j^n) > \beta] \le 2e^{-\frac{\kappa}{4} \frac{n^2 \eps^4}{m}}$$

Thus, by union bound, $$\Pr_p\left[\exists j, \phi(Y_j^n) > \beta\right] \le \sum_{j=1}^m \Pr_p[\phi(Y_j^n) > \beta] \le 2 m e^{-\frac{\kappa n^2 \eps^4}{4m}} = 2 e^{-\frac{\kappa n^2 \eps^4}{4 m} + \log m} = 2 e^{- \frac{\kappa n^2 \eps^4}{4m} (1 + o(1))}$$

since $\frac{n^2 \eps^4}{m} = \omega(\log m)$. For $\Pr_p[\forall j, \phi(Y_j^n) \le \beta]$ and $\Pr_p[\Tilde S \ge \tau | \exists j, \phi(Y_j^n) > \beta]$, we apply the trivial upper bound of $1$. So, for $\kappa \ge 2\tau^2$, by Lemma \ref{lem:gartner_ellis_application_squared}, $$\Pr_p[\Tilde{S} \ge \tau] \le e^{- \frac{\tau^2}{4}\frac{n^2 \eps^4}{m} } + 2 e^{- \frac{\tau^2}{2} \frac{n^2 \eps^4}{m} (1 + o(1))} = e^{-\frac{\tau^2}{4}\frac{n^2 \eps^4}{m}} (1 + o(1))$$

For the lower bound, note that
$$\Pr_p[\forall j, \phi(Y_j^n) \le \beta] \ge 1 - \sum_{j} \Pr_p[\phi(Y_j^n) > \beta] \ge 1 - 2m e^{-\frac{\kappa}{4} \frac{n^2 \eps^4}{m}} = 1 - 2 e^{-\frac{\kappa}{4} \frac{n^2 \eps^4}{m} (1 + o(1))} $$

For the second term, i.e., $\Pr_p\left[\Tilde{S} \ge \tau | \exists j, \phi(Y_j^n) > \beta \right] \Pr_p[\exists j, \phi(Y_j^n) > \beta]$, we apply the trivial lower bound of $0$. Again, setting $\kappa \ge 2 \tau^2$, we get $$\Pr_p\left[\Tilde S \ge \tau \right] \ge e^{-\frac{n^2\eps^4}{m} \frac{\tau^2}{4} (1 + o(1))}$$

So, we have the claim for the uniform distribution. We will upper and lower bound the probability of crossing $\tau$ for $\eps$-far distribution $q$ in the same way. For ease, let $$c_\tau := \frac{(\tau \gamma(\gamma-1) + 1)^2}{4 \gamma^2 (\gamma-1)^2}$$

By Bernstein's inequality, we have that since $\gamma = \Theta(1)$ and $1-\gamma = \Theta(1)$ so that $q_j = 1/m + O(\eps/m)$ for every $j$, and since $n \eps^4 = \omega(1)$ by \eqref{eq:squared_eps_larger_than} and $\beta = \kappa \frac{n^2 \eps^4}{m}$, for every $j$, under $\eps$-far distribution $q$, $$\Pr_q[\phi(Y_j^n) > \beta] \le 2 e ^{-\frac{\kappa}{4} \frac{n^2 \eps^4}{m} (1 + o(1))}$$

Then, setting $\kappa \ge 8 c_\tau$, and repeating the same argument as in the uniform case, we obtain the required upper and lower bounds on $\Pr_q[\Tilde{S} \le \tau]$. The claim follows.
\end{proof}

\subsection{Setting the threshold}

We need to set our threshold $\tau$ so that the minimum of the error probability under the uniform distribution $p$, and any $\eps$-far distribution $q$ is maximized. Note that, it is sufficient to consider a threshold $\tau$ such that $0 < \tau < \frac{1}{\gamma(1-\gamma)}$, since otherwise, the error probability in one of the two cases is at least constant.  To set our threshold, we will first observe that for any $\tau$ in this range, the ``error exponent'' under $\eps$-far distributions is minimized for a particular $\eps$-far distribution. Then, we will set the threshold to maximize the minimum of the error exponent under the uniform distribution, and under this $\eps$-far distribution.

\begin{lemma}
\label{lem:threshold_squared}
Setting the threshold $\tau = 2$, we have for the uniform distribution $p$, $$\lim_{n \to \infty} -\frac{m}{n^2 \eps^4} \log \left( \Pr_p\left[\Tilde{S} \ge \tau \right] \right) = 1$$
and for any $\eps$-far distribution $q$ such that $q_j = \frac{1}{m} + \frac{\eps}{\gamma m}$ for $j \le l$ and $q_j = \frac{1}{m} - \frac{\eps}{(1-\gamma)m}$ for $j > l$ and $\gamma = \Theta(1), 1-\gamma = \Theta(1)$,$$\lim_{n \to \infty} - \frac{m}{n^2 \eps^4} \log \left( \Pr_q \left[ \Tilde{S} \le \tau \right]\right) \ge 1$$
 with equality for $q$ such that $q_j = 1/m + 2 \eps/m$ for $j \le m/2$ and $q_j = 1/m - 2 \eps/m$ for $j > m/2$.
\end{lemma}

\begin{proof}
By Lemma \ref{lem:squared_removed_conditioning_probs}, for $0 < \tau < \frac{1}{\gamma(1-\gamma)}$, $$\lim_{n \to \infty} \frac{m}{n^2 \eps^4} \log \left( \Pr_q \left[\Tilde{S} \le \tau \right]\right) = \frac{\left(\tau \gamma (\gamma-1) + 1\right)^2}{4 \gamma^2 (\gamma - 1)^2}$$

Now, the numerator of the right hand side is minimized when $\gamma = 1/2$, and the denominator is maximized when $\gamma = 1/2$. Thus, the right hand side is minimized when $\gamma = 1/2$. So, we have that,
$$\lim_{n \to \infty} -\frac{m}{n^2 \eps^4} \log\left(\Pr_{q}\left[ \Tilde{S} \le \tau \right] \right) \ge \frac{1}{4} (\tau - 4)^2$$
 with equality for distribution $q$ such that $q_j = 1/m + 2 \eps/m$ for $j \le m/2$ and $q_j = 1/m - 2 \eps/m$ for $j > m/2$.
 
Then, the claim follows by substituting in $\tau = 2$ in the expression for the uniform distribution in Lemma \ref{lem:gartner_ellis_application_huber} and in the above expression.
\end{proof}

Recall that our target sample complexity is 
\begin{align}\tag{\ref{eq:nboundtight}}
  n = (1 + o(1)) \sqrt{m \log \frac{1}{\delta}}/\eps^2
\end{align}
We have our result for the squared/collisions tester.
\collisions*
\begin{proof}
We need to show that we can satisfy \eqref{eq:Delta_constraint}, \eqref{eq:beta_second_bound} and \eqref{eq:beta_third_bound} for the $\beta$ that we chose in \eqref{eq:beta_setting_squared}, and some $\Delta$. Set $$\Delta = \frac{n \eps^2}{m} = \Theta(\eps^2)$$ so that \eqref{eq:Delta_constraint} is satisfied. Now, $$\beta = \kappa \frac{n^2 \eps^4}{m} = \omega(\log m) = \omega\left(\log\left(\frac{1}{\Delta}\right) \right)$$
and since $n = \Theta(m)$,
$$\beta = \kappa \frac{n^2 \eps^4}{m} = \Theta\left(\sqrt{\frac{n}{m}} \frac{n^2 \eps^4}{m} \right) = \omega\left(\sqrt{\frac{n}{m}} \log m\right) = \omega\left(\sqrt{\frac{n}{m} \log\left(\frac{1}{\Delta}\right)} \right)$$
Thus, \eqref{eq:beta_second_bound} is satisfied. For \eqref{eq:beta_third_bound}, note that $$\left(\beta^2 \eps^2\right)^3 = \Theta\left(\left(\frac{n^4 \eps^{10}}{m^2}\right)^3\right) = \Theta \left(\frac{n^{12} \eps^{30}}{m^6}\right) = \Theta\left(n^6 \eps^{30}\right) = o(\eps^4) = o(\Delta^2) $$

since $\eps = o\left(\frac{1}{n^{3/13}}\right)$. So, the conditions are satisfied.

By Lemma \ref{lem:threshold_squared}, we have that $\bar c(\eps, m, C) = 1$ for every $\eps$ that satisfies our assumptions, and every $C > 2$. In particular, any $\eps'$ such that $\left(1 - \frac{1}{C'} \right) \eps(n) \le \eps'(n) \le \eps(n)$ has $\bar{c}(\eps', m, C) = 1$ for every $C > 2$. Thus, by Lemma \ref{lem:final_worst_case_lemma}, we have that $c(\eps, m) = 1$ for every $\eps$ that satisfies our assumptions. The claim follows.
\end{proof}

%% file: empty_bins_n_Theta_m_eps_to_0.tex
\section{Empty bins statistic in Sublinear Regime}\label{section:empty_bins}
\begin{assumption}\label{assumption:empty_bins}
$n/m \lesssim 1$, $\eps, \delta \ll 1$, $n \gg 1$.
\end{assumption}
We will assume that Assumption~\ref{assumption:empty_bins} holds throughout this section.
Let our statistic be $$S := \sum_{j=1}^m \1_{ \{Y_j^n = 0\}}$$ where $Y_j^n = \sum_{i=1}^n \1_{\{X_i = j\}}$, and $X_1, \dots, X_n$ are the $n$ samples drawn from the distribution $\nu$ supported on $[m]$. In other words, it is the number of empty bins. Note that this has the same error probability as the total variation statistic when $n \le m$. For ease of exposition, we will analyze the statistic
\begin{equation}
\Tilde{S} = \frac{m}{n^2 \eps^2} [S - m e^{-n/m}]
\end{equation} 
Note that this has the same error probability as $S$. Consider the moment generating function (MGF) of $\Tilde{S}$ with respect to distribution $\nu$, given by $$M_{\Tilde{S}, \nu}(\theta) = \E_\nu\left[\exp(\theta \Tilde{S})\right]$$

Note that the logarithmic moment generating function of this statistic is given by 
\begin{equation}
\Lambda_{n, \nu}(\theta) := \log(\E_\nu[\exp(\theta \Tilde{S})])
\end{equation}
We wish to study the error exponent \citep{DBLP:journals/tit/HuangM13} of this statistic with respect to normalization $\frac{n^2 \eps^4}{m}$. To do this, we will compute the asymptotic expansion of the limiting logarithmic moment generating function of the statistic, given by 
\begin{equation}
\Lambda_\nu(\theta) := \lim_{n \to \infty} \frac{m}{n^2 \eps^4} \Lambda_{n, \nu}( \frac{n^2 \eps^4}{m} \theta)\label{eq:empty_bins_lim_log_mgf_def}
\end{equation}

\subsection{Poissonization}
Define $\Tilde{S}_{Poi(\lam)}$ to be the Poissonized statistic, that is the statistic $\Tilde{S}$ when the number of samples is chosen according to the Poisson distribution with mean $\lam$. We begin by computing the MGF of $\Tilde{S}_{Poi(\lam)}$ with MGF parameter $\frac{n^2 \eps^4}{m} \theta$. Let this be $$A_\lam(\theta) := \E\left[\exp\left(\frac{n^2 \eps^4}{m} \theta \Tilde{S}_{Poi(\lam)}\right) \right] = \exp\left(- m e^{-n/m} \eps^2 \theta \right)\E\left[ \exp\left( \eps^2 \theta \sum_{j=1}^m \1_{\{Z_j = 0\}}\right)\right]$$
where $Z_j \sim Poi(\lam \nu_j)$ are independent. Due to this independence, 
\begin{equation}
\label{eq:empty_bins_A_lam_expression}
A_\lam(\theta) = \exp\left(-m e^{-n/m} \eps^2 \theta \right)\prod_{j=1}^m \E\left[\exp\left(\eps^2 \theta \1_{\{Z_j = 0 \}}\right)\right]
\end{equation}
Now, we have the following.
\begin{lemma}
\label{lem:empty_bins_indicator_mgf}
$$\E\left[\exp\left(\eps^2 \theta \1_{\{Z_j = 0\}}\right) \right] = e^{-\lam \nu_j} e^{\eps^2 \theta} - e^{-\lam \nu_j} + 1$$
when $Z_j \sim Poi(\lam \nu_j)$.
\end{lemma}
\begin{proof}
\begin{align*}
\E\left[\exp\left(\eps^2 \theta \1_{\{Z_j = 0\}}\right) \right] &= \exp(\eps^2 \theta) \Pr[Z_j = 0] + \sum_{k=1}^\infty \Pr[Z_j = k]\\
&= \exp(\eps^2 \theta) \Pr[Z_j = 0] + 1 - \Pr[Z_j = 0] = e^{-\lam \nu_j} e^{\eps^2 \theta} - e^{-\lam \nu_j} + 1
\end{align*}
\end{proof}
\subsection{Depoissonization}
First, we will show that $A_\lam(\theta)$ is analytic in $\lam$.
\begin{lemma}
$A_\lam(\theta)$ is analytic in $\lam$.
\end{lemma}
\begin{proof}
By Lemma \ref{lem:empty_bins_indicator_mgf}, $\E[\exp(\eps^2 \theta \1_{\{Z_j = 0\}})]$ is analytic $\lam$ since it is the finite sum of analytic functions. Then, $A_\lam(\theta)$ is the finite product of analytic functions, and is hence analytic. The claim follows.
\end{proof}

\begin{lemma}
\begin{equation}
\label{eq:empty_bins_depoissonization_integral}
M_{\Tilde{S}, \nu}\left(\frac{n^2 \eps^4}{m} \theta\right) = \frac{n!}{2 \pi i}\oint \frac{e^\lam}{\lam^{n+1}} A_\lam(\theta) d \lam
\end{equation}
\end{lemma}

\begin{proof}
Follows from Lemma \ref{lem:general_depoissonization_integral} since $A_\lam(\theta)$ is analytic in $\lam$.
\end{proof}

We will choose a contour passing through a particular $\lam_0$, and this will make it easy to evaluate the integral. We carry out the integration along the contour given by $\lam = \lam_0 e^{i \psi}$, where 
\begin{equation}
\label{eq:empty_bins_lam_0_setting}
\lam_0 = n(1 + e^{-n/m} \eps^2 \theta)
\end{equation}
We substitute $\lam = \lam_0 e^{i \psi}$ and $A_\lam(\theta)$ from \eqref{eq:empty_bins_A_lam_expression} into \eqref{eq:empty_bins_depoissonization_integral} to get that
\begin{equation}
\label{eq:empty_bins_mgf_integral}
M_{\Tilde{S}, \nu}\left(\frac{n^2 \eps^4}{m} \theta\right) = \exp\left(- m  e^{-n/m} \eps^2 \theta\right) \frac{n!}{2 \pi} \lam_0^{-n} Re\left[\int_{-\pi}^\pi g(\psi) d \psi\right]
\end{equation}
with 
\begin{equation}
\label{eq:empty_bins_h_def}
g(\psi) := e^{-i n \psi} e^{\lam_0 e^{i\psi}} \prod_{j=1}^m \left\{e^{-\lam_0 \nu_j e^{i \psi}} e^{\eps^2 \theta} - e^{-\lam_0 \nu_j e^{i\psi}} + 1\right\}
\end{equation}

We will split this integral into $3$ parts. Let 
\begin{equation}
    \label{eq:empty_bins_split_integrals}
    \begin{gathered}
        I_1 = Re\left[\int_{-\pi/3}^{\pi/3} g(\psi) d \psi \right]\\
        I_2 = Re\left[\int_{-\pi}^{-\pi/3} g(\psi) d \psi \right]\\
        I_3 = Re\left[\int_{\pi/3}^{\pi} g(\psi) d \psi \right]
    \end{gathered}
\end{equation}

We will show that $I_1$ dominates. We show this by bounding $g(\psi)$ in the region $\psi \in [-\pi, -\pi/3]\cup [\pi/3, \pi]$ as follows.
\begin{lemma}
\label{lem:empty_bins_small_integrals_bound}
For $\psi \in [-\pi, -\pi/3] \cup [\pi/3, \pi]$, $$|g(\psi)| \le \exp\{0.5 n(1 + O(\eps^2)) + 0.5  m  e^{-n/m} \eps^2 \theta(1 + O(\eps^2))\}$$
\end{lemma}

\begin{proof}
By definition of $g$ from \eqref{eq:empty_bins_h_def} and Lemma \ref{lem:empty_bins_poissonized_mgf_bound}, we have that for $Z_j \sim Poi(\lam_0 \nu_j)$,
\begin{align*}
|g(\psi)| &=\left| e^{-i n \psi} e^{\lam_0 e^{i\psi}} \prod_{j=1}^m \left\{e^{-\lam_0 \nu_j e^{i\psi}} e^{\eps^2 \theta} - e^{-\lam_0 \nu_j e^{i\psi}} + 1 \right\}\right|\\
&= \Big|e^{-i n \psi} e^{\lam_0 e^{i\psi}}  \exp\left\{m e^{-\frac{n}{m} e^{i \psi}} \eps^2 \theta (1 + O(\eps^2)) + O(n \eps^3) \right\} \Big |
\end{align*}

Since $\lam_0 = n(1 + O(\eps^2))$, this is
$$\Big|e^{-i n \psi} \exp\{n e^{i \psi} + m e^{-\frac{n}{m} e^{i \psi}} \eps^2 \theta (1 + O(\eps^2)) + O(n \eps^2) \}\Big|$$

The claim follows since $|e^{-i n \psi}| = 1$ and $$|\exp(e^{i \psi})| = |\exp(\cos( \psi ) + i \sin(\psi))| = \exp(\cos(\psi)) \le \exp(0.5)$$ for $\psi$ in the range stated.
\end{proof}

Note that this implies that for the integrals defined in \eqref{eq:empty_bins_split_integrals} that 
\begin{equation}
\label{eq:empty_bins_small_integrals}
I_2 + I_3 = O(e^{0.6 n + 0.6 m e^{-n/m} \eps^2 \theta })
\end{equation}

Now, we will compute $I_1$. Define $G(\psi) := \log(g(\psi))$. Then, by definition of $g$ in \eqref{eq:empty_bins_h_def},

\begin{equation}
\label{eq:empty_bins_H(psi)}
G(\psi) = - i n\psi  + \lam_0 e^{i \psi} + \sum_{j=1}^m \log\left\{e^{-\lam_0 \nu_j e^{i\psi}} e^{\eps^2 \theta} - e^{-\lam_0 \nu_j e^{i\psi}} + 1\right\}
\end{equation}

Note that \begin{equation}
\label{eq:empty_bins_G(0)_imaginary_0}
    \Im(G(0)) = 0
\end{equation}

Then, applying Lemma \ref{lem:squared_huber_H_derivatives},
\begin{equation}
\label{eq:empty_bins_G'(0)_0}
\Re(G'(0)) = 0 
\end{equation}
Now, computing the asymptotic expansion of $G''(\psi)$ by Lemma \ref{lem:empty_bins_G_second_derivative}, we have
\begin{equation}
\label{eq:empty_bins_H''(psi)_asymptotic}
G''(\psi) = - n e^{i \psi} + O\left(n \eps^2\right)
\end{equation}

Now, by Taylor's theorem, for any $\psi \in [- \pi/3, \pi/3]$ there exists $\tilde{\psi} \in (0, \psi)$ such that
\begin{equation}
\label{eq:empty_bins_taylors_theorem}
G(\psi) = G(0) + G'(0) \psi + \frac{G''(\tilde{\psi})}{2} \psi^2
\end{equation}

But, by \eqref{eq:empty_bins_H''(psi)_asymptotic}, $Re[G''(\psi)] \le -0.4 n$ for any $\psi \in [-\pi/3, \pi/3]$. So, for $\psi \in [-\pi/3, \pi/3]$, 
\begin{equation}
\label{eq:empty_bins_G(psi)_at_most_G(0)_with_residual}
Re(G(\psi)) \le G(0) - 0.2 n \psi^2
\end{equation}

Now, we have the following upper bound on $I_1$.
\begin{lemma}
\label{lem:empty_bins_I_1_upper_bound}
$$I_1 \le e^{G(0)} \frac{\sqrt{\pi}}{\sqrt{0.2n}}$$
\end{lemma}

\begin{proof}
\begin{equation}
I_1 = Re\left[\int_{-\pi/3}^{\pi/3} e^{G(\psi)} d \psi\right] \le \int_{- \pi /3}^{\pi/3} e^{Re(G(\psi))} d \psi \le e^{G(0)} \int_{-\pi/3}^{\pi/3} e^{-0.2 n \psi^2} d \psi
\end{equation}

\end{proof}

$$ \le e^{G(0)} \int_{-\infty}^\infty e^{-0.2 n \psi^2}d \psi = e^{G(0)} \frac{\sqrt{\pi}}{\sqrt{0.2n}}$$

The next lemma shows that $I_1$ is also lower bounded by the above quantity (up to constants).

\begin{lemma}
\label{lem:empty_bins_I_1_lower_bound}
$$I_1 \ge e^{G(0)} \frac{0.5 \sqrt{\pi}}{\sqrt{1.1n}} (1 + o(1))$$
where $I_1$ is defined in \eqref{eq:empty_bins_split_integrals}
\end{lemma}

\begin{proof}
By \eqref{eq:empty_bins_H''(psi)_asymptotic},  $\Im(G''(\psi)) = - n \sin(\psi) + O\left(n \eps^2\right)$. So, for large enough $n$, since $|\sin(\psi)| \le |\psi|$, for any $\psi \in [-\pi/3, \pi/3]$, $|\Im(G''(\psi))| \le 1.1 n |\psi|$. So, by \eqref{eq:empty_bins_taylors_theorem}, \eqref{eq:empty_bins_G(0)_imaginary_0} and \eqref{eq:empty_bins_G'(0)_0}, we have that for constant $c > 0 $ and $\psi \in [-\pi/3, \pi/3]$, 
$$|\Im(G(\psi))| \le 1.1 n |\psi|^3 + c n \eps^2 \psi^2$$

Also, $\Re(G''(\psi)) \ge -1.1 n$ by a similar argument. So, by \eqref{eq:empty_bins_taylors_theorem} and \eqref{eq:empty_bins_G'(0)_0}, for $\psi \in [-\pi/3, \pi/3]$,
$$\Re(G(\psi)) \ge G(0) - 1.1 n \psi^2$$

Now, for $t_n = 0.1 \min\{n^{-1/3}, \frac{1}{\eps\sqrt{c n} }\}$, we have that for $\psi \in [-t_n, t_n]$, $\cos(\Im(G(\psi))) \ge 0.5$ so that $\Re(e^{G(\psi)}) \ge 0.5 e^{\Re(G(\psi)}$. We can split $I_1$ further into $3$ parts:

$$I_1 = \Re\left[ \int_{-\pi/3}^{-t_n} e^{G(\psi)} d \psi\right] + \Re\left[\int_{t_n}^{\pi/3} e^{G(\psi)} d \psi \right] + \Re\left[\int_{-t_n}^{t_n} e^{G(\psi)} d \psi \right]$$

Now, by \eqref{eq:empty_bins_G(psi)_at_most_G(0)_with_residual}, $$\left| \int_{-\pi/3}^{-t_n}e^{G(\psi)} d \psi \right| \le e^{G(0)} \int_{- \infty}^{-t_n} e^{-0.2 n \psi^2} d \psi = t_n e^{G(0)} \int_{-\infty}^{-1} e^{-0.2 n t_n^2 \bar{\psi}^2} d \bar \psi$$
$$ \le t_n e^{G(0)} \int_{-\infty}^{-1} e^{-0.2 n t_n^2 |\bar \psi|} d \bar \psi = e^{G(0)} O\left(\frac{1}{n t_n} \right) = e^{G(0)} o\left(\frac{1}{\sqrt{n}} \right)$$

In a similar way, we can bound the second term. For the third term, we have $$\Re\left[ \int_{-t_n}^{t_n} e^{G(\psi)} d \psi \right] \ge \int_{-t_n}^{t_n} 0.5 e^{\Re(G(\psi))} d \psi \ge 0.5 e^{G(0)} \int_{-t_n}^{t_n} e^{-1.1 n \psi^2} d \psi$$

$$\ge 0.5 e^{G(0)} \left[ \int_{-\infty}^\infty e^{-1.1 n \psi^2} d \psi - 2 \int_{-\infty}^{- t_n} e^{-1.1 n \psi^2} d \psi \right]$$
$$\ge 0.5 e^{G(0)} \left(\frac{\sqrt{\pi}}{\sqrt{1.1 n}} + O\left(\frac{1}{n t_n}\right) \right) = 0.5 e^{H(0)} \frac{\sqrt{\pi}}{\sqrt{1.1 n}} (1 + o(1))$$

Combining the bounds, we get that $$I_1 \ge e^{G(0)} \frac{0.5 \sqrt{\pi}}{\sqrt{1.1 n}} (1 + o(1))$$

\end{proof}

Combining the upper bound on $I_1$ from Lemma \ref{lem:empty_bins_I_1_upper_bound} and the lower bound from Lemma \ref{lem:empty_bins_I_1_lower_bound}, we have

$$I_1 = e^{G(0)} \frac{1}{\sqrt{n}} e^{O(1)}$$

So, by \eqref{eq:empty_bins_mgf_integral} and \eqref{eq:empty_bins_small_integrals}, 
\begin{equation}
\label{eq:empty_bins_mgf_H(0)}
M_{\Tilde S, \nu}\left(\frac{n^2 \eps^4}{m} \theta \right) = \exp(- m e^{-n/m} \eps^2 \theta) \frac{n!}{2 \pi} \lam_0^{-n} e^{G(0)} \frac{\sqrt{\pi}}{\sqrt{0.2n}}(1+o(1))
\end{equation}
So, it remains to compute $G(0)$.

\begin{lemma}
\label{lem:empty_bins_uniform_H(0)}
Under the uniform distribution given by $\nu_j = 1/m$ for all $j$, and $\lam_0 = n(1 + e^{-n/m} \eps^2 \theta)$,
$$G(0) = \lam_0 + m e^{-n/m} \eps^2 \theta + \eps^4 \theta^2[- n e^{-2n/m} + m \frac{e^{-n/m}}{2} - m \frac{e^{-2n/m}}{2}] + O(m \eps^6)$$
\end{lemma}
\begin{proof}
By definition of $G(\psi)$ in equation \eqref{eq:empty_bins_H(psi)} and Lemma \ref{lem:empty_bins_indicator_mgf},
$$G(0) = \lam_0 + \sum_{j=1}^m \log[1 + e^{- \lam_0 \nu_j} (e^{\eps^2 \theta} - 1)]$$
Now, since $e^{\eps^2\theta} - 1 = \eps^2 \theta + \frac{\eps^4 \theta^2}{2} + O(\eps^6)$, this is 
$$ \lam_0 + \sum_{j=1}^m [ e^{-\lam_0 \nu_j} (\eps^2\theta + \frac{\eps^4\theta^2}{2}) - \frac{\eps^4 \theta^2}{2} e^{-2\lam_0 \nu_j} + O(\eps^6) ]$$
Substituting in $\lam_0$ from \eqref{eq:empty_bins_lam_0_setting}, and $\nu_j = 1/m$ for all $j$, this is
$$ = \lam_0 + \sum_{j=1}^m [e^{-\frac{n}{m}} e^{-\frac{n}{m} e^{-\frac{n}{m}} \eps^2\theta} (\eps^2 \theta + \frac{\eps^4 \theta^2}{2}) - \frac{\eps^4 \theta^2}{2} e^{-2 \frac{n}{m} (1 + O(\eps^2))}) + O(\eps^6)]$$
$$ = \lam_0 + \sum_{j=1}^m [e^{-\frac{n}{m}}(1 - \frac{n}{m} e^{-\frac{n}{m}} \eps^2 \theta) (\eps^2 \theta + \frac{\eps^4 \theta^2}{2}) - \frac{\eps^4 \theta^2}{2}  e^{-2\frac{n}{m}} + O(\eps^6)]$$
$$ = \lam_0 + \sum_{j=1}^m [e^{-\frac{n}{m}} \eps^2 \theta - \frac{n}{m} e^{-2\frac{n}{m}} \eps^4 \theta^2 + \frac{e^{-\frac{n}{m}}}{2} \eps^4 \theta^2 - \frac{\eps^4 \theta^2}{2} e^{-2\frac{n}{m}} + O(\eps^6)]$$

$$ = \lam_0 + m e^{-\frac{n}{m}} \eps^2 \theta + \eps^4 \theta^2[- n e^{-2\frac{n}{m}} + m \frac{e^{-\frac{n}{m}}}{2} - m \frac{e^{-2\frac{n}{m}}}{2}] + O(m \eps^6)$$
\end{proof}

\begin{lemma}
\label{lem:empty_bins_worst_case_G(0)}
Under distribution $\nu$ such that $\nu_j = \frac{1}{m} + \frac{\eps}{\gamma m}$ for $j \le \gamma m$ and $\nu_j = \frac{1}{m} - \frac{\eps}{(1-\gamma) m}$ for $j > l$, for $\gamma = \Theta(1)$, $(1-\gamma) = \Theta(1)$, and $\lam_0 = n(1 + e^{-n/m} \eps^2 \theta)$, we have 
\begin{dmath*}
G(0) = \lam_0 + m e^{-\frac{n}{m}} \eps^2 \theta + \eps^4 \theta^2\left[-n e^{-2n/m} + m\frac{e^{-n/m}}{2} - m \frac{e^{-2n/m}}{2} \right] + \eps^4 \theta \frac{e^{-n/m}  n^2}{2m \gamma (1-\gamma)}
\end{dmath*}
\end{lemma}

\begin{proof}
By definition of $G(\psi)$ in equation \eqref{eq:empty_bins_H(psi)} and Lemma \ref{lem:empty_bins_indicator_mgf},
$$G(0) = \lam_0 + \sum_{j=1}^m \log[1 + e^{- \lam_0 \nu_j} (e^{\eps^2 \theta} - 1)]$$
Now, since $e^{\eps^2 \theta} - 1 = \eps^2 \theta + \frac{\eps^4 \theta^2}{2} + O(\eps^6)$, this is $$ \lam_0 + \sum_{j=1}^m [ e^{-\lam_0 \nu_j} (\eps^2\theta + \frac{\eps^4\theta^2}{2}) - \frac{\eps^4 \theta^2}{2} e^{-2\lam_0 \nu_j} + O(\eps^6) ]$$
Substituting in $\lam_0$ and $\nu$, this is
\begin{dmath*}
= \lam_0 + m e^{-\frac{n}{m}} \eps^2 \theta + \eps^4 \theta^2\left[-n e^{-2n/m} + m\frac{e^{-n/m}}{2} - m \frac{e^{-2n/m}}{2} \right] + \eps^4 \theta \frac{e^{-n/m} m n^2}{2m\gamma(1-\gamma)}
\end{dmath*}
\end{proof}

Finally, we compute the MGF.

\begin{lemma}
\label{lem:empty_bins_mgfs}
Under the uniform distribution $p$ over $[m]$,
$$M_{\Tilde{S}, p}\left(\frac{n^2 \eps^4}{m} \theta\right)= (1 + O(1/n)) \exp\left\{\frac{n^2\eps^4}{m} \theta^2 \left[-\frac{m}{n} \frac{e^{-\frac{2n}{m}}}{2} + \frac{m^2}{n^2} \frac{e^{-n/m}}{2} - \frac{m^2}{n^2} \frac{e^{-\frac{2n}{m}}}{2} \right] \right\}$$
and over distribution $q$ such that $q_j = \frac{1}{m} + \frac{\eps}{\gamma m}$ for $j \le \gamma m$ and $\nu_j = \frac{1}{m} - \frac{\eps}{(1-\gamma) m}$ for $j > \gamma m$, for $\gamma = \Theta(1)$ and $1-\gamma = \Theta(1)$, we have
\begin{align*}
    M_{\Tilde S, q}\left(\frac{n^2 \eps^4}{m} \theta \right) = (1 + O(1/n))\exp\left\{ \frac{n^2 \eps^4}{m} \theta^2 \left[-\frac{m}{n} \frac{e^{-\frac{2n}{m}}}{2} + \frac{m^2}{n^2} \frac{e^{-n/m}}{2} - \frac{m^2}{n^2} \frac{e^{-\frac{2n}{m}}}{2} \right] + \eps^4 \theta \frac{e^{-n/m} n^2}{2 m \gamma(1-\gamma)}\right\}\\
\end{align*}
\end{lemma}

\begin{proof}
For the uniform distribution $p$, by \eqref{eq:empty_bins_mgf_H(0)}, substituting $\lam_0 = n(1 + e^{-n/m} \eps^2 \theta)$ and $G(0)$ from Lemma \ref{lem:empty_bins_uniform_H(0)},  we have 
\begin{align*}
M_{\Tilde{S}, p}\left(\frac{n^2 \eps^4}{m} \theta \right) =\exp(-m e^{-n/m} \eps^2 \theta) \frac{e^n n!}{\sqrt{2 \pi n}} (n(1 + e^{-n/m} \eps^2 \theta))^{-n} \\
\exp\left(n e^{-n/m} \eps^2 \theta + m e^{-n/m} \eps^2 \theta + \eps^4 \theta^2 \left[ -n e^{-2\frac{n}{m}} + m \frac{e^{-\frac{n}{m}}}{2} - m \frac{e^{-2 \frac{n}{m}}}{2} \right]  + O(m \eps^6) \right)\\
= \frac{e^{n} n!}{\sqrt{2 \pi n}} \exp(-n( \log n + \log (1 + e^{-n/m}\eps^2 \theta)))\\
\exp\left(n e^{-n/m} \eps^2 \theta + \eps^4 \theta^2\left[ -n e^{-2n/m} + m \frac{e^{-n/m}}{2} - m \frac{e^{-2n/m}}{2} \right] + O(m \eps^6)\right)\\
= \frac{e^n n!}{n^n \sqrt{2 \pi n}} \exp\left\{-n (e^{-n/m}\eps^2 \theta - \frac{e^{-2 n /m}\eps^4 \theta^2}{2})\right\}\\
\exp\left\{n e^{-n/m} \eps^2 \theta + \eps^4 \theta^2 \left[ -n e^{-2n/m} + m \frac{e^{-n/m}}{2} - m \frac{e^{-2n/m}}{2} \right] + O(m \eps^6)\right\}\\
= (1 + O(1/n)) \exp\left\{\frac{n^2\eps^4}{m} \theta^2 \left[-\frac{m}{n} \frac{e^{-\frac{2n}{m}}}{2} + \frac{m^2}{n^2} \frac{e^{-n/m}}{2} - \frac{m^2}{n^2} \frac{e^{-\frac{2n}{m}}}{2} \right] \right\}
\end{align*}
by Stirling's approximation.

Similarly, for $q$ as stated, we have the claim.
\end{proof}
\subsection{Application of the G{\"a}rtner-Ellis Theorem}

In this section, we apply the G{\"a}rtner-Ellis Theorem to obtain the probability that our statistic crosses a threshold, under the uniform distribution, and under one of the worst-case $\eps$-far distributions.

\begin{lemma}
\label{lem:gartner_ellis_application_empty_bins_n_Theta(m)}
When $n = \Theta(m)$, let $\alpha = n/m$. Then, under the uniform distribution $p$, we have that for $\tau > 0$, $$\lim_{n \to \infty} - \frac{m}{n^2 \eps^4} \log\left(\Pr_p\left[\Tilde{S} \ge \tau \right] \right) = \frac{\tau^2 \alpha^2 e^{2 \alpha}}{2e^\alpha - 2 - 2 \alpha }$$


Under an $\eps$-far distribution $q$ of the form $q_j = \frac{1}{m} + \frac{\eps}{\gamma m}$ for $j \le \gamma m$ and $q_j = \frac{1}{m} - \frac{\eps}{(1-\gamma)m}$ for $j > \gamma m$, and $\gamma = \Theta(1)$, $1-\gamma = \Theta(1)$, for $\tau < \frac{e^{-\alpha} }{2\gamma(1-\gamma)}$,

$$\lim_{n \to \infty} -\frac{m}{n^2 \eps^4} \log \left(\Pr_q\left[\Tilde{S} \le \tau \right] \right) =  \frac{\alpha^2  (2 \tau e^\alpha \gamma (\gamma-1) + 1)^2}{8(-1 - \alpha + e^\alpha) \gamma^2 (\gamma-1)^2}$$

\end{lemma}
\begin{proof}
Note that by Lemma \ref{lem:empty_bins_mgfs}, the limiting logarithmic moment generating function with respect to the uniform distribution $p$ is given by $$\Lambda_p(\theta) = \lim_{n \to \infty} \frac{m}{n^2 \eps^4} \log\left(M_{\Tilde S, p} \left(\frac{n^2 \eps^4}{m} \theta \right)\right) = \theta^2 \left[-\frac{e^{-2 \alpha}}{2\alpha} + \frac{e^{-\alpha}}{2\alpha^2} - \frac{e^{-2\alpha}}{2 \alpha^2} \right]$$

Thus, Assumption \ref{assumption:gartner_ellis} holds for $\mathcal{D}_{\Lambda_p} = \R$. Furthermore, the Fenchel-Legendre Transform (defined in equation \ref{eq:fenchel_legendre}) of $\Lambda_p$ is given by $$\Lambda_p^*(\tau) = \sup_{\theta } \left\{\theta \tau - \theta^2\left[-\frac{e^{-2 \alpha}}{2\alpha} + \frac{e^{-\alpha}}{2\alpha^2} - \frac{e^{-2\alpha}}{2 \alpha^2} \right]\right\} = \frac{\tau^2 \alpha^2 e^{2 \alpha}}{2e^\alpha - 2 - 2 \alpha }$$

This is a strongly convex function of $\tau$, so the set of exposed points of $\Lambda_p^*$ whose exposing hyperplane belongs to $\mathcal{D}_{\Lambda_p}^o$ is all of $\R$. Thus, by the  Theorem \ref{thm:gartner_ellis} (G{\"a}rtner-Ellis), for $\tau > 0$,

$$ \lim_{n \to \infty} -\frac{m}{n^2 \eps^4} \log\left(\Pr_p\left[\Tilde{S} \ge \tau\right]\right) = \inf_{x \ge \tau} \Lambda_p^*(x) = \frac{\tau^2 \alpha^2 e^{2 \alpha}}{2e^\alpha - 2 - 2 \alpha }$$

%

Similarly, the limiting logarithmic moment generating function with respect to an alternate distribution $q$ is given by $$\Lambda_q(\theta) = \lim_{n \to \infty} - \frac{m}{n^2 \eps^4} \log \left( \E_q\left[\exp\left(\frac{n^2 \eps^4}{m} \theta \Tilde{S} \right) \right] \right) = \theta^2 \left[-\frac{e^{-2 \alpha}}{2\alpha} + \frac{e^{-\alpha}}{2\alpha^2} - \frac{e^{-2\alpha}}{2 \alpha^2} \right] + \theta \frac{e^{-\alpha} }{2 \gamma (1-\gamma)}$$

The Fenchel-Legendre transform is given by $$\Lambda_q^*(\tau) = \sup_{\theta} \left\{\theta \tau - \theta^2 \left[-\frac{e^{-2 \alpha}}{2\alpha} + \frac{e^{-\alpha}}{2\alpha^2} - \frac{e^{-2\alpha}}{2 \alpha^2} \right] - \theta \frac{e^{-\alpha}}{2 \gamma (1-\gamma)}\right\} = \frac{\alpha^2 (2 \tau e^\alpha \gamma (\gamma-1) + 1)^2}{8(-1 - \alpha + e^\alpha) \gamma^2 (\gamma-1)^2}$$

Again, applying the G{\"a}rtner-Ellis Theorem gives, for $\tau < \frac{e^{-\alpha} }{2 \gamma (1-\gamma)}$,
$$\lim_{n \to \infty} - \frac{m}{n^2 \eps^4} \log\left(\Pr_q\left[\Tilde{S} _n^* \le \tau \right] \right) = \inf_{x \le \tau} \Lambda^*_q(x) = \frac{\alpha^2 (2 \tau e^\alpha \gamma (\gamma-1) + 1)^2}{8(-1 - \alpha + e^\alpha) \gamma^2 (\gamma-1)^2}$$


\end{proof}

\begin{lemma}
\label{lem:empty_bins_application_empty_bins_n_o(m)}
When $n = o(m)$, under the uniform distribution $p$, we have for $\tau > 0$, $$\lim_{n \to \infty} -\frac{m}{n^2 \eps^4} \log \left(\Pr_p \left[\Tilde{S} \ge \tau\right] \right) = \tau^2$$
and under an $\eps$-far distribution $q$ such that $q_j = \frac{1}{m} + \frac{\eps}{\gamma m}$ for $j \le \gamma m$ and $q_j = \frac{1}{m} - \frac{\eps}{(1-\gamma)m}$ for $ j>\gamma m$, for $\gamma = \Theta(1), 1 - \gamma = \Theta(1)$, and $\tau < \frac{1}{2 \gamma(1-\gamma)}$,
$$\lim_{n \to \infty} -\frac{m}{n^2 \eps^4} \log \left( \Pr_q \left[ \Tilde{S} \le \tau \right] \right) = \frac{(2 \tau \gamma(\gamma-1) + 1)^2}{4 \gamma^2 (\gamma-1)^2}$$
\end{lemma}

\begin{proof}
Since $n = o(m)$, we have by Lemma \ref{lem:empty_bins_mgfs}, Taylor expanding, $$M_{\Tilde{S}, p}\left(\frac{n^2 \eps^4}{m} \right) = \exp\left\{\frac{n^2 \eps^4}{m} \theta^2 \left[\frac{1}{4} + O(n/m)\right]\right\}$$
and similarly, $$M_{\Tilde S, q} \left( \frac{n^2 \eps^4}{m} \right) = \exp \left\{ \frac{n^2 \eps^4}{m} \theta \left[ \frac{1}{4}\theta + \frac{1}{2 \gamma(1-\gamma)} + O(n/m)\right]\right\}$$

Then, by a similar argument as in \ref{lem:gartner_ellis_application_empty_bins_n_Theta(m)}, the claim follows.
\end{proof}

\subsection{Setting the threshold}

We need to set our threshold $\tau$ so that the minimum of the error probability under the uniform distribution $p$, and any $\eps$-far distribution $q$ is maximized. Note that by Lemmas \ref{lem:gartner_ellis_application_empty_bins_n_Theta(m)} and \ref{lem:empty_bins_application_empty_bins_n_o(m)}, it is sufficient to consider a threshold $\tau$ such that $0 < \tau < \frac{e^{-\alpha}}{2\gamma(1-\gamma)}$ when $n = \Theta(m)$ and $0 < \tau < \frac{1}{2 \gamma (1-\gamma)}$ when $n = o(m)$, since otherwise, the error probability in one of the two cases is at least constant.  To set our threshold, we will first observe that for any $\tau$ in this range, the ``error exponent'' under $\eps$-far distributions is minimized for a particular $\eps$-far distribution. Then, we will set the threshold to maximize the minimum of the error exponent under the uniform distribution, and under this $\eps$-far distribution.

\begin{lemma}
\label{lem:threshold_empty_bins_n_Theta(m)}
When $n = \Theta(m)$ so that $\alpha = n/m$, setting the threshold $\tau = e^{-\alpha}$, we have for the uniform distribution $p$, $$\lim_{n \to \infty} -\frac{m}{n^2 \eps^4} \log \left( \Pr_p\left[\Tilde{S} \ge \tau \right] \right) = \frac{\alpha^2}{2(-1 - \alpha + e^\alpha)}$$
and for any $\eps$-far distribution $q$ such that $q_j = 1/m + \frac{\eps}{\gamma m}$ for $j \le \gamma m$ and $q_j = 1/m - \frac{\eps}{(1-\gamma)m}$ for $j > \gamma m$ and $\gamma = \Theta(1), 1-\gamma = \Theta(1)$,$$\lim_{n \to \infty} - \frac{m}{n^2 \eps^4} \log \left( \Pr_q \left[ \Tilde{S} \le \tau \right]\right) \ge \frac{\alpha^2}{2(-1 - \alpha + e^\alpha)}$$
 with equality for $q$ such that $q_j = 1/m + 2 \eps/m$ for $j \le m/2$ and $q_j = 1/m - 2 \eps/m$ for $j > m/2$.
\end{lemma}

\begin{proof}
By Lemma \ref{lem:gartner_ellis_application_empty_bins_n_Theta(m)}, for $0 < \tau < \frac{e^{-\alpha} }{2\gamma(1-\gamma)}$, $$\lim_{n \to \infty} \frac{m}{n^2 \eps^4} \log \left( \Pr_q \left[\Tilde{S} \le \tau \right]\right) = \frac{\alpha^2 (2 \tau e^\alpha \gamma (\gamma-1) + 1)^2}{8(- 1 - \alpha + e^\alpha) \gamma^2 (\gamma - 1)^2}$$

Now, the numerator of the right hand side is minimized when $\gamma = 1/2$, and the denominator is maximized when $\gamma = 1/2$. Thus, the right hand side is minimized when $\gamma= 1/2$. So, we have that,
$$\lim_{n \to \infty} -\frac{m}{n^2 \eps^4} \log\left(\Pr_{q}\left[ \Tilde{S} \le \tau \right] \right) \ge \frac{2 \alpha^2(1 - \frac{1}{2} \tau e^{\alpha})^2}{(-1 - \alpha + e^\alpha) }$$
 with equality for distribution $q$ such that $q_j = 1/m + 2 \eps/m$ for $j \le m/2$ and $q_j = 1/m - 2 \eps/m$ for $j > m/2$.
 
Then, the claim follows by substituting in $\tau = e^{-\alpha}$ in the expression for the uniform distribution in Lemma \ref{lem:gartner_ellis_application_empty_bins_n_Theta(m)} and in the above expression.
\end{proof}

\begin{lemma}
\label{lem:threshold_empty_bins_n_o(m)}
When $n = o(m)$, setting the threshold $\tau = 1$, we have for the uniform distribution $p$, $$\lim_{n \to \infty} -\frac{m}{n^2 \eps^4} \log\left(\Pr_p\left[\Tilde S \ge \tau\right] \right) = 1$$
and for any $\eps$-far distribution $q$ such that $q_j = 1/m + \frac{\eps}{\gamma m}$ for $j \le \gamma m$ and $q_j = 1/m - \frac{\eps}{(1-\gamma) m}$ for $j > \gamma m$, and $\gamma = \theta (1), 1-\gamma=\Theta(1)$,
$$\lim_{n \to \infty} -\frac{m}{n^2 \eps^4} \log\left( \Pr_q\left[ \Tilde S \le \tau\right]\right) \ge 1$$
 with equality for distribution $q$ such that $q_j = 1/m + 2 \eps/m$ for $j \le m/2$ and $q_j = 1/m - 2 \eps/m$ for $j > m/2$.
\end{lemma}

\begin{proof}
Follows from Lemma \ref{lem:empty_bins_application_empty_bins_n_o(m)} and using a similar argument as in Lemma \ref{lem:threshold_empty_bins_n_Theta(m)}.
\end{proof}

Finally, we have our results.

\tv*
\begin{proof}
Recall that our statistic is equivalent to the TV tester when $n \le m$. When $n = \Theta(m)$, by Lemma \ref{lem:threshold_empty_bins_n_Theta(m)}, we have that $\bar c(\eps, m, C) = 1$ for every $\eps$ that satisfies our assumptions, and every $C > 2$. In particular, any $\eps'$ such that $\left(1 - \frac{1}{C'} \right) \eps(n) \le \eps'(n) \le \eps(n)$ has $\bar{c}(\eps', m, C) = 1$ for every $C > 2$. Thus, by Lemma \ref{lem:final_worst_case_lemma}, we have that $c(\eps, m) = 1$ for every $\eps$ that satisfies our assumptions.

So, we have shown that our tester fails with probability $e^{-\frac{n^2 \eps^4}{m} (\xi + o(1))}$ for $$\xi = \frac{\alpha^2}{2(-1 - \alpha + e^\alpha)}$$
where $\alpha = n/m$. This implies that the TV tester uses $n = \frac{\sqrt{m \log(1/\delta)}}{\eps^2}(c + o(1))$ samples for failure probability $\delta$, where $$c = \frac{\sqrt{2 (e^\alpha - 1 - \alpha)}}{\alpha}$$ as required. Now, $c > 1$ for $0 \le \alpha \le 1$. The claim for the $n = \Theta(m)$ case follows. By a similar argument using Lemma~\ref{lem:threshold_empty_bins_n_o(m)}, the claim for the $n = o(m)$ case follows.
\end{proof}

\subsection{Empty bins lemmas}
We assume that Assumption~\ref{assumption:empty_bins} holds throughout this section.
\begin{lemma}
\label{lem:empty_bins_first_expansion}
Suppose Assumption~\ref{assumption:empty_bins} holds.
  \begin{equation*}
      \sum_{k=0}^\infty \left\{\frac{(\lam_0 \nu_j e^{i \psi})^k}{k!} e^{-\lam_0 \nu_j e^{i\psi}} \exp(\eps^2 \theta \1_{\{k = 0\}})\right\} = 1 + O(\eps^2)
  \end{equation*}
  for $\lam_0 = n(1 + O(\eps))$ and $\nu_j = 1/m + O(\eps/m)$ for all $j$,
\end{lemma}
\begin{proof}
By Lemma \ref{lem:empty_bins_indicator_mgf},
  \begin{align*}
      \sum_{k=0}^\infty \left\{\frac{(\lam_0 \nu_j e^{i \psi})^k}{k!} e^{-\lam_0 \nu_j e^{i\psi}} \exp(\eps^2 \theta \1_{\{k = 0\}})\right\} &= e^{-\lam_0 \nu_j e^{i \psi}} e^{\eps^2 \theta} - e^{-\lam_0 \nu_j e^{i \psi}} + 1\\
      &= e^{-\lam_0 \nu_j e^{i \psi}}(\eps^2 \theta + O(\eps^4)) + 1
  \end{align*}
  Substituting $\lam_0$ and $\nu_j$, this is
  \begin{align*}
      &1 + e^{-\frac{n}{m} e^{i \psi} (1 + O(\eps))} (\eps^2 \theta + O(\eps^4))\\
      &= 1 + O(\eps^2)
  \end{align*}
\end{proof}

\begin{lemma}
\label{lem:empty_bins_poissonized_mgf_bound}
Suppose Assumption~\ref{assumption:empty_bins} holds. For $\lam_0 = n(1 + e^{-n/m} \eps^2 \theta)$, $\nu_j = \frac{1}{m} + O\left(\frac{\eps}{m}\right)$ for all $j$,
$$\prod_{j=1}^m\left\{ e^{-\lam_0 \nu_j e^{i\psi}} e^{\eps^2 \theta} - e^{-\lam_0 \nu_j e^{i \psi}} + 1 \right\} = \exp\left\{m e^{-\frac{n}{m} e^{i \psi}} \eps^2 \theta (1 + O(\eps^2)) + O(n \eps^3)\right\}$$
\end{lemma}

\begin{proof}
By Lemma \ref{lem:empty_bins_indicator_mgf}, 
\begin{align*}
\prod_{j=1}^m \left\{ e^{-\lam_0 \nu_j e^{i \psi}} e^{\eps^2 \theta} - e^{-\lam_0 \nu_j e^{i \psi}} + 1\right\} &= \prod_{j=1}^m \left\{ e^{-\lam_0 \nu_j e^{i \psi}} (\eps^2\theta + O(\eps^4)) + 1\right\}\\
&= \exp\left\{\sum_{j=1}^m \log \left[ e^{-\lam_0 \nu_j e^{i \psi}} (\eps^2 \theta + O(\eps^4)) + 1\right] \right\}
\end{align*}
Substituting $\lam_0 = n(1 + O(\eps^2))$ and $\nu_j = 1/m + O(\eps/m)$, since $n = O(m)$, this is 
\begin{align*}
&= \exp\left\{\sum_{j=1}^m \log \left[ e^{-\frac{n}{m} e^{i \psi} (1 + O(\eps))} (\eps^2 \theta + O(\eps^4)) + 1\right] \right\}\\
&= \exp\left\{m \left[ e^{-\frac{n}{m} e^{i \psi} (1 + O(\eps))} (\eps^2 \theta + O(\eps^4))\right] \right\}\\
&= \exp\left\{m e^{-\frac{n}{m} e^{i \psi}} \eps^2 \theta (1 + O(\eps^2)) + O(n \eps^3)\right\}
\end{align*}

\end{proof}

\begin{lemma}
\label{lem:empty_bins_G_second_derivative}
Suppose Assumption~\ref{assumption:empty_bins} holds. For $\lam_0 = n(1 + e^{-n/m} \eps^2 \theta)$, $\nu_j = \frac{1}{m} + O(\frac{\eps}{m})$, and $$f(k) = \exp(\eps^2 \theta \1_{\{k = 0\}})$$ and $$G(\psi) = - i n \psi + \lam_0 e^{i \psi} + \sum_{j=1}^m \log \left\{e^{-\lam_0 \nu_j e^{i\psi}} e^{\eps^2 \theta} - e^{\lam_0 \nu_j e^{i\psi}} + 1 \right\}$$
we have $$G''(\psi) = - n e^{i \psi} + O(n \eps^2)$$
\end{lemma}

\begin{proof}
First, note that for $c > 0$, since $\1_{\{k+c = 0\}} = 0$ for every $k \ge 0$, $f(k+c) = 1$, for every $k \ge 0$. So,
$$\sum_{k=0}^\infty \left\{\frac{(\lam_0 \nu_j e^{i \psi})^k}{k!} e^{-\lam_0 \nu_j e^{i \psi}} f(k+c)  \right\} = 1$$

By Lemma \ref{lem:empty_bins_first_expansion}, we have that $$\sum_{k=0}^\infty\left\{ \frac{(\lam_0 \nu_j e^{i\psi})^k}{k!}e^{-\lam_0 \nu_j e^{i \psi}} f(k)\right\} = 1 + O(\eps^2)$$

So, we have 
\begin{align*}
&\left(\sum_{k=0}^\infty \frac{(\lam_0 \nu_j e^{i \psi})^k}{k!} e^{-\lam_0 \nu_j e^{i \psi}} f(k+1) \right)^2 \\
&- \left(\sum_{k=0}^\infty\frac{(\lam_0 \nu_j e^{i \psi})^k}{k!} e^{-\lam_0 \nu_j e^{i \psi}} f(k)\right)\left( \sum_{k=0}^\infty \frac{(\lam_0 \nu_j e^{i \psi})^k}{k!} e^{-\lam_0 \nu_j e^{i\psi}} f(k+2) \right) = O(\eps^2)
\end{align*}
and $$\left(\sum_{k=0}^\infty\frac{(\lam_0 \nu_j e^{i \psi})^k}{k!} e^{-\lam_0 \nu_j e^{i \psi}} f(k)\right)\left( \sum_{k=0}^\infty \frac{(\lam_0 \nu_j e^{i \psi})^k}{k!} e^{-\lam_0 \nu_j e^{i\psi}} f(k+2) \right)= 1 + O(\eps^2)$$ and $$\left(\sum_{k=0}^\infty \frac{(\lam_0 \nu_j e^{i \psi})^k}{k!} e^{-\lam_0 \nu_j e^{i \psi}} f(k) \right)^2 = 1 + O(\eps^2)$$

So, by Lemma \ref{lem:squared_huber_H_derivatives}, we have that 
$$G''(\psi) = \sum_{j=1}^m \frac{(\lam_0 \nu_j e^{i \psi})^2 O(\eps^2) - (\lam_0 \nu_j e^{i \psi}) (1 + O(\eps^2))}{1 + O(\eps^2)}$$
$$= \sum_{j=1}^m \left\{ O(\frac{n^2 \eps^2}{m^2}) - (\lam_0 \nu_j e^{i \psi}) (1 + O(\eps^2))\right\} (1 + O(\eps^2))$$
$$ = -(\lam_0 e^{i \psi}) (1 + O(\eps^2)) + O\left(\frac{n^2 \eps^2}{m}\right) = - n e^{i \psi} + O(n \eps^2)$$

since $n = O(m)$ and $\lam_0 = n(1 + O(\eps^2))$.
\end{proof}